\documentclass[preprint,12pt]{elsarticle}

\usepackage{amssymb}
\usepackage{amsmath}
\usepackage{amsthm}
\newtheorem{observation}{Observation}
\usepackage{booktabs}
\usepackage{multirow}
\usepackage{graphicx}
\usepackage{svg}
\usepackage{tcolorbox}
\usepackage{enumitem}
\usepackage{caption}
\usepackage{subcaption} 
\usepackage{wrapfig}
\usepackage{url}
\usepackage{enumitem}
\usepackage{placeins}
\usepackage{xcolor}
\usepackage{soul}
\usepackage{hyperref}
\usepackage{tabularx}
\usepackage{geometry}

\definecolor{delim}{RGB}{20,105,176}
\definecolor{numb}{RGB}{106, 109, 32}
\definecolor{string}{rgb}{0.64,0.08,0.08}
\definecolor{mainBlue}{HTML}{00677D}
\definecolor{darkBlue}{HTML}{004E5F}
\definecolor{darkBlueRGB}{rgb}{0,0.3,0.37}
\definecolor{mainGrey}{HTML}{83939A}
\definecolor{mainPurple}{HTML}{49007d}
\definecolor{mainRed}{HTML}{7d2600}
\definecolor{mainBrown}{HTML}{7d6400}
\definecolor{palecerulean}{rgb}{0.61, 0.77, 0.89}
\definecolor{mintbg}{rgb}{.63,.79,.95}

\DeclareRobustCommand{\hlyellow}[1]{{\sethlcolor{yellow!40}\hl{#1}}}
\DeclareRobustCommand{\hlblue}[1]{{\sethlcolor{mintbg!40}\hl{#1}}}

\journal{Computers \& Education}

\begin{document}

\begin{frontmatter}



%\title{LLMs Grade Essays Differently from Humans but Consistently with the Feedback They Generate} %% Article title

\title{LLMs Do Not Grade Essays Like Humans}
\author[uofa]{Jerin George Mathew\corref{cor1}}
\ead{jeringe1@ualberta.ca}

\author[uofa]{Sumayya Taher}
\ead{sntaher@ualberta.ca}

\author[uofa]{Anindita Kundu}
\ead{kundu2@ualberta.ca}

\author[uofa]{Denilson Barbosa}
\ead{denilson@ualberta.ca}

\cortext[cor1]{Corresponding author}

% Author affiliation
% \affiliation[uofa]{organization={University of Alberta},%Department and Organization
%             city={Edmonton},
%             state={AB},
%             country={Canada}}
\affiliation[uofa]{
    organization={University of Alberta},
    addressline={116 St \& 85 Ave}, 
    city={Edmonton},
    state={AB},
    postcode={T6G 2R3},
    country={Canada}
}

%% Abstract
\begin{abstract}
Large language models have recently been proposed as tools for automated essay scoring, but their agreement with human grading remains unclear. In this work, we evaluate how LLM-generated scores compare with human grades and analyze the grading behavior of several models from the GPT and Llama families in an out-of-the-box setting, without task-specific training. Our results show that agreement between LLM and human scores remains relatively weak and varies with essay characteristics. In particular, compared to human raters, LLMs tend to assign higher scores to short or underdeveloped essays, while assigning lower scores to longer essays that contain minor grammatical or spelling errors. We also find that the scores generated by LLMs are generally consistent with the feedback they generate: essays receiving more praise tend to receive higher scores, while essays receiving more criticism tend to receive lower scores. These results suggest that LLM-generated scores and feedback follow coherent patterns but rely on signals that differ from those used by human raters, resulting in limited alignment with human grading practices. Nevertheless, our work shows that LLMs produce feedback that is consistent with their grading and that they can be reliably used in supporting essay scoring.
\end{abstract}


% \begin{highlights}
% \end{highlights}

%% Keywords

\end{frontmatter}



\section{Introduction}
\label{sec:intro}

%% Labels are used to cross-reference an item using \ref command.

% Section text. See Subsection \ref{subsec1}.

%% Use \subsection commands to start a subsection.
% \subsection{Example Subsection}
% \label{subsec1}

% Subsection text.

Student assessment plays a central role in education, shaping how learning is measured and valued. Among the various assessment formats, essays are distinctive in that they require students to synthesize information and articulate their understanding in written form. In educational contexts, essays are commonly defined as \textit{``scholarly analytical or interpretative compositions with a specific focus on a phenomenon in question''}~\citep{ifenthaler2022automated}. Beyond evaluating factual knowledge, essay writing provides insights into the comprehension, critical thinking, and communication skills of students~\citep{gibbs2005conditions}. Consequently, essay assessment serves not only as a mechanism for measuring learning outcomes but also as a means of encouraging deeper engagement with course material and the development of analytical reasoning abilities~\citep{scouller1998influence,Joughin2009}.

Despite their pedagogical value, the evaluation of essays remains a demanding task. Traditionally, essays are graded by human instructors, a process that is both time-consuming and cognitively intensive. Providing careful evaluation and meaningful feedback requires instructors to read each essay closely, assess the quality of arguments and writing, and apply grading rubrics consistently across many submissions~\citep{ramesh2022automated}, which is hard and time-consuming. In modern educational settings, this challenge is amplified by ever-increasing class sizes, increasing student-teacher ratios, and the growing demand for scalable forms of education such as distance learning~\citep{fomba2023institutional}. At the same time, many countries are experiencing persistent shortages of qualified teachers, which further increases the workload placed on educators~\citep{koc2015impact}. Under such conditions, instructors often face substantial time pressure, which can delay feedback and limit the level of individualized attention students receive.

Human grading also introduces additional challenges related to subjectivity and consistency. Even when clear rubrics are provided, evaluations may vary across graders and over time, for the same grader. Cognitive biases, fatigue, and prior expectations can influence judgments, potentially leading to inconsistencies in how essays are assessed~\citep{williamson1999mental,benson2020unconscious}. These challenges have motivated research on automated approaches that can support or complement human evaluation in large-scale assessments.

To address these challenges, Automated Essay Scoring (AES) systems have been developed to assist in evaluating written responses. AES refers to the use of computer-based systems to automatically assign scores to essays without manual intervention. Early AES approaches relied on rule-based methods or statistical models built on hand-crafted linguistic features. More recently, machine learning and deep learning methods have been applied to learn scoring patterns from large collections of essays annotated by human graders~\citep{ramesh2022automated}. 

The emergence of large language models (LLMs) has introduced new possibilities for automated essay assessment. LLMs are trained on large collections of text and can understand and generate coherent natural language across a wide range of tasks, including text summarization~\citep{ouyang2022training}, reasoning~\citep{ko2024hierarchical}, and content evaluation~\citep{mizumoto2023exploring}. Because of these capabilities, LLMs can be prompted to assign essay scores directly, without requiring task-specific training, making them promising tools for scalable assessment. In addition to scoring, LLMs can generate textual feedback that highlights strengths and weaknesses in writing, such as clarity of argumentation and language quality. Such feedback not only offers interpretability for educators but also has the potential to guide students in improving their writing~\citep{stahl2024exploring,han2024llm}. 

Despite these advances, the reliability and interpretability of LLM-based essay evaluation remain open questions. While LLMs can be fine-tuned for specific tasks, in this study we focus on their out-of-the-box performance. Our goal is not to improve alignment through additional training, but to investigate whether LLMs, as provided by AI companies, produce scores and feedback that correspond meaningfully with human evaluations. We focus on this ``out-of-the box'' setting (as opposed to fine-tuning models for the task) because it currently readily available and because we believe this is how most educators will use such systems.
Moreover, while multiple studies~\citep{mansour2024can,sessler2025can,stahl2024exploring,kundu2024large} have examined the numerical agreement between automated systems and human raters, less attention has been paid to how model predictions relate to specific aspects of an essay. The relationship between specific essay elements, such as rubric traits or other signals of essay quality, and LLM-generated scores can provide insight into the signals reflected in model predictions.

\paragraph{Contribution} To address these gaps, we investigate the performance of both open and closed-source LLMs. Our evaluation setup includes GPT-3.5, GPT-4.1, GPT-5-mini, and Meta Llama 2, Llama 3 and Llama 4 models, examining how their scores align with human graders and reflect the presence of rubric-defined traits. Specifically, we aim to answer the following research questions:

\begin{enumerate}[label=\textbf{RQ\arabic*:}, leftmargin=*]
    \item \textbf{Do LLMs grade similarly to humans?}  
    We evaluate the extent to which LLM-generated scores agree with human grades across datasets, and analyze how differences between LLM and human grading vary with essay characteristics such as length and the presence of language errors.

    \item \textbf{Do LLMs exhibit consistent grading behavior across essays?}  
    We investigate whether the scores assigned by LLMs are coherent with the feedback they generate, and whether positive and negative evaluative signals in the feedback systematically correspond to higher or lower essay scores.
\end{enumerate}

Our findings indicate that LLM-assigned scores only partially align with human grades. In particular, the agreement varies with essay quality: shorter or poorly developed essays tend to receive lower scores from humans compared to LLMs, whereas longer or more developed essays often receive higher human scores despite minor errors. In addition, our results suggest that LLM-generated scores are consistent with the feedback they produce: feedback emphasizing positive aspects tends to coincide with higher scores, while feedback highlighting criticisms generally corresponds to lower scores. This suggests that LLM-generated scores are consistent with the evaluative signals present in the generated feedback.

\paragraph{Outline}The remainder of this paper is structured as follows. In Section~\ref{sec:related}, we review related work on automated essay scoring and the evaluation of LLMs for AES. Section~\ref{sec:methodology} introduces the proposed framework used in this study. Section~\ref{sec:experiments} presents the experimental setup and reports the results of our evaluation. In Section~\ref{sec:discussion}, we discuss the main findings and their implications for automated essay scoring. Finally, Section~\ref{sec:conclusions} concludes the paper and outlines directions for future work.
\section{Related Work}
\label{sec:related}
In this section, we review prior work related to AES and the use of LLMs for essay evaluation. We organize the literature into three main areas: (i) traditional AES systems, (ii) the use of LLMs for essay scoring, and (iii) previous studies evaluating the performance of LLMs on this task.

\paragraph{Automatic essay scoring} Automated essay scoring has been studied for more than five decades, beginning with early systems such as Project Essay Grade (PEG) \citep{page-1966}. Early AES systems relied primarily on handcrafted linguistic features and statistical models to approximate human scoring patterns. For example, systems such as e-rater \citep{attali2006automated} and the Intelligent Essay Assessor (IEA) \citep{foltz1999intelligent} used features capturing aspects such as lexical diversity, syntactic complexity, and essay length to predict scores. While effective in some cases, these approaches have been criticized for relying heavily on surface-level textual features rather than deeper aspects of writing quality.

With the development of machine learning methods, AES systems increasingly adopted neural architectures capable of learning representations directly from text. These include models based on recurrent neural networks (RNNs), convolutional neural networks (CNNs), and long short-term memory networks (LSTMs) \citep{taghipour-ng-2016-neural, dong2017attention}. Some work also combines neural representations with handcrafted linguistic features to improve performance \citep{uto2020neural_handcraft}.

Progress in this area has been facilitated by the availability of annotated datasets such as the Automated Student Assessment Prize (ASAP) dataset~\citep{asap-aes}, one of the most widely used benchmarks for AES research. Essay scoring datasets typically consist of a (i) writing prompt, (ii) the student essay responding to that prompt, and (iii) scores assigned by one or more human raters. Depending on the dataset, these scores may correspond to a single holistic grade reflecting overall essay quality, or to multiple analytic trait scores evaluating specific aspects of writing such as grammar, coherence, organization, or argumentation. Other datasets used in AES research include  ICLE++~\citep{granger2003international,li2024icle++}, ASAP++~\citep{mathias2018asap++}, ICNALE GRA~\citep{ishikawa2020aim,ishikawa2023icnale}, and CELA~\citep{xue2021hierarchical}, which provide essays annotated with either holistic or analytic trait scores \citep{ke2019automatedsurvey}. These datasets differ substantially in scoring rubrics, trait definitions, and essay prompts, making cross-study comparisons challenging.

\paragraph{LLMs for AES} Recent advances in large language models (LLMs) have introduced new approaches to automated essay scoring. Unlike earlier AES systems that relied on task-specific architectures and extensive feature engineering, LLMs leverage large-scale pretraining to capture general linguistic and semantic knowledge that can be adapted to essay evaluation tasks. Early work explored adapting pretrained transformer models~\citep{vaswani2017attention} such as BERT~\citep{devlin2019bert} to AES through supervised fine-tuning. For example, prior studies proposed BERT-based architectures that learn essay representations and predict scores through regression or ranking objectives \citep{wang-bert-use, yang-etal-2020-enhancing}. Other work further improved performance by combining neural representations with traditional linguistic features or by leveraging transfer learning for long-text scoring tasks.

More recently, prompt-based approaches to essay scoring using LLMs have been explored. In these approaches, the model receives a scoring rubric and task description in natural language and produces a score in zero-shot or few-shot settings. For instance, \citet{mizumoto2023exploring} evaluated GPT-3 on essays from the TOEFL11 corpus and showed that LLM-based scoring exhibited moderate agreement with human graders. A subsequent work by \citet{mansour2024can} showed that model performance is sensitive to prompt design, with prompting strategies such as few-shot examples or chain-of-thought reasoning improving scoring accuracy.

Beyond score prediction, several studies explored using LLMs to generate feedback or explanations for student writing. Some approaches train models to predict scores for individual traits of an essay, providing more detailed feedback to students \citep{Hussein2020, kumar2021many}. Other work investigates the use of generative models to produce hints or comments aimed at supporting student revision \citep{LLMonassignments, MEYER2024100199}. These developments highlight the broader potential of LLMs not only for automated grading but also for supporting formative assessment and writing improvement.

\paragraph{Evaluating LLMs for AES} Research evaluating LLMs agreement with human raters for AES remains relatively limited. Existing studies report mixed results, often due to differences in datasets, languages, evaluation metrics, and experimental protocols. For example, \citet{mansour2024can} evaluated ChatGPT and Llama 2 on the ASAP benchmark using several prompting strategies and measured agreement with reference essay scores using Quadratic Weighted Kappa. While both models achieved moderate agreement levels, they still lagged behind specialized AES systems, although the authors noted that LLMs can generate useful feedback to support essay revision. Other studies report substantially weaker agreement with human grading. For instance, \citet{gaggioli2025assessing} evaluated multiple LLMs on Italian university essays scored across four rubric dimensions. Their results showed consistently low agreement between LLM and human scores (near-zero weighted Kappa scores) and limited repeatability across runs, suggesting that LLMs struggle to replicate human judgments in context-specific evaluation settings. Similarly, \citet{sessler2025can} compared several LLMs on German middle-school essays across ten rubric dimensions and found that closed-source GPT models achieved higher consistency and stronger correlation with teacher ratings than open-source models, although they tended to overestimate content quality and required further calibration. Another work has also examined the performance of LLMs for essay grading on the ASAP benchmark and reported weak agreement between LLM-generated and human-assigned scores, with models often assigning  different grades than human raters~\citep{kundu2024large}.

\paragraph{Discussion} These studies present a mixed picture of LLM performance for automated essay scoring. While LLMs can approximate human grading in some settings, agreement with human raters remains inconsistent and highly dependent on the dataset and evaluation setup. This raises the question of how LLM-generated scores relate to the aspects of writing that human raters consider when grading essays. In addition to evaluating agreement using standard metrics, our study examines the nature of this divergence by analyzing how LLMs assign scores, how these scores relate to the feedback they generate, and which traits they emphasize when evaluating essays.
\section{Methodology}
\label{sec:methodology}

\begin{figure}[!t]
\centering
\includegraphics[width=0.8\textwidth]{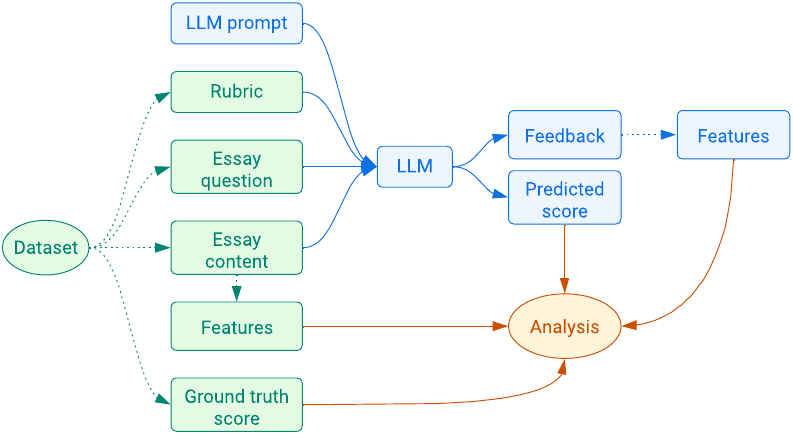}
\caption{Overview of the proposed analysis framework. Essays are first evaluated by LLMs to produce both predicted scores and textual feedback. Essay-level features and feedback-derived signals are then extracted and analyzed to compare LLM scores with human ratings and to examine how feedback sentiment and essay characteristics relate to the scores assigned by the models.}\label{fig:methodology}
\end{figure}

Figure~\ref{fig:methodology} illustrates the overall framework used in this study. The framework combines model-generated feedback with features extracted from the essay in order to study how LLM scoring behavior relates to both essay content and the feedback produced by the models. The process consists of three main stages: (i) obtaining LLM-based scoring and feedback, (ii) extracting features from essays and generated feedback, and (iii) analyzing the relationship between scores, feedback, and essay characteristics.

\paragraph{Data}
Each example in the dataset consists of an essay prompt, a student essay response, a scoring rubric, and a score assigned by human raters. The essay prompt defines the writing task, the rubric specifies the evaluation criteria, and the human score serves as the reference used to evaluate the behavior of LLM-based grading.

\paragraph{LLM-based essay scoring}
To obtain automated evaluations, the LLM is prompted with the essay prompt, the scoring rubric, and the student essay response. We explicitly ask the LLM to output both a numeric score and written feedback explaining that score according to the rubric. The predicted score represents the assessment of essay quality produced by the model, while the generated feedback provides a textual explanation of the evaluation.

\paragraph{Feature extraction}
The generated feedback is used to derive feedback-based features capturing signals in the evaluation produced by the model. In particular, trait-specific mentions of praise and criticism related to rubric dimensions (e.g. grammar) are extracted from the feedback, and the associated sentiment is analyzed. 

In parallel, we extract a small set of essay-level characteristics from the essay text. These features capture properties of the writing itself and are independent of the feedback produced by the LLM. We consider indicators such as essay length, readability, and the number of grammatical or spelling errors in our analysis.

\paragraph{Analysis}
Finally, the predicted scores, feedback-based features, and essay-level characteristics are used in the analysis stage. First, we compare the predicted scores with those assigned by human raters to measure agreement between LLM-generated scores and human grading, while examining how differences between the two relate to essay characteristics such as length, readability, and language errors. Second, we examine the extracted feedback features to determine whether the sentiment expressed in trait-related feedback is consistent with the scores assigned by the models.

\section{Experiments}
\label{sec:experiments}

\subsection{Datasets}
We conduct our experiments on two essay datasets commonly used in AES: ASAP  and DREsS. Table~\ref{tab:datasets} summarizes key statistics of the datasets used in our experiments.

\paragraph{ASAP} Originally released as part of a Kaggle competition, the Automated Student Assessment Prize (ASAP) benchmark~\citep{asap-aes} contains essays written in English by students in grades 7-10 across eight prompts covering different writing tasks. Each essay is graded by multiple human raters according to prompt-specific rubrics. To protect privacy, personally identifiable information in the essays were anonymized by replacing entities such as names, locations, or organizations with NER placeholders (e.g.  \texttt{@PERSON1}, \texttt{@ORGANIZATION1}). 

\begin{figure}[!ht]
\centering

\begin{subfigure}{\textwidth}
\centering
\includegraphics[width=\textwidth]{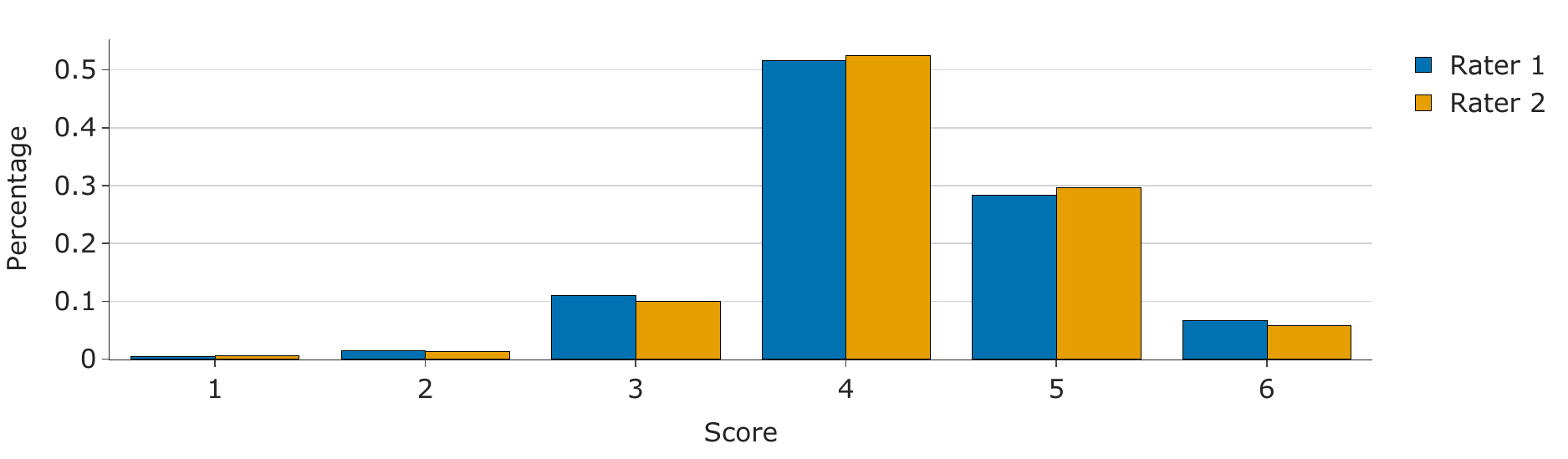}
\caption{ASAP Task 1}
\label{fig:asap_1_score_histogram}
\end{subfigure}

\vspace{0.5cm}

\begin{subfigure}{\textwidth}
\centering
\includegraphics[width=\textwidth]{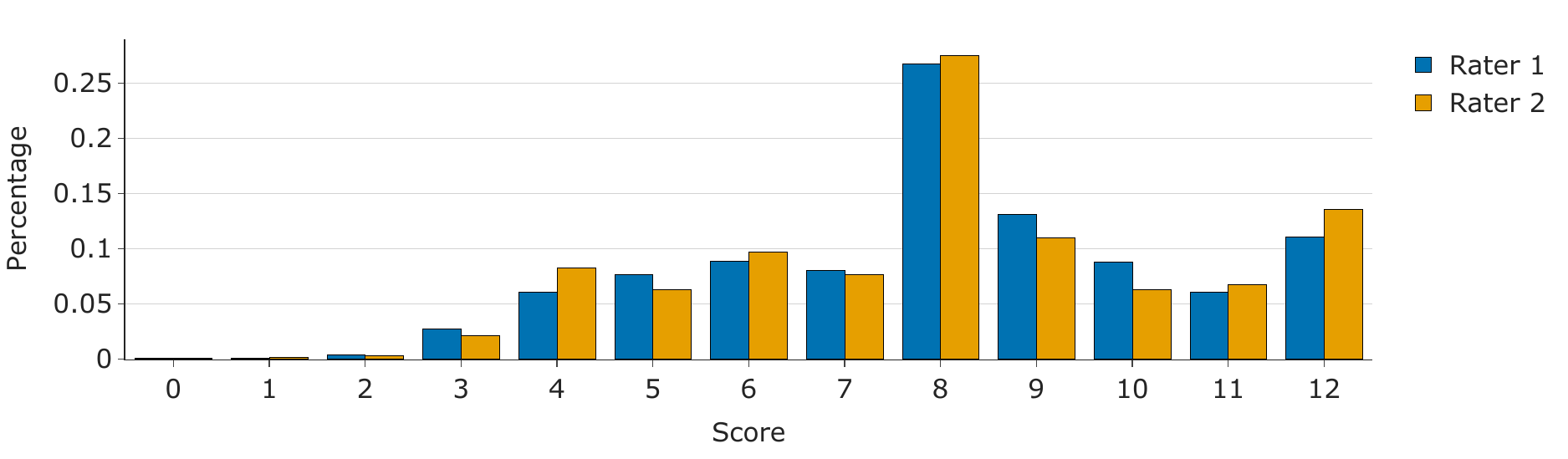}
\caption{ASAP Task 7}
\label{fig:asap_7_score_histogram}
\end{subfigure}

\vspace{0.5cm}

\begin{subfigure}{\textwidth}
\centering
\includegraphics[width=\textwidth]{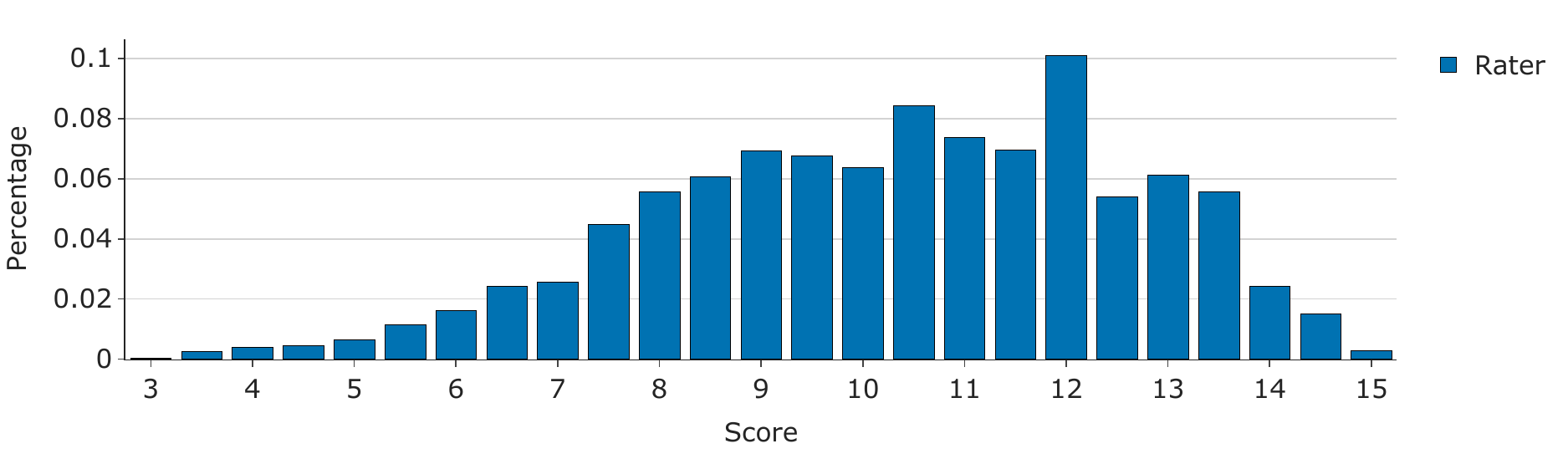}
\caption{DREsS}
\label{fig:dress_score_histogram}
\end{subfigure}

\caption{Distribution of human-assigned essay scores across datasets.}
\label{fig:score_histograms}
\end{figure}

From this benchmark, we use the Task 1 and Task 7 subsets, which differ in their scoring schemes and rubric structure. Task 1 contains 1,783 essays graded holistically by two raters on a scale from 1 to 6. Task 7 contains 1,569 essays evaluated using trait-based scoring, where raters assign scores for multiple writing aspects, including \textit{ideas}, \textit{organization}, \textit{style}, and \textit{conventions}, resulting in an aggregated score between 0 and 12. The distributions of scores assigned by the raters are shown in Figure~\ref{fig:asap_1_score_histogram} and Figure~\ref{fig:asap_7_score_histogram}. The essay prompts and scoring rubrics for ASAP Task 1 and Task 7 are provided in ~\ref{sec:dataset_prompts_rubrics} (Figures~\ref{fig:asap_1_rubric_prompt} and~\ref{fig:asap_7_rubric_prompt}). Since Task 1 is graded holistically, the  original ASAP dataset does not provide trait-level grades for Task 1. To enable trait-level analysis for this subset, we use the ASAP++ dataset~\citep{mathias2018asap++}, which adds manual trait-level grades (e.g. \textit{organization}, \textit{word choice} and \textit{sentence fluency}) for prompts that originally contained only holistic scores. Each trait is scored on a 1-6 scale by a separate annotator, allowing trait-level analysis for Task 1. The list of traits for each dataset subset is provided in~\ref{sec:trait_terms} (Table~\ref{tab:trait_terms}).

\paragraph{DREsS}
The DREsS (Dataset for Rubric-based Essay Scoring) dataset~\citep{yoo2025dress} is designed for rubric-based automated essay scoring in English-as-a-Foreign-Language (EFL) writing and provides trait-level grades aligned with educational assessment rubrics. The dataset consists of three subsets: (i) \textit{DREsS\_New}, a dataset of essays collected from undergraduate EFL students and scored by English education experts, (ii) \textit{DREsS\_Std}, which standardizes several existing rubric-based AES datasets under a unified scoring scheme; and (iii) \textit{DREsS\_CASE}, a large synthetic dataset generated using corruption-based essay augmentation. In this work, we use DREsS\_New, which contains essays written in response to multiple prompts and annotated with rubric-based scores for three traits: \textit{content}, \textit{organization}, and \textit{language}, each scored on a 1-5 scale with 0.5 increments. The original split contains 2,279 essays. However, some entries have empty essay texts, which we filter out, resulting in 1,917 essays used in our experiments. As summarized in Table~\ref{tab:datasets}, this subset covers 59 prompts, with trait scores aggregated into an overall score ranging from 3 to 15. The distribution of scores for this subset is shown in Figure~\ref{fig:dress_score_histogram}, while the scoring rubric is provided in~\ref{sec:dataset_prompts_rubrics} (Figure~\ref{fig:dress_rubric_prompt}).

\begin{table}[]
\caption{Summary of the datasets used in the experiments.}
\label{tab:datasets}
\resizebox{\textwidth}{!}{%
\begin{tabular}{@{}llllllllll@{}}
\toprule
\textbf{Name} & \textbf{Subset} & \textbf{\begin{tabular}[c]{@{}l@{}}Num \\ essays\end{tabular}} & \textbf{\begin{tabular}[c]{@{}l@{}}Essay length \\ (words)\end{tabular}} & \textbf{\begin{tabular}[c]{@{}l@{}}Num \\ prompts\end{tabular}} & \textbf{\begin{tabular}[c]{@{}l@{}}Num \\ raters\end{tabular}} & \textbf{\begin{tabular}[c]{@{}l@{}}Score \\ range\end{tabular}} & \textbf{\begin{tabular}[c]{@{}l@{}}Score \\ step\end{tabular}} & \textbf{\begin{tabular}[c]{@{}l@{}}Num \\ traits\end{tabular}} & \textbf{\begin{tabular}[c]{@{}l@{}}NER \\ tags\end{tabular}} \\ \midrule
\multirow{2}{*}{ASAP} & Task 1 & 1,783 & 410.30 $\pm$ 136.18 & 1 & 2 & 1--6 & 1 & 5 & Yes \\
 & Task 7 & 1,569 & 191.67 $\pm$ 100.12 & 1 & 2 & 0--12 & 1 & 4 & Yes \\
DREsS & New & 1,917 & 360.65 $\pm$ 107.01 & 59 & 1 & 3--15 & 0.5 & 3 & No \\ \bottomrule
\end{tabular}
}
\end{table}

\subsection{Models}
\begin{figure}[!t]
\centering
\begin{subfigure}{\textwidth}
    \centering
    \includegraphics[width=\textwidth]{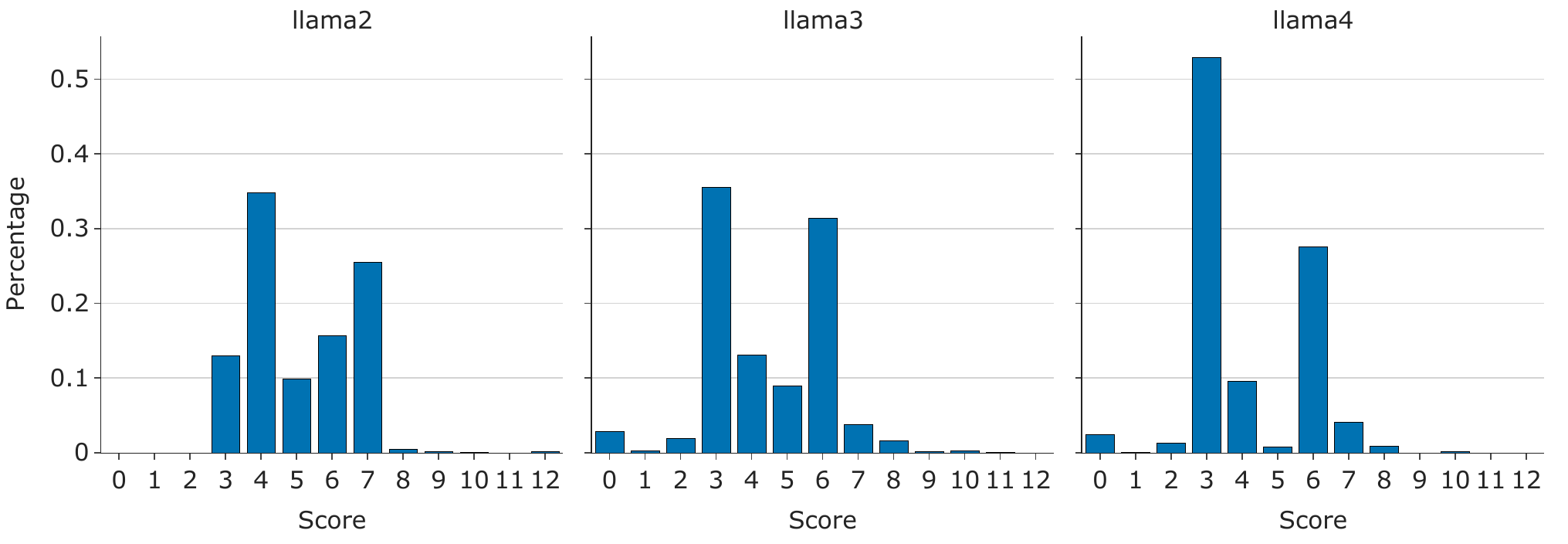}
    \caption{Llama}
\end{subfigure}
\vspace{0.5cm}
\begin{subfigure}{\textwidth}
    \centering
    \includegraphics[width=\textwidth]{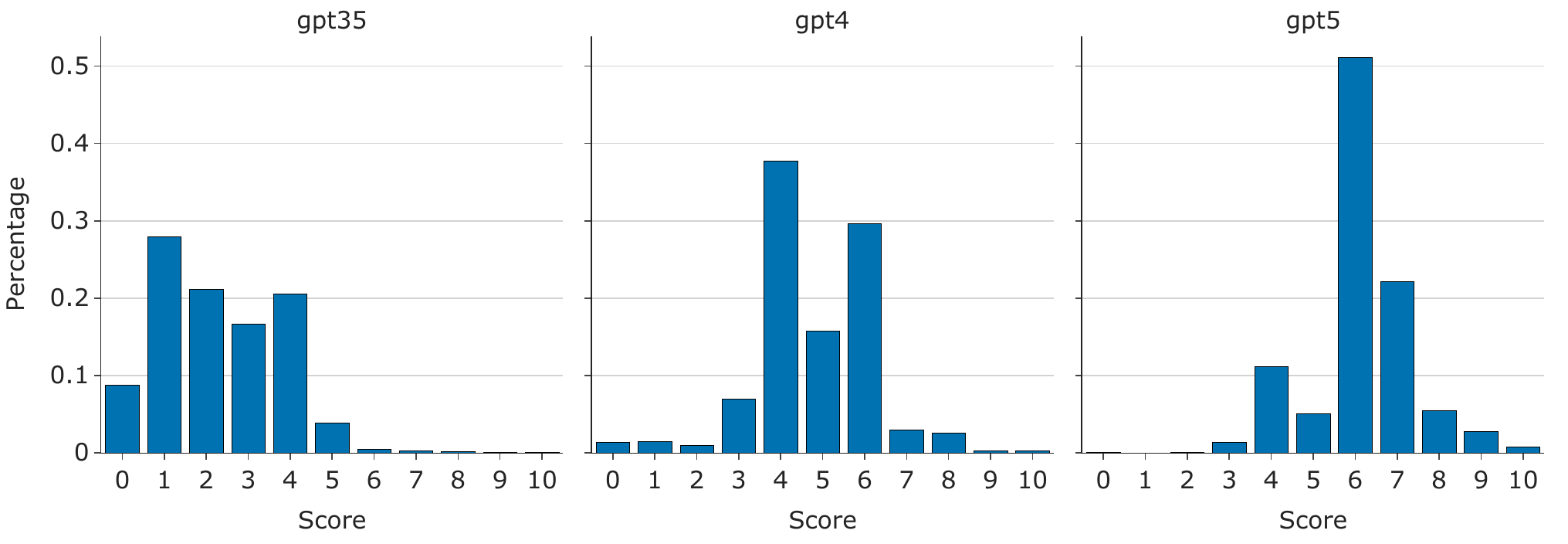}
    \caption{GPT}
\end{subfigure}
\caption{Distribution of essay scores assigned by Llama and GPT models on the ASAP Task 7 dataset. Most models produce scores concentrated around the middle of the scale, while GPT-3.5 assigns comparatively lower scores with predictions concentrated toward the lower end.}
\label{fig:score_distribution_asap_7_llama_gpt}
\end{figure}

We evaluate multiple versions of two widely used LLM families: GPT and Llama. For the GPT family, we use GPT-3.5, GPT-4.1, and GPT-5 mini, which we refer to as GPT-3.5, GPT-4, and GPT-5, respectively. For the Llama family, we use Llama 2 70B, Llama 3 70B, and Llama 4 Scout, which we denote as Llama 2, Llama 3, and Llama 4. For each essay in the ASAP and DREsS datasets, the input to the model follows a standardized prompt template (Figure~\ref{fig:base_llm_prompt_template} in~\ref{sec:llm_prompts}), that includes the essay prompt, the evaluation rubric, the scoring range, and the essay text. The prompt instructs the LLM to assign a numeric score reflecting essay quality and to generate a short explanation justifying the evaluation according to the rubric. For ASAP Task 1, the models are asked to produce a single holistic score together with an explanation. For ASAP Task 7 and DREsS, the models are asked to generate scores for each rubric trait as well as the overall score, each accompanied by a short explanation. All inferences are performed using the default generation parameters of the respective models. Score distributions for ASAP Task~7 are shown in Figure~\ref{fig:score_distribution_asap_7_llama_gpt}. The corresponding distributions for ASAP Task 1 and DREsS are provided in~\ref{sec:llm_score_distributions}. 

\subsection{Feature extraction}  
\label{subsec:implementation_details}
%In this section we describe the implementation of the feature extraction components used in our analysis.

\paragraph{Essay level features} We extract three types of features from each essay: length, readability, and language quality. Essay length is measured by the total number of tokens, supported by the assumption that longer essays are likely to contain more and better arguments while very short essays might lack sufficient detail. Readability is measured using the Flesch Reading Ease (FRE) score~\citep{kincaid}. FRE is a commonly used readability metric that typically ranges from 0 to 100, with higher values indicating easier-to-read text. In some cases the score may exceed 100, corresponding to very simple text with short sentences. Finally, language mistakes are identified and categorized using \textit{LanguageTool}\footnote{\url{https://languagetool.org}}. The tool detects several types of writing errors, including spelling, grammar, punctuation, style, and capitalization mistakes. For each essay, we count the number of detected errors in each category.

\paragraph{LLM feedback-level features} To analyze the feedback generated by the LLMs, we extract trait-specific mentions of praise and criticism using an aspect-based sentiment analysis (ABSA) model. First, for each dataset we manually define a set of trait-related terms corresponding to the rubric dimensions. These terms are identified by inspecting a sample of 50 feedback responses generated by different LLMs for each dataset. The resulting lists include multiple lexical variants associated with each rubric trait (e.g., \textit{content}, \textit{organization}, \textit{grammar}) and is reported in \ref{sec:trait_terms}. The feedback text is processed using spaCy~\citep{honnibal2020spacy} for sentence segmentation and lemmatization. Sentences containing trait-related terms are identified using exact or lemma-based matching. These sentence are then are then classified using a pretrained ABSA model\footnote{\url{https://huggingface.co/yangheng/deberta-v3-base-absa-v1.1}} based on DeBERTa-v3~\citep{hedebertav3}. The model predicts the sentiment associated with the trait mention (\textit{positive}, \textit{negative}, or \textit{neutral}). To reduce noise, we retain only predictions with confidence above 0.9 and discard neutral predictions. The resulting mentions are aggregated to obtain counts of positive and negative feedback associated with each rubric trait.

\subsection{Evaluation metrics}
\label{subsec:metrics}
To evaluate how closely LLM-generated scores align with human grades, we use three metrics commonly adopted in automated essay scoring (AES): Quadratic Weighted Kappa (QWK), Pearson correlation, and Mean Absolute Error (MAE). QWK measures the level of agreement between two raters while accounting for agreement expected by chance and penalizing larger disagreements more heavily. The metric ranges from -1 (complete disagreement) to 1 (perfect agreement), with values closer to 1 indicating stronger agreement. Pearson correlation measures the strength of the linear relationship between predicted and human scores, with values ranging from -1 to 1. Higher values indicate that essays receiving higher human scores also tend to receive higher model scores. However, Pearson correlation captures only the strength of this relationship and does not directly measure exact agreement between raters. Finally, MAE measures the average absolute difference between predicted and reference scores, where lower values indicate better alignment. Because the datasets used in our experiments employ different scoring ranges, we report normalized MAE, obtained by dividing the MAE by the score range (maximum minus minimum possible score).

\begin{figure}[!ht]
\centering
\includegraphics[width=\textwidth]{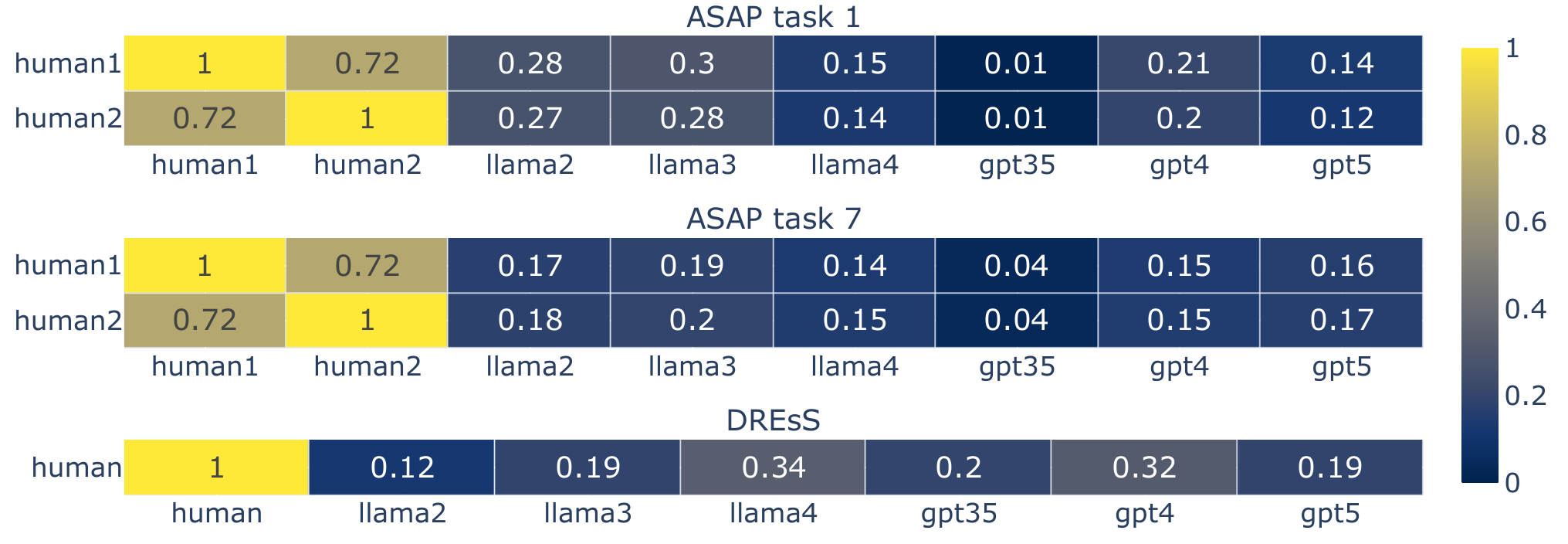}
\caption{Quadratic Weighted Kappa (QWK) scores between LLM-generated scores and human ratings across datasets. Lighter cells indicate greater agreement. While human raters exhibit relatively strong agreement with each other, agreement between LLM predictions and human scores remains generally lower across models and datasets.}\label{fig:kappa}
\end{figure}

\begin{figure}[!ht]
\centering
\includegraphics[width=\textwidth]{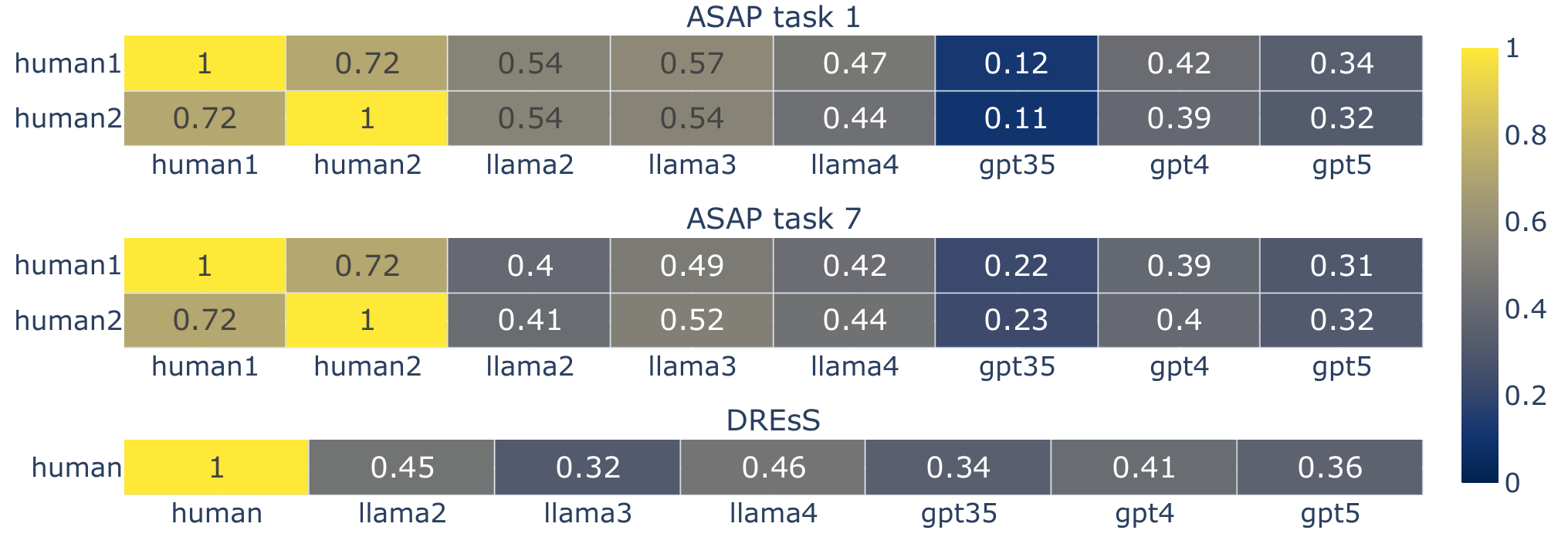}
\caption{Pearson correlation between LLM-generated scores and human ratings across datasets. Lighter cells indicate greater agreement. Although some models show moderate correlations with human scores, the overall agreement remains lower than the correlation observed between human raters.}\label{fig:pearson}
\end{figure}

\begin{figure}[!ht]
\centering
\includegraphics[width=\textwidth]{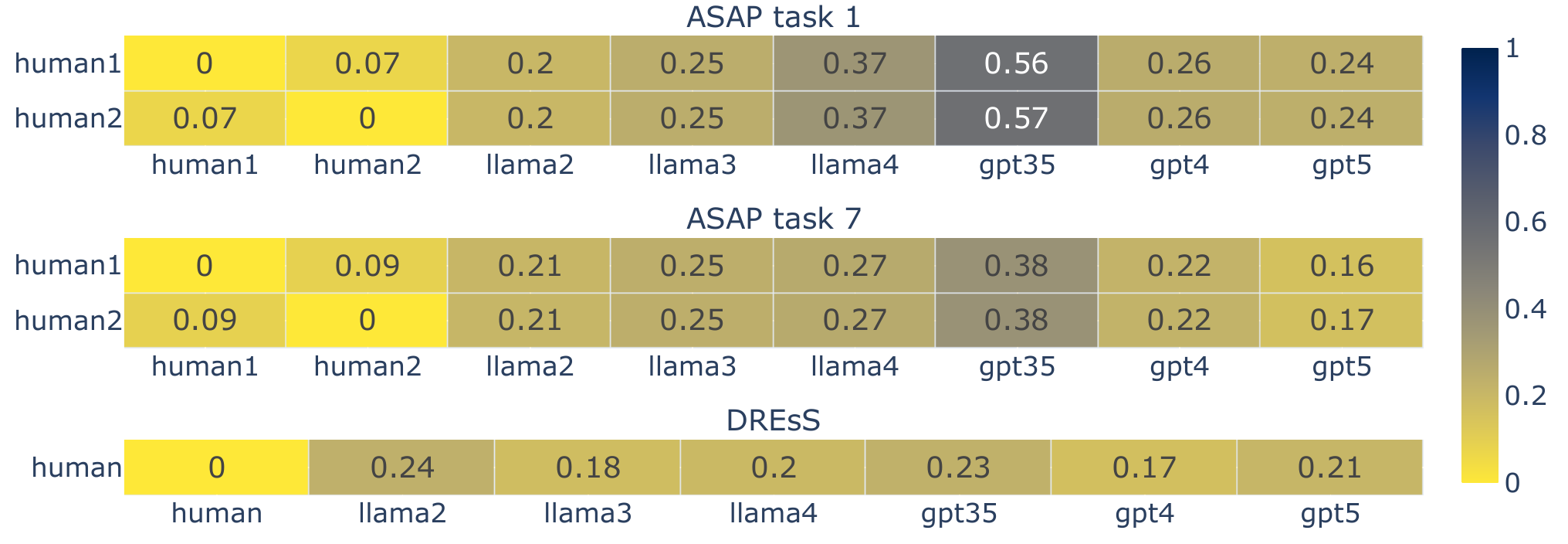}
\caption{Mean Absolute Error (MAE) between LLM-generated scores and human ratings across datasets. Lighter cells indicate greater agreement (smaller differences between scores). Across datasets and models, LLM predictions consistently exhibit larger errors than the differences observed between human raters.}\label{fig:mae}
\end{figure}

\subsection{Experimental results}
\label{subsec:experimental_results}
In this section, we present the results of our experiments and analyze how closely LLM-generated scores align with human grades across the datasets introduced earlier. Our analysis is organized around the two research questions introduced in Section~\ref{sec:intro}: (i) whether LLMs assign scores similarly to human raters (RQ1), and (ii) whether the scores produced by LLMs are consistent with the feedback they generate (RQ2).

\subsection*{RQ1: LLMs exhibit weak agreement with human raters}

We compare LLM-assigned grades with human scores using the metrics defined in Section~\ref{subsec:metrics}. Results for all datasets are shown in Figures~\ref{fig:kappa},  \ref{fig:pearson} and \ref{fig:mae}.

Across all datasets, LLMs show substantially weaker agreement with human raters than the agreement observed between human annotators. This trend is clearly visible in the QWK scores shown in Figure~\ref{fig:kappa}. For the ASAP datasets, the agreement between the two human raters is relatively high ($QWK=0.72$), indicating strong inter-rater reliability. In contrast, all LLMs achieve considerably lower agreement with human scores, with QWK values below 0.30 across models and tasks. A similar pattern is observed on the DREsS dataset, where the highest agreement with human scores reaches only 0.34. Pearson correlation results (Figure~\ref{fig:pearson}) shows a similar trend. While human raters exhibit a strong correlation on the ASAP datasets ($r=0.72$), correlations between LLM predictions and human scores remain substantially lower across models. The highest correlations are observed for Llama 3 on ASAP Task 1 ($r=0.57$) and Llama-4 on DREsS ($r=0.46$), but overall correlations remain moderate at best. Mean Absolute Error (MAE) results in Figure~\ref{fig:mae} further confirm this mismatch. The disagreement between the two human raters on the ASAP datasets is relatively small ($MAE=0.07$-$0.09$), whereas the error between LLM predictions and human scores is consistently higher. Depending on the model and dataset, MAE values range from approximately 0.16 to 0.57 on ASAP and from 0.17 to 0.24 on DREsS. 

Among the evaluated models, GPT-3.5 consistently shows the weakest alignment with human raters. For example, on ASAP Task 1 it achieves a QWK of only 0.01, substantially lower than the other models, and also exhibits the lowest Pearson correlation and the second-highest MAE across datasets. This behavior is partly reflected in the distribution of scores assigned by the model, shown in Figure~\ref{fig:score_distribution_asap_7_llama_gpt} for ASAP Task 7. Compared to human raters, GPT-3.5 assigns lower scores more frequently and produces substantially fewer high-score predictions, resulting in a distribution that is skewed toward the lower end of the grading scale. In general, we observe differences in how models use the available score range. While some models produce relatively concentrated score distributions, others generate a wider spread of grades, indicating that different LLMs may use distinct implicit grading criteria when evaluating essays.

\begin{observation}
LLM-human score agreement is substantially weaker than human inter-rater agreement across all datasets and models, with different LLMs exhibiting distinct score distributions that suggest divergent implicit grading criteria.
\end{observation}

Also, we observe no consistent improvement across successive model generations.  While the GPT family shows an overall reduction in MAE from GPT-3.5 to GPT-5, intermediate versions do not consistently follow this trend across datasets. For instance, GPT-4 achieves the lowest MAE on the DREsS dataset, whereas GPT-5 performs better on the ASAP datasets. A similar inconsistency is observed for the Llama models, where Llama 4 shows lower agreement than earlier versions on the ASAP datasets but higher agreement on DREsS. These results suggest the following:

\begin{observation}
Improvements in general language modeling capability do not necessarily translate into better agreement with human essay grading across different datasets.
\end{observation}

In addition to the base prompt configuration, we evaluated two prompt variants. The first, \textit{student-grade}, extends the base prompt template by explicitly specifying the student's grade level (e.g., 8th grade), as shown in in~\ref{sec:llm_prompts} (Figure~\ref{fig:student_grade_llm_prompt_template}). The second, \textit{few-shot}, includes two example essays together with their human-assigned scores (Figure~\ref{fig:few_shot_llm_prompt_template} in~\ref{sec:llm_prompts}). In some cases these variants lead to modest improvements in agreement with human scores, particularly for the few-shot setting, across QWK, Pearson correlation, and MAE. However, these improvements are not consistent across datasets or models.

\begin{observation}
Prompt variants such as few-shot prompting can modestly improve LLM-human score agreement, but these gains are not consistent across datasets or models.
\end{observation}

Detailed results are reported in Appendix~\ref{sec:human_llm_agreement_prompt_variants} (Figures~\ref{fig:prompt_variant_agreement_with_human_asap1}, \ref{fig:prompt_variant_agreement_with_human_asap7}, and \ref{fig:prompt_variant_agreement_with_human_dress}).

We also examine agreement between LLMs themselves in~\ref{sec:llm_llm_prompt_agreement}. We observe that agreement between LLMs varies substantially across model pairs. In many cases, models belonging to the same family exhibit relatively strong agreement with each other. For example, successive generations of GPT models (e.g., GPT-4 and GPT-5) show high correlations across prompt variants, and similar patterns are observed among Llama 3 and Llama 4 models. In several cases, these model-model correlations exceed the correlations observed between individual models and human raters, suggesting that some LLMs share similar implicit grading strategies even when those strategies diverge from human scoring behavior.  We also observe that prompt variants of the same model (base, student-grade, and few-shot) can exhibit particularly high agreement with each other across datasets and metrics. This indicates that while prompt modifications may affect the absolute scores assigned by a model, they could generally preserve the relative ordering of essays. In contrast, GPT-3.5 consistently shows weaker agreement with most other models, indicating a noticeably different scoring behavior.

\begin{observation}
Models within the same family seem to share implicit grading strategies that can exceed their agreement with human raters, while prompt modifications tend to preserve the relative ordering of essays rather than changing absolute scores.
\end{observation}

\begin{figure}[!ht]
\centering
\includegraphics[width=\textwidth]{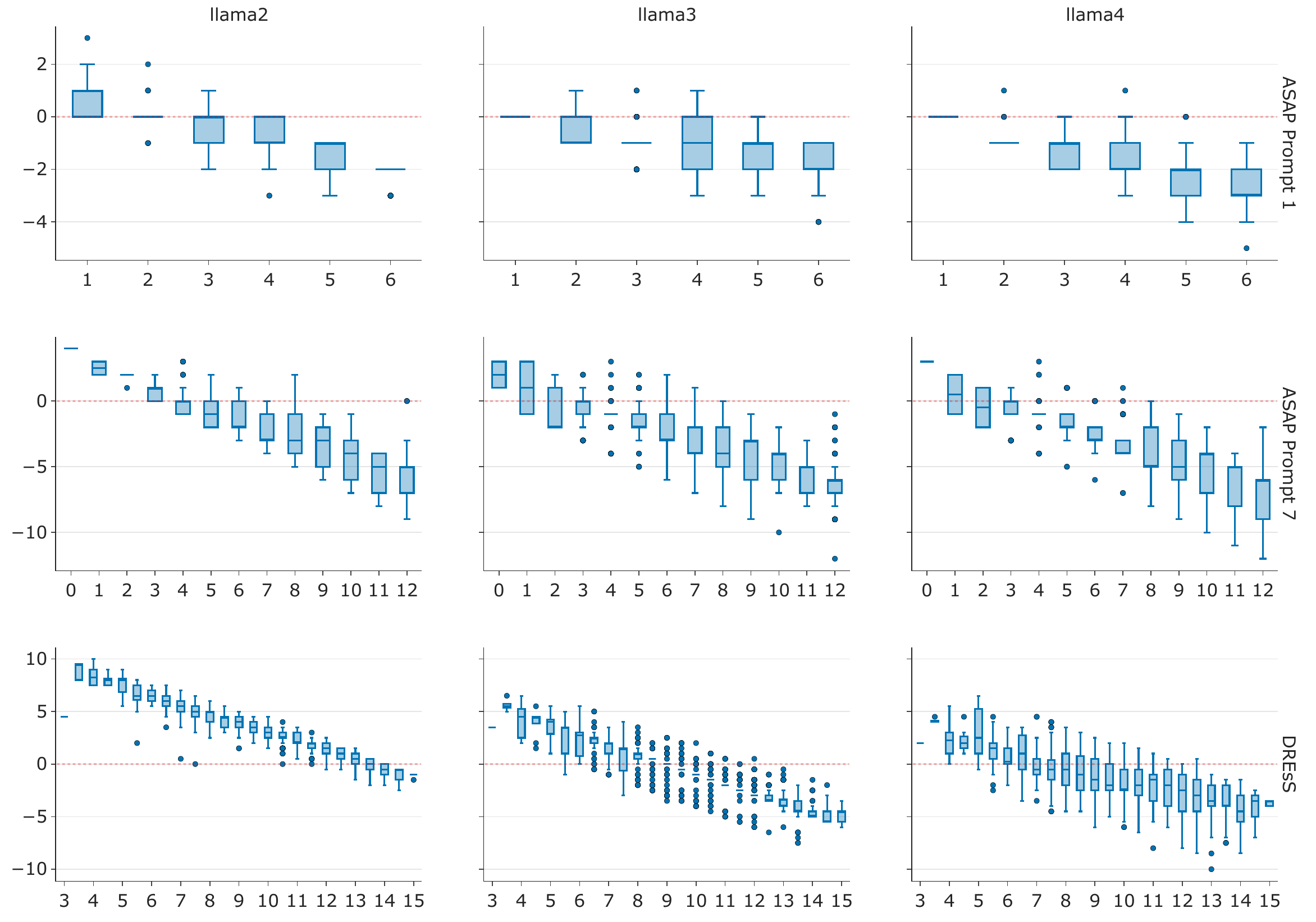}
\caption{Differences between Llama-assigned scores and human ratings grouped by human score across datasets. Columns correspond to Llama models and rows to datasets. Positive values indicate higher model scores than human ratings. Llama models assign relatively higher scores to low-scoring essays and lower scores to high-scoring essays, indicating compression of the score range.}\label{fig:llama_human_diff}
\end{figure}

\begin{figure}[!ht]
\centering
\includegraphics[width=\textwidth]{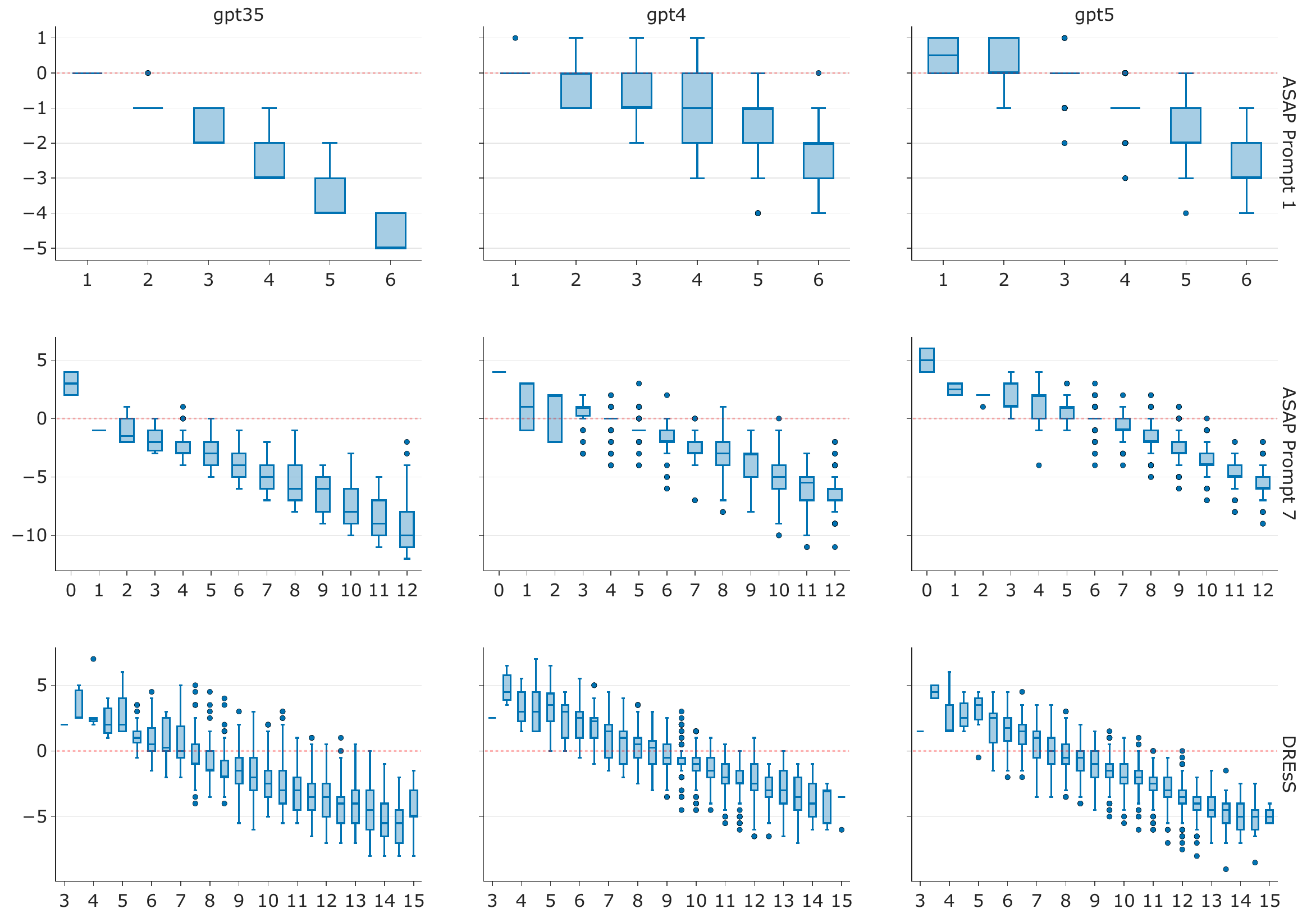}
\caption{Differences between GPT-assigned scores and human ratings grouped by human score across datasets. Columns correspond to GPT models and rows to datasets. Positive values indicate higher model scores than human ratings. Similar to Llama models, GPT models assign higher scores to low-scoring essays and lower scores to high-scoring essays, indicating compression of the score range.}\label{fig:gpt_human_diff}
\end{figure}

\subsection*{RQ1: LLMs compress the essay scoring range compared to human raters}

To better understand the nature of the disagreement between LLMs and human raters, we analyze how score differences vary across essays with different human-assigned scores. For each essay, we compute the difference between the LLM-assigned score and the human-assigned score\footnote{We report on scores from the first human rater for the ASAP datasets We observe similar results when using the second rater.}. A positive value indicates that the LLM assigns a higher score than the human rater, while a negative value indicates that the LLM assigns a lower score. We group these differences by the human-assigned score and report the results in Figures~\ref{fig:llama_human_diff} and~\ref{fig:gpt_human_diff}.

Across datasets and models, we observe a clear asymmetric pattern in the score differences. In general, LLMs tend to avoid assigning extreme scores, compressing the scoring range. As a result, LLMs often assign higher scores than humans to low-scoring essays and lower scores than humans to high-scoring essays. This pattern follows an approximately linear trend: score differences are positive for essays with low human scores, decrease as human scores increase, and become negative for essays with high human scores.

Some differences across datasets and models can still be observed. On ASAP Task 1, for example, Llama 2 tends to assign higher scores than human raters for very low-scoring essays ($score \leq 1$), while Llama 3 and Llama 4 are more closely aligned with human scores in this range. However, for essays with higher human scores, these two models generally assign slightly lower scores than human raters, as indicated by the negative score differences observed across most score levels. Among the GPT models, GPT-3.5 and GPT-4 are relatively aligned with human raters for the lowest scores, while GPT-5 shows larger positive differences in this range.

On ASAP Task 7, both Llama and GPT models generally assign slightly lower scores than human raters, as most score differences fall below zero. In contrast, on the DREsS dataset, Llama 2 tends to assign higher scores than humans across most score levels, whereas Llama 3, Llama 4, and the GPT models show a more balanced pattern, with positive differences for low-scoring essays and negative differences for high-scoring essays.

\begin{observation}
LLMs systematically compress the scoring range, assigning higher scores than humans to low-scoring essays and lower scores to high-scoring ones, following an approximately linear trend that varies across models and datasets.
\end{observation}

To better understand these LLM-human disagreements, we manually inspected essays with large absolute LLM-human score differences. For low-scoring essays, we observed that many responses were readable but short and underdeveloped, offering limited justification for their claims and containing grammatical or spelling errors. While human raters assigned very low scores to these essays, LLMs often assigned slightly higher scores, frequently noting that the essay addressed the prompt even if the argument was weak. An example of such a case is shown in Figure~\ref{fig:llm_human_disagreement_lowest_score}.

\begin{figure}[!ht]
\centering
\includegraphics[width=\textwidth]{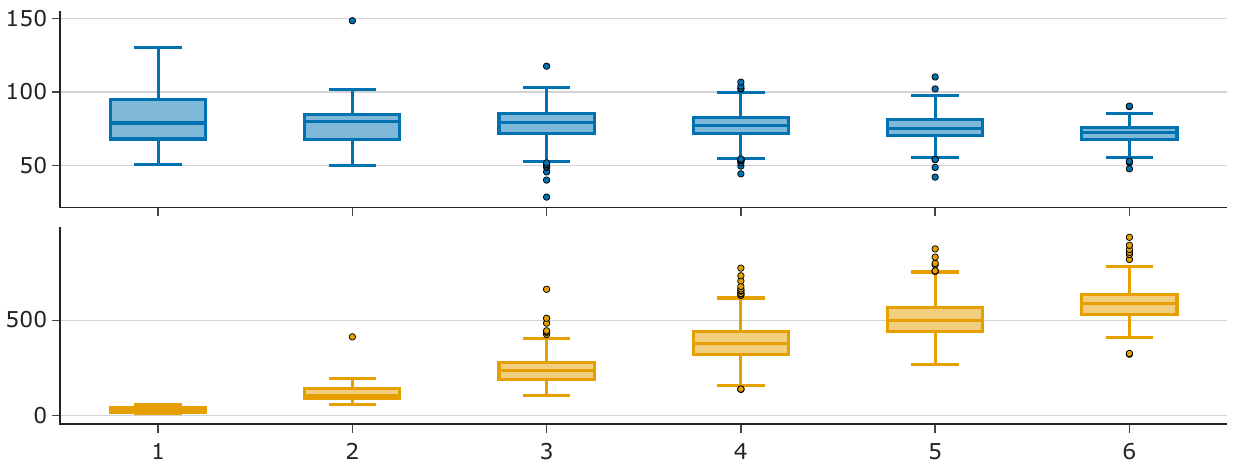}
\caption{Readability scores (Flesch Reading Ease, top) and essay lengths (number of tokens, bottom) for the ASAP Task 1 dataset grouped by human score. Essays receiving higher scores tend to be substantially longer, while readability varies less across score levels.}
\label{fig:readability_length_asap1}
\end{figure}

Interestingly, analysis of essay characteristics (Figures~\ref{fig:readability_length_asap1}, \ref{fig:readability_length_asap7} and \ref{fig:readability_length_dress}) shows that readability remains relatively stable across score levels, while essay length increases substantially with human scores. This suggests that some low-scoring essays may still be easy to read despite being brief or underdeveloped, which may lead LLMs to assign slightly higher scores. In contrast, human raters may place greater emphasis on essay development and length, which are correlated with higher human-assigned scores.

\begin{figure}[!ht]
\centering
\includegraphics[width=\textwidth]{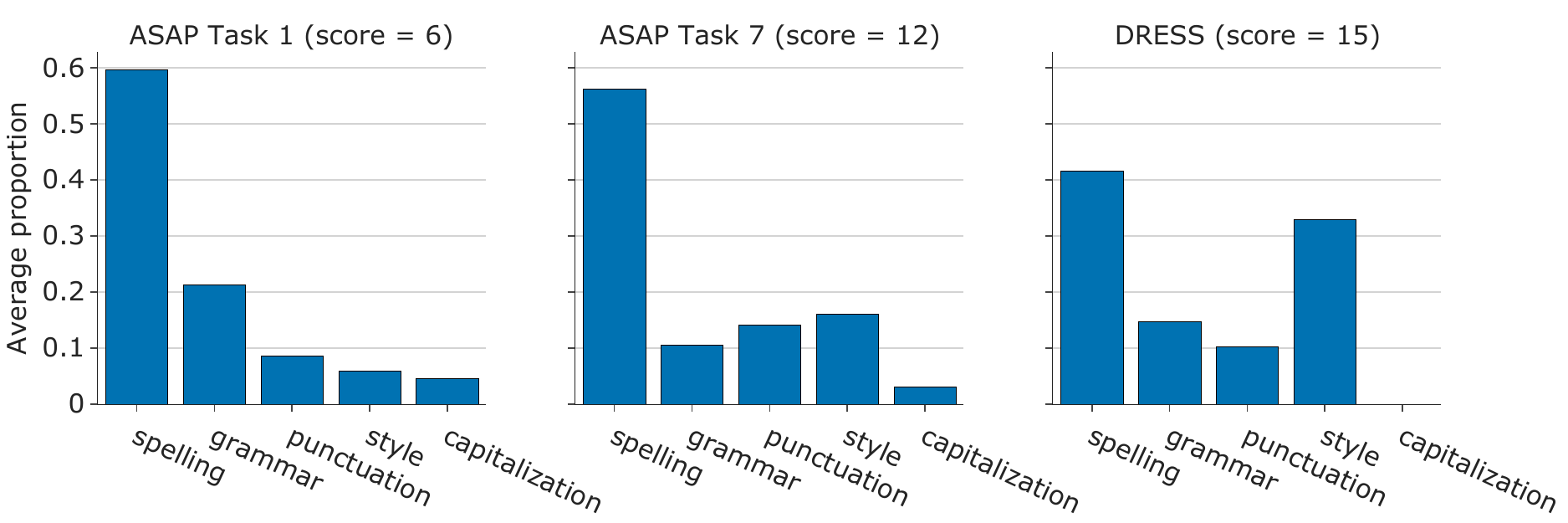}  
\caption{Average distribution of LanguageTool-detected error types in essays that received the maximum possible score from human raters. All such essays contain at least one detected error, most commonly spelling mistakes, suggesting that minor language issues are often tolerated by human raters.}\label{fig:language_tool_stats}
\end{figure}

At the other extreme, we observed that essays receiving the maximum human score often still contain minor grammatical or syntactic errors (e.g. punctuation or capitalization). Figure~\ref{fig:language_tool_stats} shows the average distribution of LanguageTool-detected error types within essays that received the maximum score. All of these essays are flagged for errors, most commonly spelling mistakes. Although these errors typically do not affect the overall comprehension of the text and are overlooked by the humans, the LLMs often consistently score the essays lower than humans and highlight such issues in their feedback. An example of this type of disagreement is shown in Figure~\ref{fig:llm_human_disagreement_highest_score}. These observations suggest that human raters may be more tolerant of minor language errors when the essay content and argumentation are well-developed\footnote{This hypothesis, however, cannot be studied with the datasets available today.}.

\begin{observation}
Compared to humans, LLMs tend to overrate short or underdeveloped essays that address the prompt, and to over-penalize minor language errors in otherwise well-developed essays, suggesting they take a more literal approach to grading according to humans.
\end{observation}

\begin{figure}[!ht]
\centering
\begin{subfigure}{\textwidth}
    \centering
    \includegraphics[width=\textwidth]{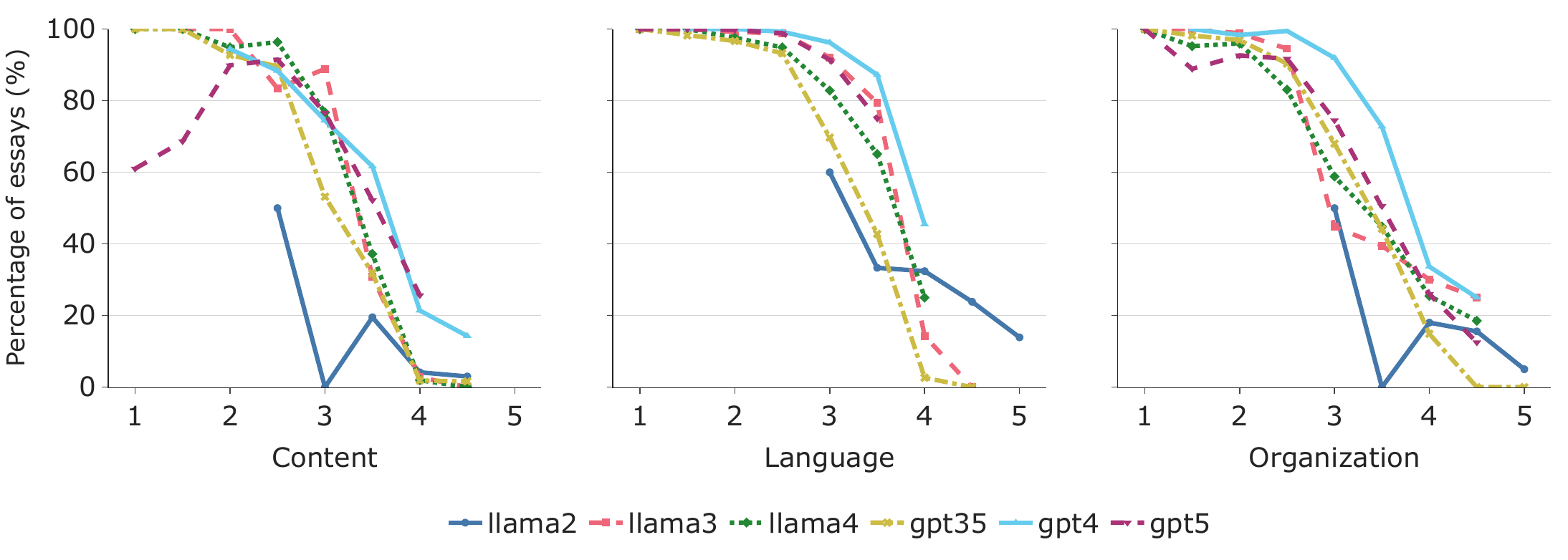}
    \caption{Criticism}
\end{subfigure}
\vspace{0.5cm}
\begin{subfigure}{\textwidth}
    \centering
    \includegraphics[width=\textwidth]{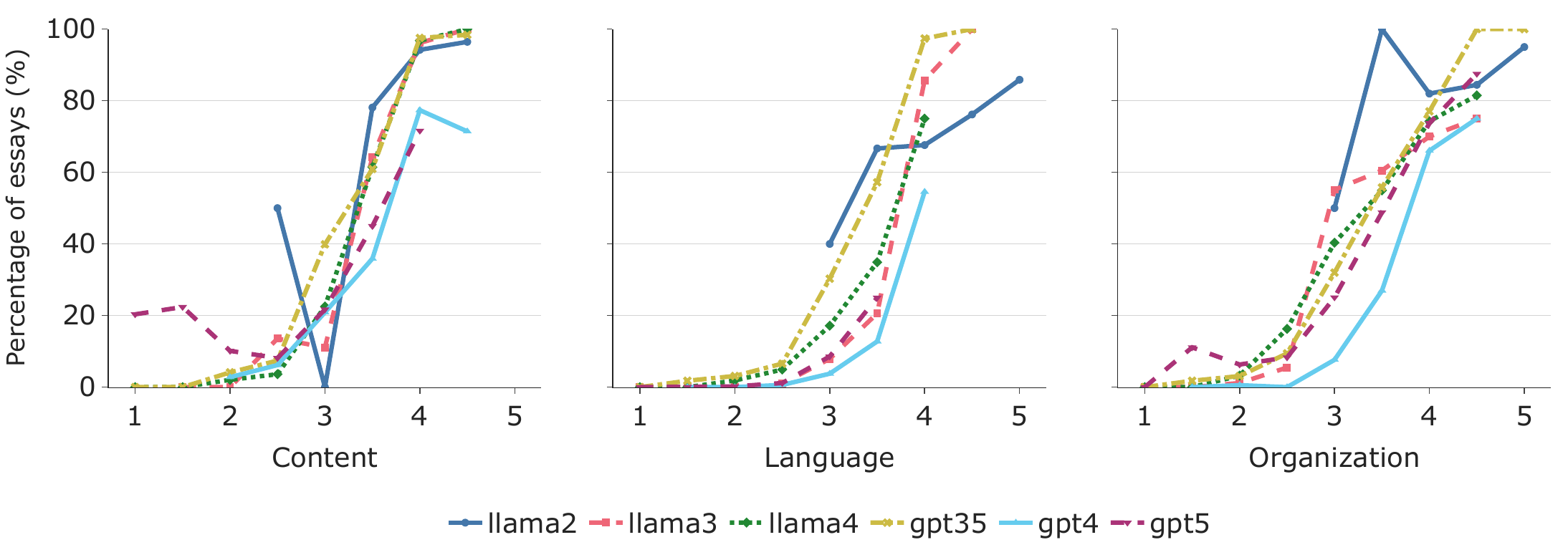}
    \caption{Praise}
\end{subfigure}
\caption{Proportion of praise (bottom) and criticism (top) associated with different rubric traits across essays with different LLM-assigned scores on the DREsS dataset. As scores increase, praise becomes more frequent while criticism decreases across most traits, indicating that feedback sentiment generally corresponds to the assigned scores.
}
\label{fig:absa_trend_dress}
\end{figure}

\begin{figure}[!ht]
\centering
\begin{subfigure}{\textwidth}
    \centering
    \includegraphics[width=\textwidth]{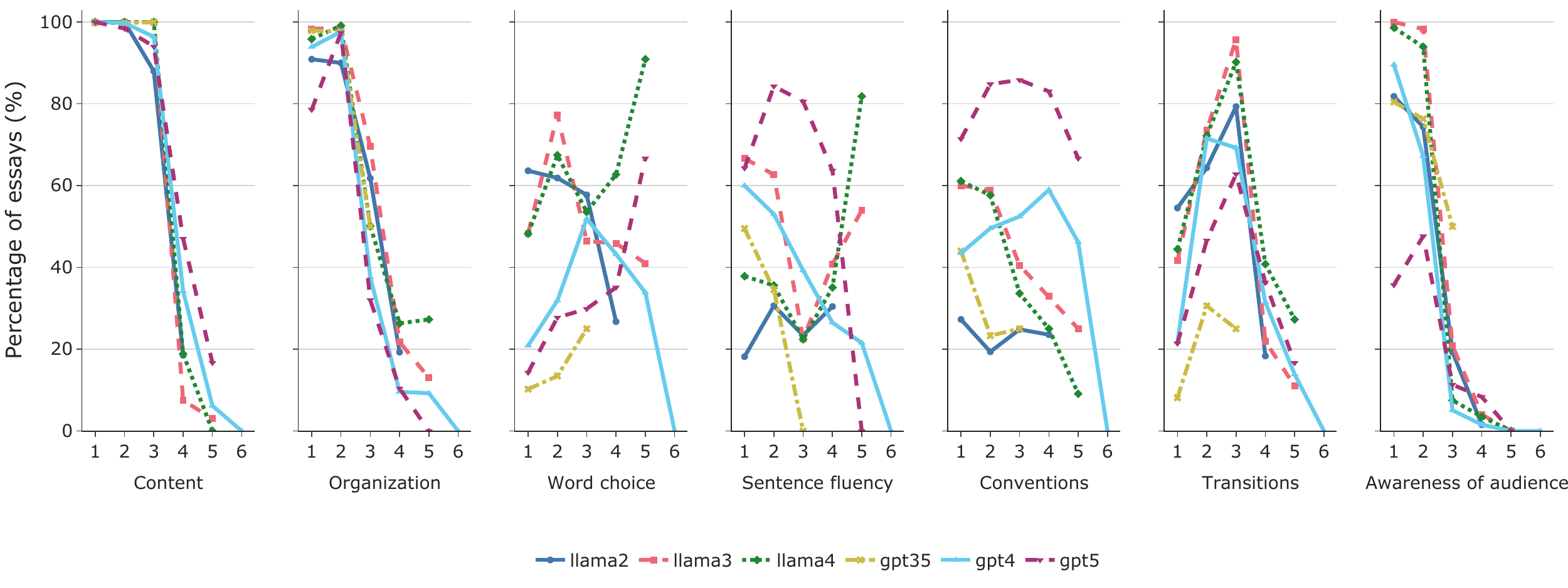}
    \caption{Negative sentiment}
\end{subfigure}
\vspace{0.5cm}
\begin{subfigure}{\textwidth}
    \centering
    \includegraphics[width=\textwidth]{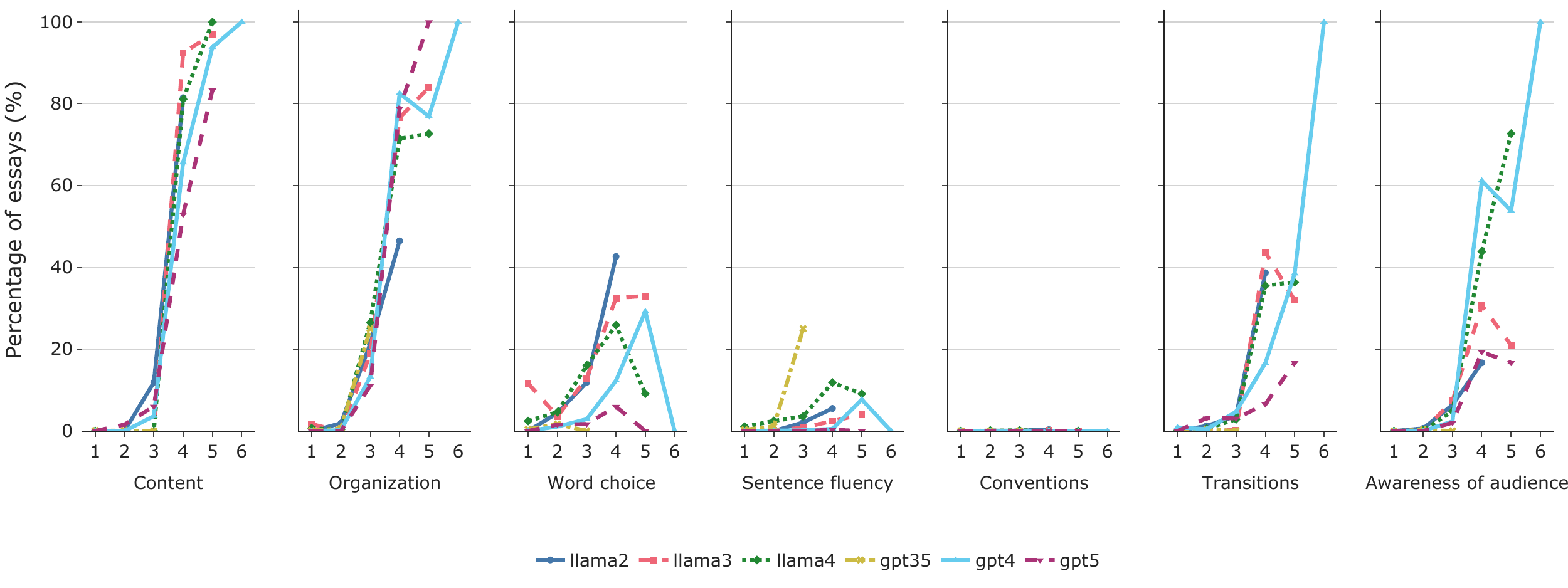}
    \caption{Positive sentiment}
\end{subfigure}
\caption{Proportion of criticism (top) and praise (bottom) associated with rubric traits across essays with different LLM-assigned scores on ASAP Task 1. Clear score-dependent trends are observed for traits such as content, organization, and awareness of audience, while other traits show more irregular patterns across score levels.}
\label{fig:absa_trend_asap_1}
\end{figure}

\subsection*{RQ2: LLM-assigned scores are consistent with the sentiment of their feedback}

To examine whether the feedback generated by the models reflects the scores they assign, we analyze the sentiment of trait-specific feedback using the ABSA model introduced in Section~\ref{subsec:implementation_details}. Because LLM feedback can contain both positive and negative statements about the same trait, and to mitigate potential prediction errors from the ABSA model, we aggregate sentiment in two steps. First, we aggregate sentiment at the level of individual trait-specific terms using majority voting. Then, we compute a second majority vote across terms belonging to the same rubric trait.

For each model, essays are grouped by their assigned scores, and we report the proportion of essays with positive and negative aggregated sentiment for each trait. Results for the DREsS and ASAP Task 1 datasets are shown in Figures~\ref{fig:absa_trend_dress} and~\ref{fig:absa_trend_asap_1} respectively. Some score levels are missing for certain models because no essays were assigned those scores.

Across datasets and models, we observe a clear pattern between feedback sentiment and assigned scores: essays receiving low scores are typically associated with a high proportion of negative mentions across traits, while essays receiving higher scores contain increasingly positive feedback. This trend is particularly clear in the DREsS dataset, where the proportion of essays with negative sentiment generally decreases as scores increase, often reaching near zero at the highest score levels. Conversely, the proportion of positive mentions increases with the assigned score, frequently exceeding 80-100\% for high-scoring essays. This trend appears consistently across both Llama and GPT models. Some fluctuations across score levels can also be observed. These may partly reflect noise introduced by the ABSA model or the fact that the manually defined trait-specific terms do not cover all expressions used in the feedback.

However, we also observe that this pattern is not equally strong across datasets and traits. On ASAP Task 1 (Figure~\ref{fig:absa_trend_asap_1}), clear score-dependent trends are observed for traits such as \textit{content}, \textit{organization}, and \textit{awareness of audience}. For other traits, including \textit{word choice}, \textit{sentence fluency}, and \textit{conventions}, the relationship between sentiment and score is less evident, with more irregular patterns across score levels.

\begin{observation}
LLM feedback sentiment consistently tracks assigned scores — low scores correlate with more criticism and high scores with more praise — though the strength of this relationship varies across traits and datasets.
\end{observation}

While the previous analysis shows that feedback sentiment correlates with the assigned scores, it does not reveal how these traits contribute to the scoring decisions. To better understand the relative influence of different traits, we train a proxy model to predict LLM-generated scores using the counts of positive and negative mentions for each rubric trait as features. Specifically, for each dataset and each LLM, we train an XGBoost model to predict the overall score assigned by the LLM. Each feature corresponds to the number of positive or negative mentions associated with a specific trait. We then apply SHAP~\citep{lundberg2017unified} to estimate the contribution of each feature to the predicted score.

\begin{figure}[!t]
\centering

\begin{subfigure}{0.32\textwidth}
    \centering
    \includegraphics[width=\textwidth]{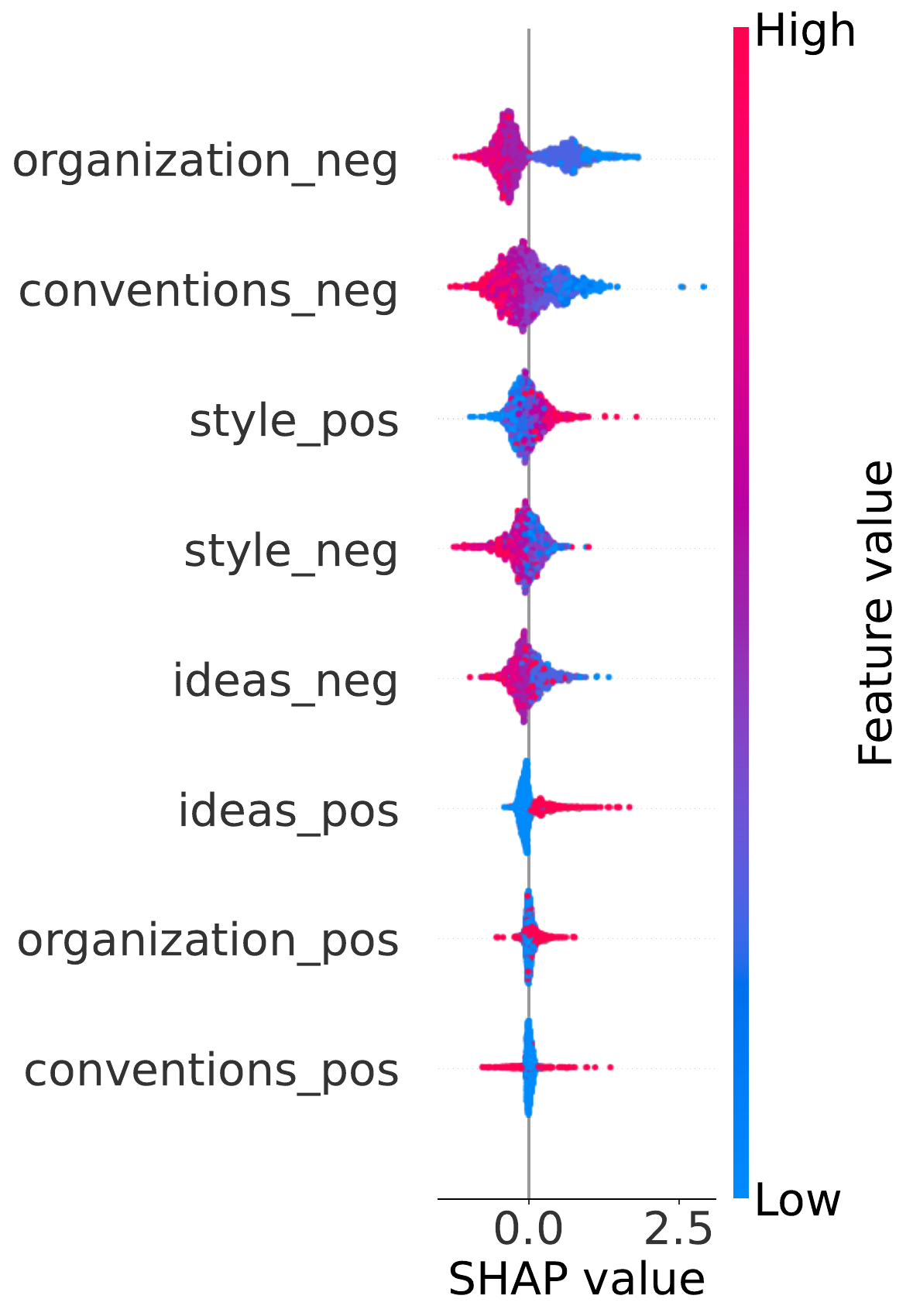}
    \caption{Llama2}
\end{subfigure}
\hfill
\begin{subfigure}{0.32\textwidth}
    \centering
    \includegraphics[width=\textwidth]{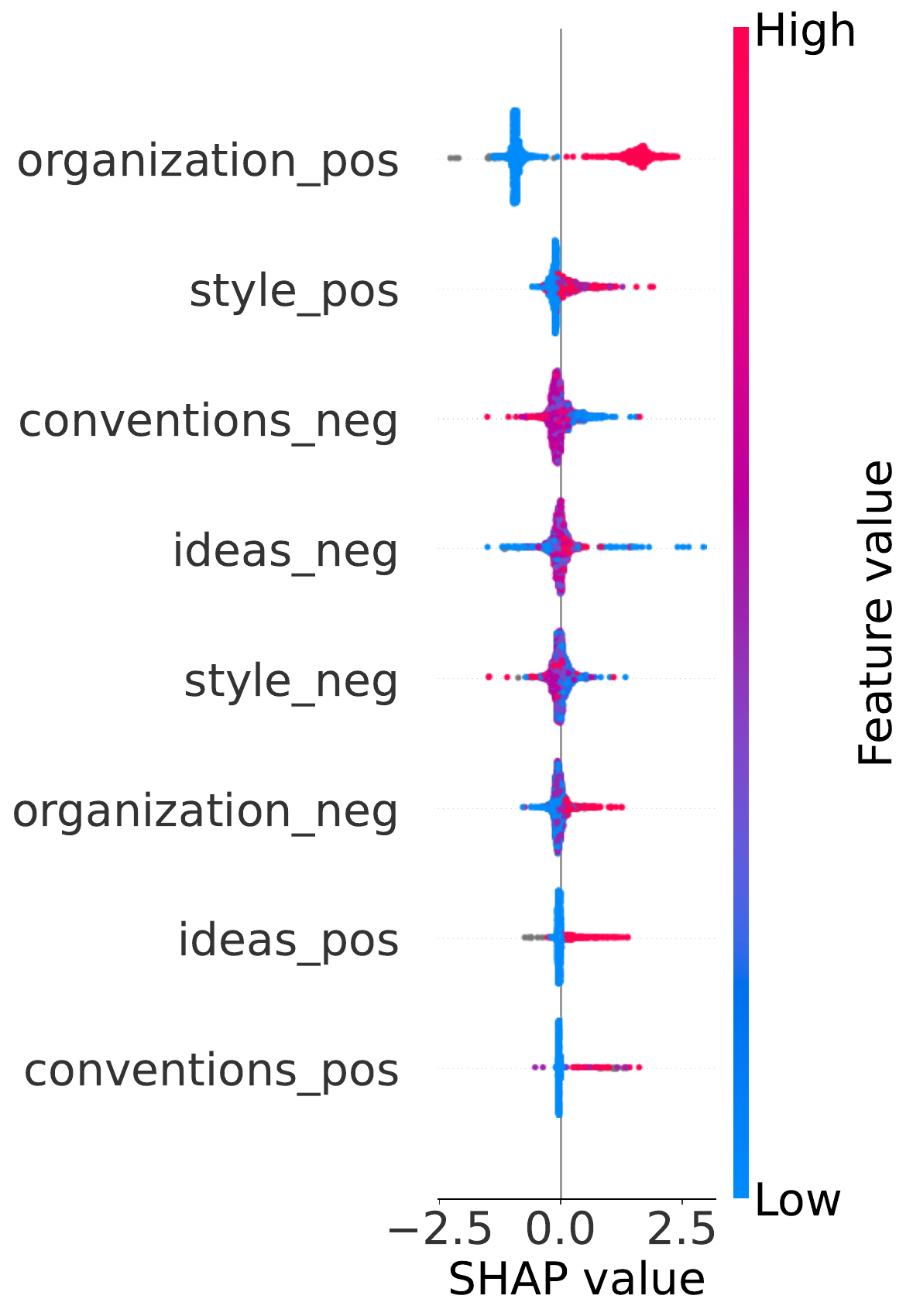}
    \caption{Llama3}
\end{subfigure}
\hfill
\begin{subfigure}{0.32\textwidth}
    \centering
    \includegraphics[width=\textwidth]{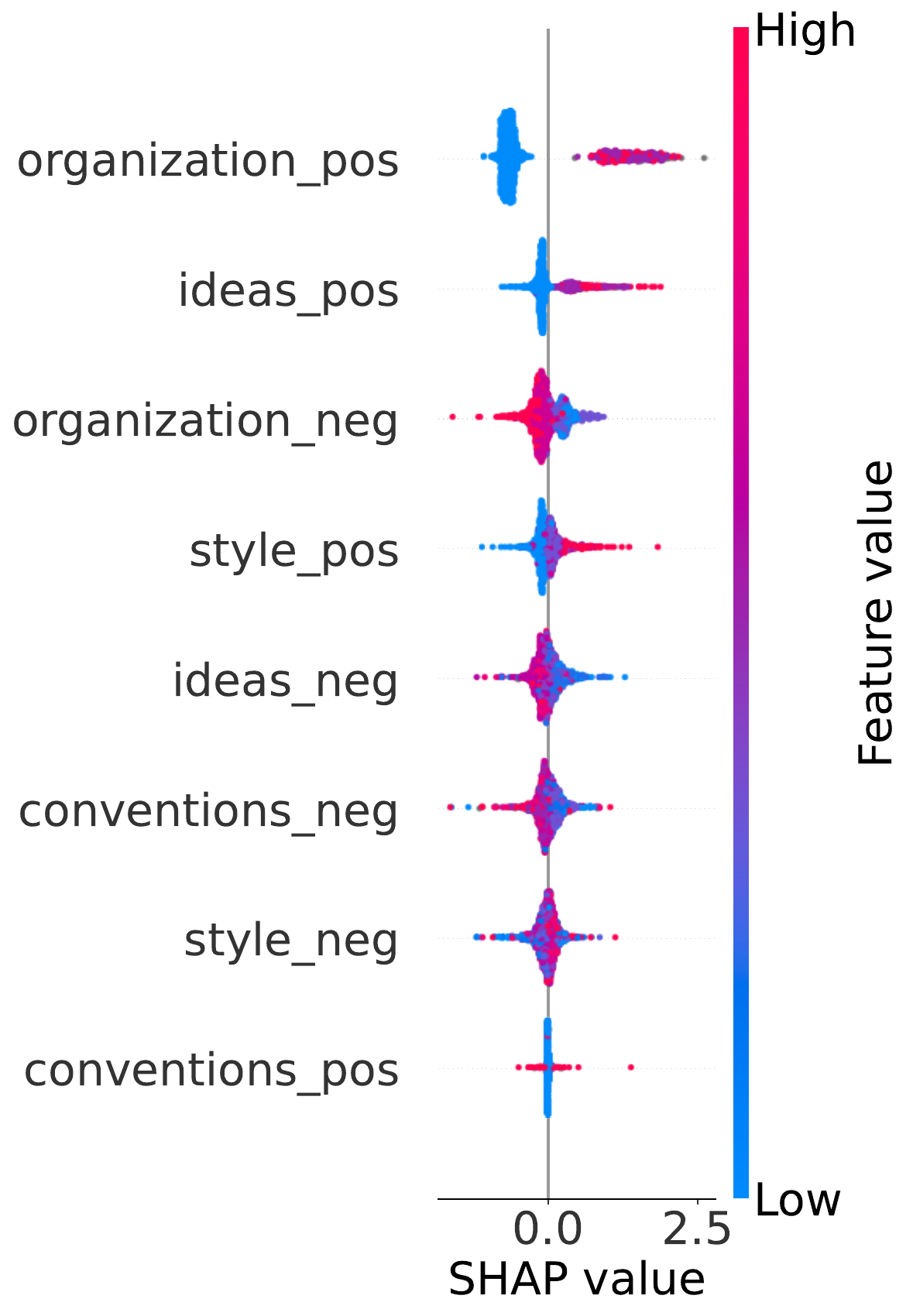}
    \caption{Llama4}
\end{subfigure}

\caption{SHAP beeswarm plots showing the contribution of trait-specific praise and criticism mentions to predicted essay scores on ASAP Task 7. Features sorted in decreasing order of importance (mean absolute SHAP value). Points represent individual essays, and horizontal position indicates the impact on the predicted score (SHAP value). Praise-related features generally increase predicted scores, while criticism-related features decrease them.}
\label{fig:beeswarm_llama_asap7}
\end{figure}

Figure~\ref{fig:beeswarm_llama_asap7} shows the resulting SHAP beeswarm plots for the Llama models on ASAP Task~7. Each point represents an essay, with red points indicating higher feature values (i.e., more mentions of a given trait) and blue points indicating lower feature values. For praise-related features (e.g., $organization_{pos}$), higher feature values correspond to positive SHAP values, increasing the predicted score. In contrast, criticism-related features (e.g., $organization_{neg}$) tend to produce negative SHAP values, lowering the predicted score. 

These results suggest that LLM-assigned scores are coherent with the evaluative signals expressed in their feedback: praise-related mentions increase predicted scores, whereas criticism-related mentions decrease them. Similar trends are observed for GPT models and across datasets (\ref{sec:shap_analysis}).

\begin{observation}
SHAP analysis suggests that praise-related mentions in the feedback increase LLM-assigned scores while criticism-related mentions decrease them, indicating that LLM scores are coherent with the evaluative signals in their feedback.
\end{observation}

\subsection*{RQ2: LLMs rely on different set of traits when assigning grades} 

The previous analysis showed that LLM-assigned scores are generally consistent with the sentiment expressed in the generated feedback. We now examine whether all rubric traits contribute equally to the scoring decision, or whether models tend to emphasize specific aspects of writing quality.

\begin{figure}[!ht]
\centering
\includegraphics[width=\textwidth]{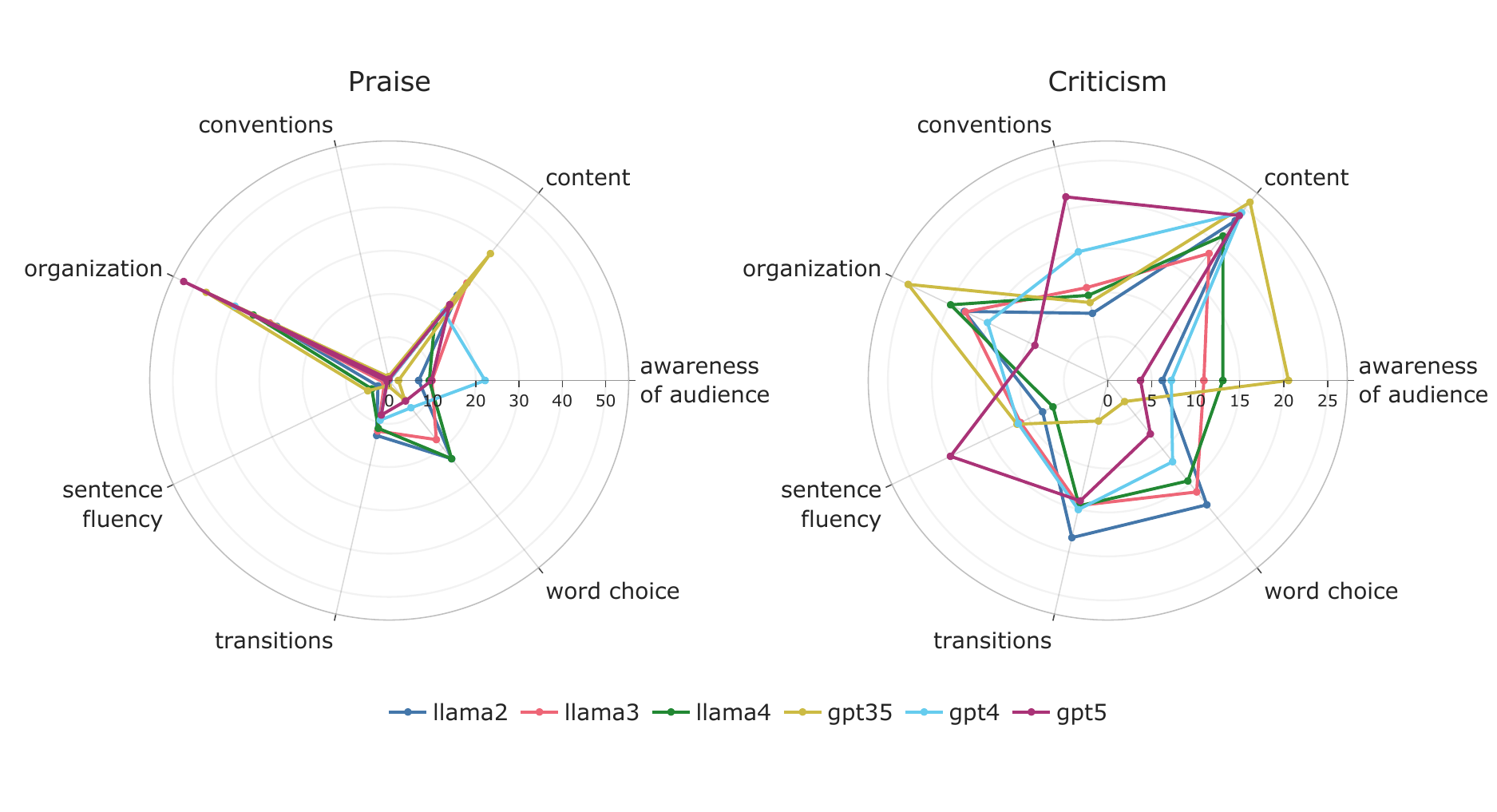}
\caption{Distribution of trait-specific praise (left) and criticism (right) across models for essays in ASAP Task 1. Praise is concentrated on a small number of traits, particularly organization and content, while criticism is distributed more broadly across traits.}
\label{fig:spider_asap1}
\end{figure}

\begin{figure}[!ht]
\centering
\includegraphics[width=\textwidth]{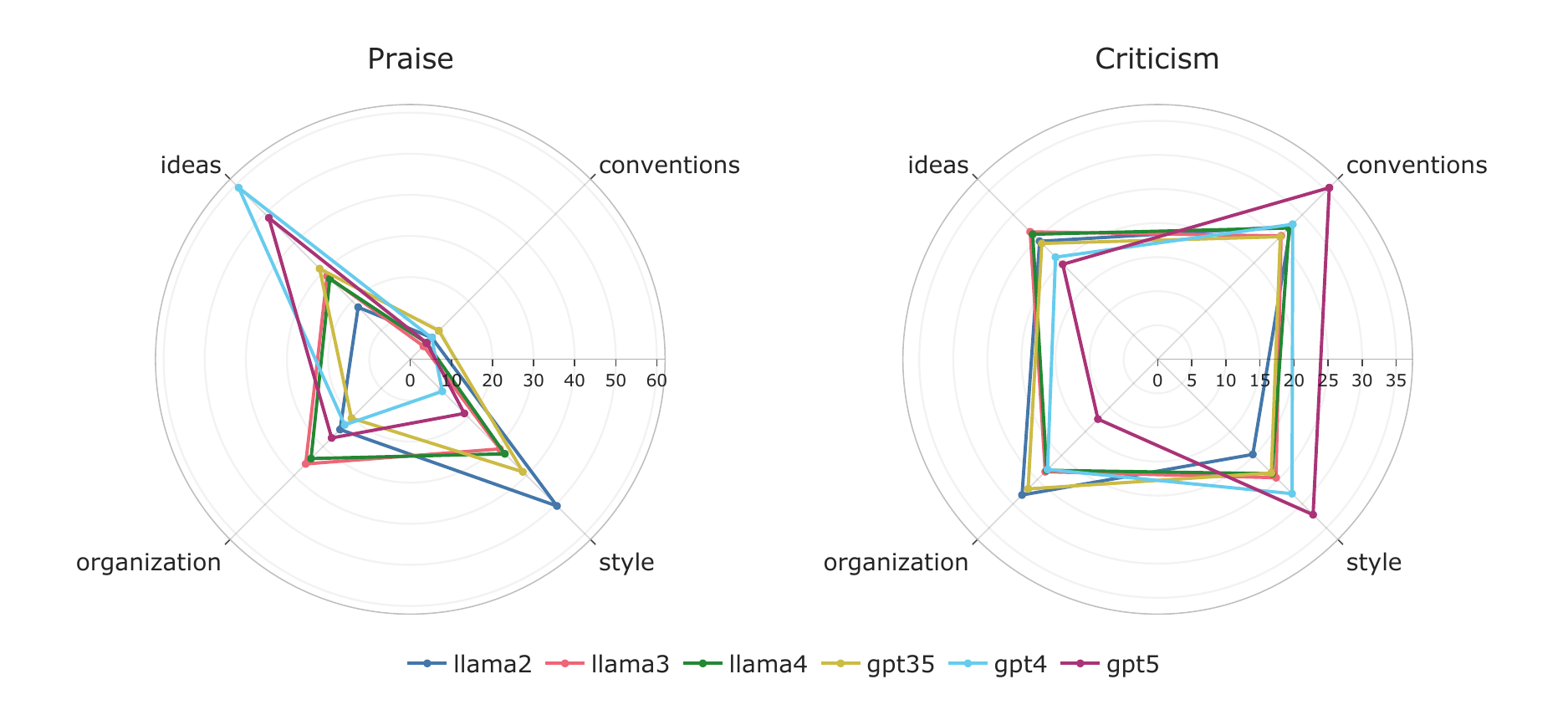}
\caption{Distribution of trait-specific praise (left) and criticism (right) across models for essays in ASAP Task 7. Positive mentions are concentrated mainly on \textit{ideas}, with some models also placing substantial emphasis on \textit{style}, whereas negative mentions are distributed more broadly across the rubric traits.}
\label{fig:spider_asap7}
\end{figure}

\begin{figure}[!ht]
\centering
\includegraphics[width=\textwidth]{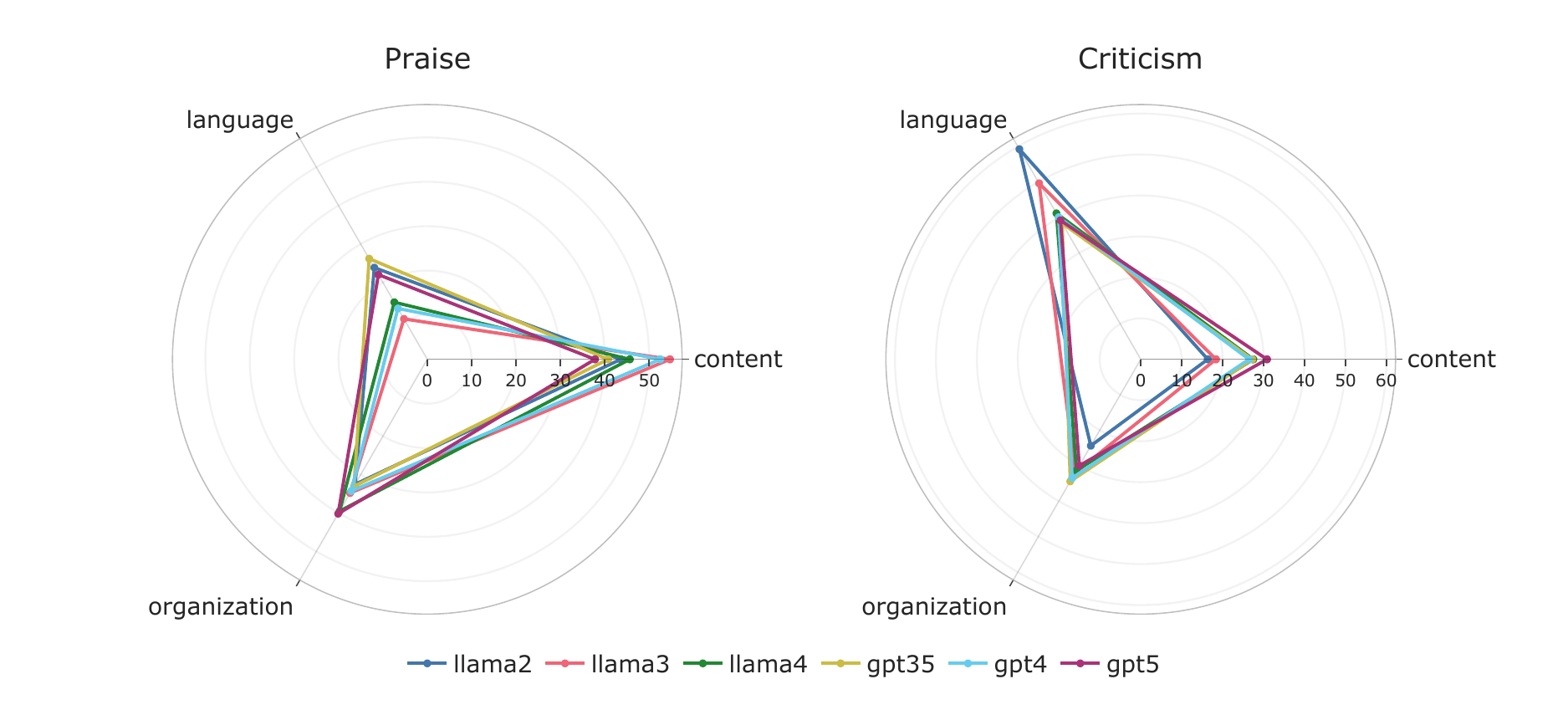}
\caption{Distribution of trait-specific praise (left) and criticism (right) across models for essays in the DREsS dataset. Trait mentions are uneven: praise is most frequently associated with \textit{content}, while criticism places greater emphasis on \textit{language}, with \textit{organization} playing a secondary role.}
\label{fig:spider_dress}
\end{figure}

To investigate this question, we first analyze how trait-specific praise and criticism mentions are distributed across essays. For each dataset and model, we collect all sentence-level mentions of positive and negative feedback terms, group them by their corresponding rubric trait, and compute their relative distribution. The resulting distributions are shown in Figures~\ref{fig:spider_asap1}, \ref{fig:spider_asap7}, and \ref{fig:spider_dress} for ASAP Task 1, ASAP Task 7, and DREsS respectively.

Across datasets and models, trait mentions are clearly not uniformly distributed. Instead, models tend to emphasize certain traits more than others when generating feedback. For example, in the ASAP Task 7 dataset, while GPT-4 focuses strongly on \textit{ideas} (around 60\%), Llama 2 produces substantially more praise related to \textit{style} (around 50\%) and fewer mentions of \textit{ideas}. Notably, praise related to \textit{conventions} does not exceed 10\% for any model.

We also observe differences in how praise and criticism are distributed across traits. In the ASAP datasets, praise mentions tend to concentrate on certain traits, whereas criticism is more evenly distributed across multiple aspects of writing. For example, in ASAP Task 7, most of the praise is related to the \textit{ideas} and \textit{style} traits, while criticism is spread across the other traits as well, such as \textit{organization} and \textit{conventions}. A similar pattern can be observed in ASAP Task 1, where praise is primarily associated with the \textit{organization} trait, while criticism spans a broader set of traits. In the DREsS dataset, trait mentions remain uneven: praise focuses primarily on \textit{content}, with \textit{organization} playing a secondary role and \textit{language} appearing less frequently, while criticism places more emphasis on \textit{language}, with \textit{content} and \textit{organization} receiving lower and more similar proportions.  This suggests that the traits emphasized in model feedback depend not only on the model but also on the dataset and the rubric structure.

We observe that variation in trait distributions across models is particularly pronounced in ASAP Task 1 compared to ASAP Task 7 and DREsS. One possible explanation is that the rubric for ASAP Task 1 does not explicitly define the trait categories used in our analysis, which are derived from the ASAP++ dataset. Because these traits are not directly referenced in the original rubric, models may rely on different linguistic cues when generating feedback, resulting in greater variability in trait-specific mentions.

\begin{observation}
LLMs do not evaluate rubric traits uniformly - praise tends to concentrate on a few traits while criticism is more broadly distributed, with trait emphasis varying across models and rubric structures.
\end{observation}

\begin{figure}[!ht]
\centering

\begin{subfigure}{0.32\textwidth}
    \centering
    \includegraphics[width=\textwidth]{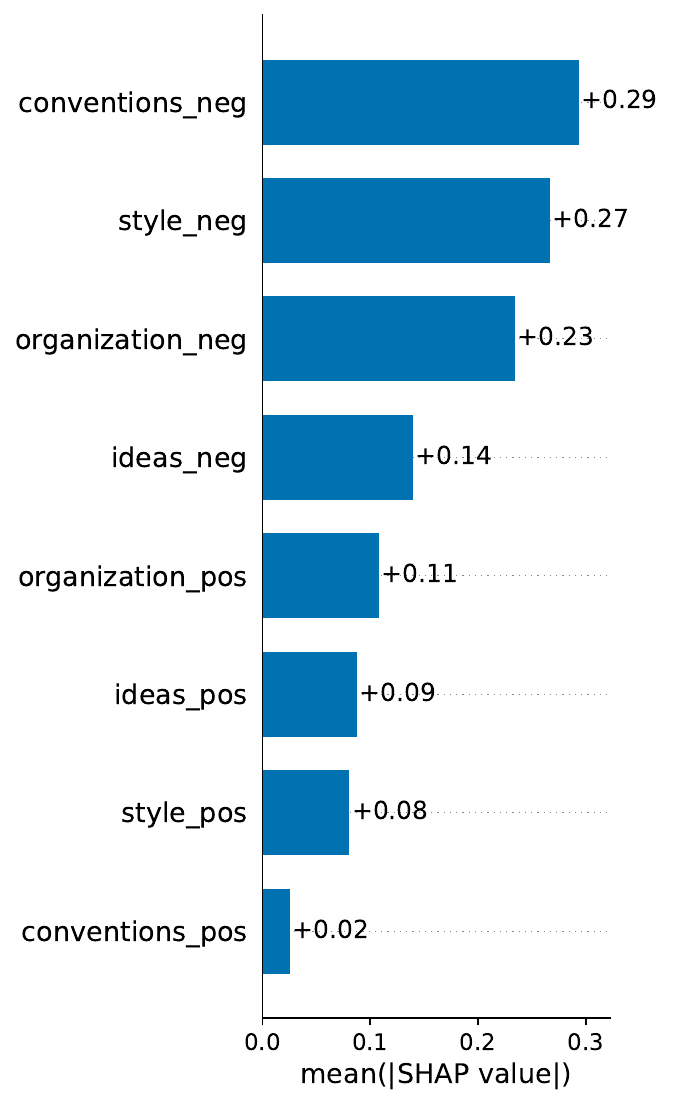}
    \caption{GPT-3.5}
\end{subfigure}
\hfill
\begin{subfigure}{0.32\textwidth}
    \centering
    \includegraphics[width=\textwidth]{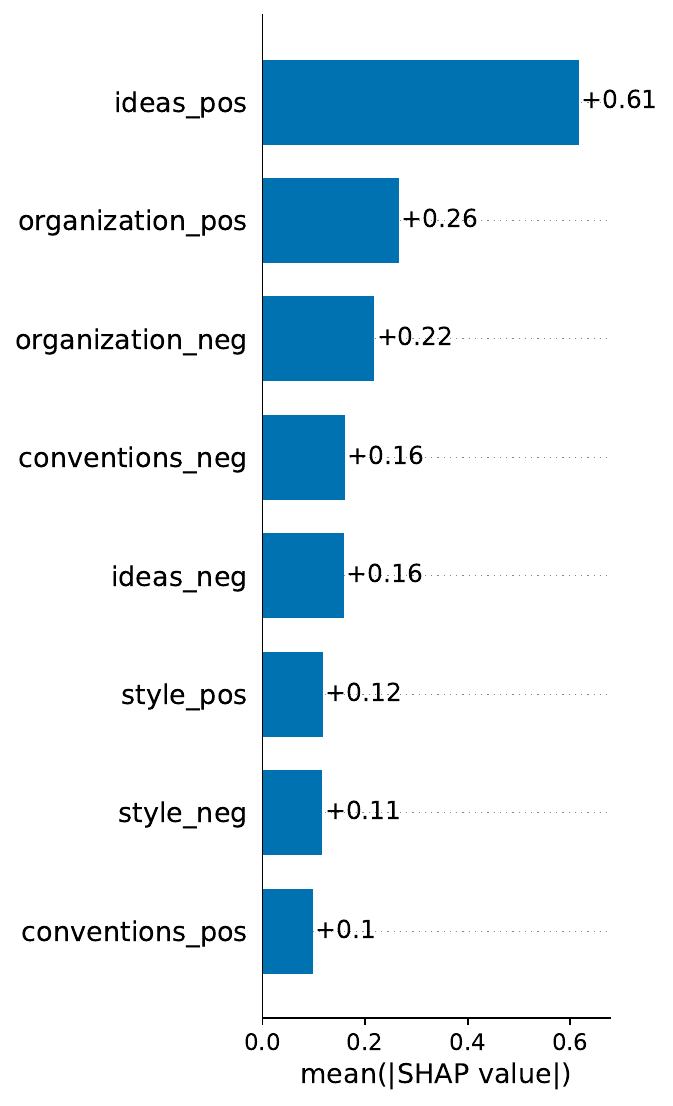}
    \caption{GPT-4}
\end{subfigure}
\hfill
\begin{subfigure}{0.32\textwidth}
    \centering
    \includegraphics[width=\textwidth]{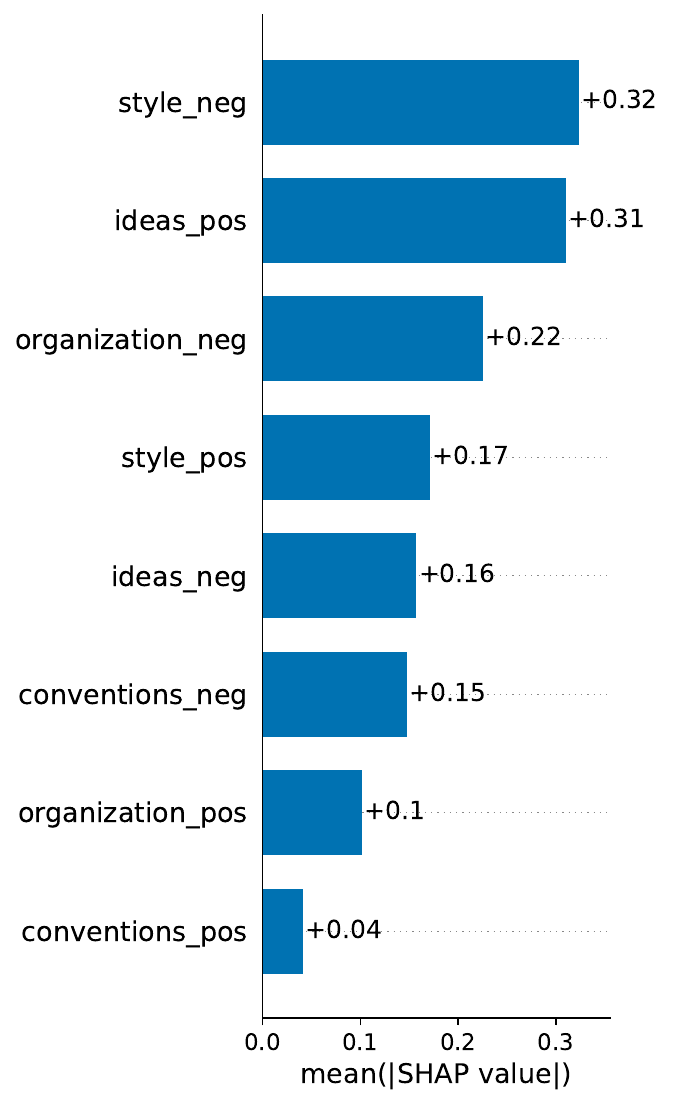}
    \caption{GPT-5}
\end{subfigure}

\caption{Average absolute SHAP values for trait-specific praise and criticism features for GPT models on ASAP Task 7. Higher values indicate stronger influence on predicted scores, highlighting differences in trait importance across models.}
\label{fig:shap_barplot_gpt_asap_7}
\end{figure}

While the previous analysis reveals which traits are most frequently mentioned in feedback, it does not directly indicate which traits influence the final score. To address this question, we analyze the contribution of trait-specific mentions using SHAP values computed from the proxy XGBoost model introduced earlier. Figure~\ref{fig:shap_barplot_gpt_asap_7} reports the average absolute SHAP values for the GPT models on the ASAP Task 7 dataset.

The results reveal clear differences in how models weigh rubric traits when assigning scores. For GPT-3.5, the most influential features correspond to negative mentions of \textit{conventions}, \textit{style}, and \textit{organization}, suggesting that the model relies strongly on criticism related to writing mechanics and structure. In contrast, GPT-4 shows a highly skewed importance distribution in which positive mentions of \textit{ideas} dominate the score prediction. GPT-5 exhibits a more balanced pattern, where both positive mentions of \textit{ideas} and negative mentions of \textit{style} and \textit{organization} contribute substantially to the predicted score. Differences are observed also for Llama models and across datasets (see~\ref{sec:shap_analysis}).

Overall, these results indicate that LLMs do not rely on a uniform set of evaluation criteria when assigning grades. Instead, different models emphasize different traits and feedback signals, which may partly explain the variability observed in their scoring behavior.

\begin{observation}
Different LLMs rely on different subsets of rubric traits when assigning scores: some are dominated by a single trait while others show more balanced weighting, which partly explains the variability in their grading behavior.
\end{observation}

\section{Discussion}
\label{sec:discussion}
In this section, we revisit our two research questions and summarize our main findings.

\paragraph{RQ1: Do LLMs grade similarly to humans?}
Our results, measured using QWK, Pearson correlation, and MAE, indicate that agreement between LLM-generated scores and human grades is generally weak. In addition, the level of agreement varies systematically with essay quality and the scores assigned by human raters. In particular, humans tend to assign lower scores than LLMs to short and poorly developed essays, even when those essays partially address the prompt. On the other hand, for longer and more developed essays, human raters often assign higher scores than LLMs despite the presence of minor grammatical or spelling errors. This pattern suggests that human graders may place greater emphasis on content development and overall argumentation, while tolerating small language mistakes that do not substantially affect the readability or meaning of the essay. In contrast, LLMs tend to explicitly highlight grammatical and spelling issues in their feedback and often penalize such errors more strongly in their assigned scores. As a result, essays that receive high scores from human raters may still receive lower scores from LLMs if the models identify language-related mistakes.

These findings suggest that LLM grading criteria differ from those implicitly used by human raters. These differences are also reflected in the distribution of scores assigned by different models. In particular, GPT-3.5 tends to assign lower scores more frequently than both human raters and other LLMs, resulting in a distribution that is skewed toward the lower end of the grading scale. In addition to that, more recent models exhibit somewhat different score distributions, but improvements in alignment with human raters are not consistent across model generations. While some newer models achieve modest gains on certain datasets, the overall agreement between LLM-generated scores and human grades remains substantially below human inter-rater agreement.

\paragraph{RQ2: Do LLMs exhibit consistent grading behavior across essays?}
Our analysis indicates that LLMs assign scores that are generally consistent with the feedback they generate. Using an ABSA model, we observe that essays receiving lower scores tend to be associated with a higher number of criticism-related mentions in the feedback, while essays receiving higher scores typically contain more praise-related mentions. Further analysis using SHAP supports this observation. Positive mentions related to rubric traits are consistently associated with positive SHAP values, indicating that they contribute to higher predicted scores, while criticism-related mentions are associated with negative SHAP values and correspond to lower predicted scores. These results suggest that LLM-generated feedback reflects the logic underlying the assigned scores and that models use praise and criticism signals in a coherent manner when grading essays.

\paragraph{Implications for AES}

Our findings highlight several implications for the use of LLMs in automated essay scoring systems. First, the systematic differences observed between LLM and human grading suggest that calibration strategies may help better align model predictions with human judgments. For example, simple post-processing heuristics could potentially adjust scores by accounting for factors such as essay length or the presence of minor language errors. Second, the observed alignment between generated feedback and assigned scores suggests that LLMs can provide coherent explanations for their evaluations. This property may make LLM-based systems useful as tools to support instructors by providing preliminary assessments and feedback that students can use to improve their writing. However, further research is needed to determine whether such feedback effectively helps students improve essay quality and learning outcomes.

\paragraph{Limitations}

While our analysis provides new insights into the behavior of LLMs for essay scoring, several limitations should be acknowledged.

First, we relied on a single prompt template for all models. Prompt design can influence LLM performance in automated essay scoring, as discussed by \citet{mansour2024can}. However, identifying a prompt that consistently improves performance across different models and essay types remains challenging.

Second, our analysis of feedback relied on an ABSA model to identify praise and criticism mentions in LLM-generated feedback. Although this approach enables large-scale analysis, the model may introduce classification errors. To mitigate this issue, we applied a confidence threshold and aggregated predictions through majority voting. Nevertheless, systematic biases in the ABSA model could still affect the extracted sentiment mentions.

Third, the extraction of trait-related terms relied on manually defined vocabularies derived from an initial inspection of feedback samples. These term lists may not capture all linguistic expressions used by the models to refer to rubric traits, which could lead to missed mentions in the analysis.

Finally, the SHAP analysis relies on a proxy model trained to predict LLM-assigned scores using derived features (i.e., counts of positive and negative trait-specific mentions). This approach assumes that the proxy model reasonably approximates the scoring behavior of the LLM and that the selected features capture the most relevant signals used during scoring. 
\section{Conclusions}
\label{sec:conclusions}

In this paper, we evaluated how large language models (LLMs) compare to human graders in essay scoring and analyzed their grading behavior using multiple existing and widely-used datasets for the task. Across several GPT and Llama models, we found that agreement between LLM-generated scores and human annotations is weak. In particular, disagreement varies systematically with essay characteristics: LLMs tend to assign higher scores to short or underdeveloped essays, while assigning lower scores to longer essays that contain minor grammatical or spelling errors. At the same time, LLM-generated scores are generally consistent with the feedback they produce, suggesting that models follow coherent internal criteria when evaluating essays. We also found that, while new generations of these models are touted to have more capabilities compared to their predecessors, the fluctuations in performance across model versions raise the question as to whether GPT and Llama became ``better'' over time.

These findings have implications for educational applications. For instructors, the consistency between scores and feedback suggests that LLMs could assist in identifying strengths and weaknesses in student essays and provide rubric-aligned preliminary assessments. For students, the feedback may be useful for identifying language-level issues such as grammatical or spelling errors. However, the systematic differences observed between LLM and human grading indicate that such systems should currently be used as support tools rather than fully automated graders.

\paragraph{Future work} Extensions to our work include evaluating the analysis on more essay datasets, studying how prompt variations influence grading behavior, and examining whether similar patterns hold across a wider range of LLM architectures. Future work should also investigate how instructors perceive and use LLM-generated evaluations in real grading settings and whether teachers agree with model-assigned scores and feedback. Another important direction is to evaluate whether feedback generated by LLMs leads to measurable improvements in student writing. 

%\input{sections/generative-ai-use}

%%       \citet{<label>} ==> Lamport (1994)
%%       \citep{<label>} ==> (Lamport, 1994)
%%

\newpage
\bibliographystyle{elsarticle-harv} 
\bibliography{main}


\newpage
\appendix

\section{Essay prompts and scoring rubrics}
\label{sec:dataset_prompts_rubrics}

\begin{tcolorbox}[title=ASAP Task 1, colback=gray!5, colframe=blue!30!black]

\scriptsize
\linespread{1}\selectfont

\medskip
\textbf{Prompt}  
More and more people use computers, but not everyone agrees that this benefits society. Those who support advances in technology believe that computers have a positive effect on people. They teach hand-eye coordination, give people the ability to learn about faraway places and people, and even allow people to talk online with other people. Others have different ideas. Some experts are concerned that people are spending too much time on their computers and less time exercising, enjoying nature, and interacting with family and friends. 

Write a letter to your local newspaper in which you state your opinion on the effects computers have on people. Persuade the readers to agree with you.
   
\medskip
\textbf{Task 1 Rubric} 

Score Point 1: An undeveloped response that may take a position but offers no more than very minimal support. Typical elements:
\begin{itemize}[nosep]
    \item Contains few or vague details.
    \item Is awkward and fragmented.
    \item May be difficult to read and understand.
    \item May show no awareness of audience.
\end{itemize}

Score Point 2: An under-developed response that may or may not take a position. Typical elements:
\begin{itemize}[nosep]
    \item Contains only general reasons with unelaborated and/or list-like details.
    \item Shows little or no evidence of organization.
    \item May be awkward and confused or simplistic.
    \item May show little awareness of audience.
\end{itemize}

Score Point 3: A minimally-developed response that may take a position, but with inadequate support and details. Typical elements:
\begin{itemize}[nosep]
    \item Has reasons with minimal elaboration and more general than specific details.
    \item Shows some organization.
    \item May be awkward in parts with few transitions.
    \item Shows some awareness of audience.
\end{itemize}

Score Point 4: A somewhat-developed response that takes a position and provides adequate support. Typical elements:
\begin{itemize}[nosep]
    \item Has adequately elaborated reasons with a mix of general and specific details.
    \item Shows satisfactory organization.
    \item May be somewhat fluent with some transitional language.
    \item Shows adequate awareness of audience.
\end{itemize}

Score Point 5: A developed response that takes a clear position and provides reasonably persuasive support. Typical elements:
\begin{itemize}[nosep]
    \item Has moderately well elaborated reasons with mostly specific details.
    \item Exhibits generally strong organization.
    \item May be moderately fluent with transitional language throughout.
    \item May show a consistent awareness of audience.
\end{itemize}

Score Point 6: A well-developed response that takes a clear and thoughtful position and provides persuasive support. Typical elements:
\begin{itemize}[nosep]
    \item Has fully elaborated reasons with specific details.
    \item Exhibits strong organization.
    \item Is fluent and uses sophisticated transitional language.
    \item May show a heightened awareness of audience.
\end{itemize}

\end{tcolorbox}
\captionof{figure}{Essay prompt and scoring rubric for ASAP Task 1.}
\label{fig:asap_1_rubric_prompt}

\begin{tcolorbox}[title=ASAP Task 7, colback=gray!5, colframe=blue!30!black]

\scriptsize
\linespread{1.1}\selectfont

\medskip
\textbf{Prompt}  
Write about patience. Being patient means that you are understanding and tolerant. A patient person experience difficulties without complaining. 

Do only one of the following: write a story about a time when you were patient OR write a story about a time when someone you know was patient OR write a story in your own way about patience.

\medskip
\textbf{Rubric} 

A rating of 0-3 on the following four traits:

\# Ideas
\begin{itemize}[nosep]
    \item Score 3: Tells a story with ideas that are clearly focused on the topic and are thoroughly developed with specific, relevant details.
    \item Score 2: Tells a story with ideas that are somewhat focused on the topic and are developed with a mix of specific and/or general details.
    \item Score 1: Tells a story with ideas that are minimally focused on the topic and developed with limited and/or general details.
    \item Score 0: Ideas are not focused on the task and/or are undeveloped.
\end{itemize}

\# Organization
\begin{itemize}[nosep]
    \item Score 3: Organization and connections between ideas and/or events are clear and logically sequenced.
    \item Score 2: Organization and connections between ideas and/or events are logically sequenced.
    \item Score 1: Organization and connections between ideas and/or events are weak.
    \item Score 0: No organization evident.
\end{itemize}

\# Style
\begin{itemize}[nosep]
    \item Score 3: Command of language, including effective and compelling word choice and varied sentence structure, clearly supports the writer's purpose and audience.
    \item Score 2: Adequate command of language, including effective word choice and clear sentences, supports the writer's purpose and audience.
    \item Score 1: Limited use of language, including lack of variety in word choice and sentences, may hinder support for the writer's purpose and audience.
    \item Score 0: Ineffective use of language for the writer's purpose and audience.
    \item 
\end{itemize}

\# Conventions
\begin{itemize}[nosep]
    \item Score 3: Consistent, appropriate use of conventions of Standard English for grammar, usage, spelling, capitalization, and punctuation for the grade level.
    \item Score 2: Adequate use of conventions of Standard English for grammar, usage, spelling, capitalization, and punctuation for the grade level.
    \item Score 1: Limited use of conventions of Standard English for grammar, usage, spelling, capitalization, and punctuation for the grade level.
    \item Score 0: Ineffective use of conventions of Standard English for grammar, usage, spelling, capitalization, and punctuation.
\end{itemize}

\end{tcolorbox}
\captionof{figure}{Essay prompt and scoring rubric for ASAP Task 7.}
\label{fig:asap_7_rubric_prompt}

\begin{tcolorbox}[title=ASAP Task 7, colback=gray!5, colframe=blue!30!black]

\footnotesize
\linespread{1}\selectfont

\medskip
\textbf{Rubric} 

Content: Paragraph is well-developed and relevant to the argument, supported with strong reasons and examples.\newline

Organization: The argument is very effectively structured and developed, making it easy for the reader to follow the ideas and understand how the writer is building the argument. Paragraphs use coherence devices effectively while focusing on a single main idea. \newline

Language: The writing displays sophisticated control of a wide range of vocabulary and collocations. The essay follows grammar and usage rules throughout the paper. Spelling and punctuation are correct throughout the paper.

\end{tcolorbox}
\captionof{figure}{Scoring rubric for DREsS.}
\label{fig:dress_rubric_prompt}

\FloatBarrier
\section{LLM prompt templates}
\label{sec:llm_prompts}

\begin{tcolorbox}[title=Base prompt template, colback=gray!5, colframe=blue!30!black]

\footnotesize
\linespread{1}\selectfont

\medskip
\textbf{System instruction} \\
\hlblue{\texttt{You are an essay grading assistant. Essay scoring range from }}
\hlyellow{\texttt{\{min\_score\}}}
\hlblue{\texttt{to }}
\hlyellow{\texttt{\{max\_score\}}}

\vspace{10pt}

\textbf{User prompt} \\
\hlblue{\texttt{\# Question:}}\\
\hlyellow{\texttt{\{ASAP\_prompt\}}}\\[4pt]

\hlblue{\texttt{\# Evaluation criteria:}}\\
\hlyellow{\texttt{\{rubric\}}}\\[6pt]

\hlblue{\texttt{\# Task: }}\\
\hlblue{\texttt{Using the evaluation criteria above, assign a score between }}
\hlyellow{\texttt{\{min\_score\}}}
\hlblue{\texttt{ and }}
\hlyellow{\texttt{\{max\_score\}}}
\hlblue{\texttt{ that best reflects the quality of the essay. Then provide a concise but detailed explanation that justifies your evaluation and references the rubric.}}\\[6pt]

\hlblue{\texttt{\# Essay:}}\\
\hlyellow{\texttt{\{essay\_text\}}}\\[6pt]

\hlblue{\texttt{\# Output format:}}\\
\hlblue{\texttt{Score: [Score according to your evaluation]}}\\
\hlblue{\texttt{Explanation: [concise but detailed explanation referencing the evaluation criteria].}}\\[6pt]

\hlblue{\texttt{\# Answer:}}

\end{tcolorbox}
\captionof{figure}{Base prompt template used to instruct LLMs for essay scoring. Blue highlights denote static prompt instructions, while yellow highlights denote placeholders replaced with dataset-specific content.}
\label{fig:base_llm_prompt_template}

\begin{tcolorbox}[title=Student-grade prompt template, colback=gray!5, colframe=blue!30!black]

\footnotesize
\linespread{1}\selectfont

\medskip
\textbf{System instruction} \\
\hlblue{\texttt{You are an essay grading assistant. Essay scoring range from }}
\hlyellow{\texttt{\{min\_score\}}}
\hlblue{\texttt{ to }}
\hlyellow{\texttt{\{max\_score\}}}

\vspace{10pt}

\textbf{User prompt} \\

\hlblue{\texttt{\# Question:}}\\
\hlyellow{\texttt{\{ASAP\_prompt\}}}\\[4pt]

\hlblue{\texttt{\# Evaluation criteria:}}\\
\hlyellow{\texttt{\{rubric\}}}\\[6pt]

\hlblue{\texttt{\# Task:}}\\
\hlblue{\texttt{Using the evaluation criteria above, assign a score between }}
\hlyellow{\texttt{\{min\_score\}}}
\hlblue{\texttt{ and }}
\hlyellow{\texttt{\{max\_score\}}}
\hlblue{\texttt{ that best reflects the quality of the essay.}}\\
\hlblue{\texttt{Grade the essay considering the expected writing skills of a grade }}
\hlyellow{\texttt{\{grade\_level\}}}
\hlblue{\texttt{ student.}}\\
\hlblue{\texttt{Then provide a concise but detailed explanation that justifies your evaluation and references the rubric.}}\\[6pt]

\hlblue{\texttt{\# Essay:}}\\
\hlyellow{\texttt{\{essay\_text\}}}\\[6pt]

\hlblue{\texttt{\# Output format:}}\\
\hlblue{\texttt{Score: [Score according to your evaluation]}}\\
\hlblue{\texttt{Explanation: [concise but detailed explanation referencing the evaluation criteria].}}\\[6pt]

\hlblue{\texttt{\# Answer:}}

\end{tcolorbox}
\captionof{figure}{Student-grade prompt template used for LLM-based essay scoring. Blue highlights denote static prompt instructions, while yellow highlights denote placeholders replaced with dataset-specific content.}
\label{fig:student_grade_llm_prompt_template}

\begin{tcolorbox}[title=Few-shot prompt template, colback=gray!5, colframe=blue!30!black]

\scriptsize
\linespread{1}\selectfont

\medskip
\textbf{System instruction} \\
\hlblue{\texttt{You are an essay grading assistant. Essay scoring range from }}
\hlyellow{\texttt{\{min\_score\}}}
\hlblue{\texttt{ to }}
\hlyellow{\texttt{\{max\_score\}}}

\vspace{10pt}

\textbf{User prompt} \\

\hlblue{\texttt{\# Examples:}}\\
\hlblue{\texttt{* Example 1:}}\\
\hlblue{\texttt{Essay 1: }}
\hlyellow{\texttt{\{student\_essay\_1\}}}\\
\hlblue{\texttt{Scores: }}
\hlyellow{\texttt{\{human\_rater\_score\_1\}}}\\[6pt]

\hlblue{\texttt{* Example 2:}}\\
\hlblue{\texttt{Essay 2: }}
\hlyellow{\texttt{\{student\_essay\_2\}}}\\
\hlblue{\texttt{Scores: }}
\hlyellow{\texttt{\{human\_rater\_score\_2\}}}\\[6pt]

\hlblue{\texttt{\# Question:}}\\
\hlyellow{\texttt{\{ASAP\_prompt\}}}\\[4pt]

\hlblue{\texttt{\# Evaluation criteria:}}\\
\hlyellow{\texttt{\{rubric\}}}\\[6pt]

\hlblue{\texttt{\# Task:}}\\
\hlblue{\texttt{Using the evaluation criteria above, assign a score between }}
\hlyellow{\texttt{\{min\_score\}}}
\hlblue{\texttt{ and }}
\hlyellow{\texttt{\{max\_score\}}}
\hlblue{\texttt{ that best reflects the quality of the essay.}}\\
\hlblue{\texttt{Then provide a concise but detailed explanation that justifies your evaluation and references the rubric.}}\\[6pt]

\hlblue{\texttt{\# Essay:}}\\
\hlyellow{\texttt{\{essay\_text\}}}\\[6pt]

\hlblue{\texttt{\# Output format:}}\\
\hlblue{\texttt{Score: [Score according to your evaluation]}}\\
\hlblue{\texttt{Explanation: [Your detailed feedback].}}\\[6pt]

\hlblue{\texttt{\# Answer:}}

\end{tcolorbox}
\captionof{figure}{Few-shot prompt template used for LLM-based essay scoring. The prompt includes example essays and corresponding human scores to guide the model's evaluation. Blue highlights denote static prompt instructions, while yellow highlights denote placeholders replaced with dataset-specific content.}
\label{fig:few_shot_llm_prompt_template}

\FloatBarrier
\section{LLM score distributions across datasets}
\label{sec:llm_score_distributions}

\begin{figure}[!ht]
\centering
\begin{subfigure}{\textwidth}
    \centering
    \includegraphics[width=\textwidth]{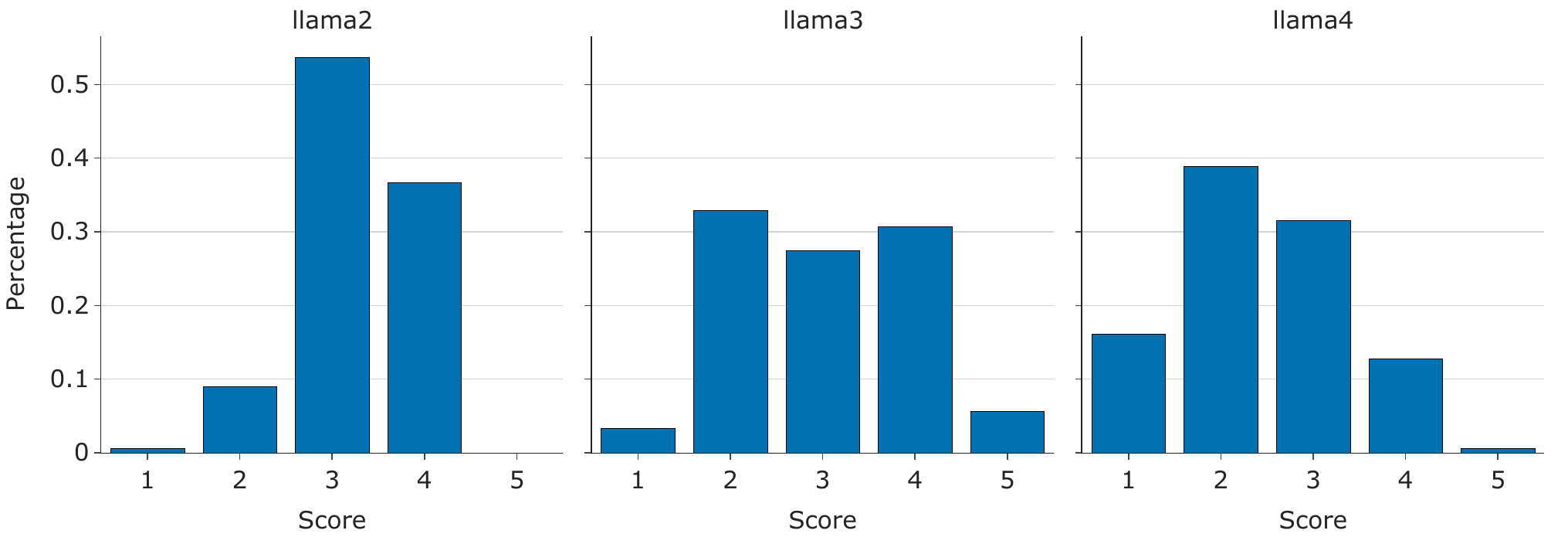}
    \caption{Llama models}
\end{subfigure}
\vspace{0.5cm}
\begin{subfigure}{\textwidth}
    \centering
    \includegraphics[width=\textwidth]{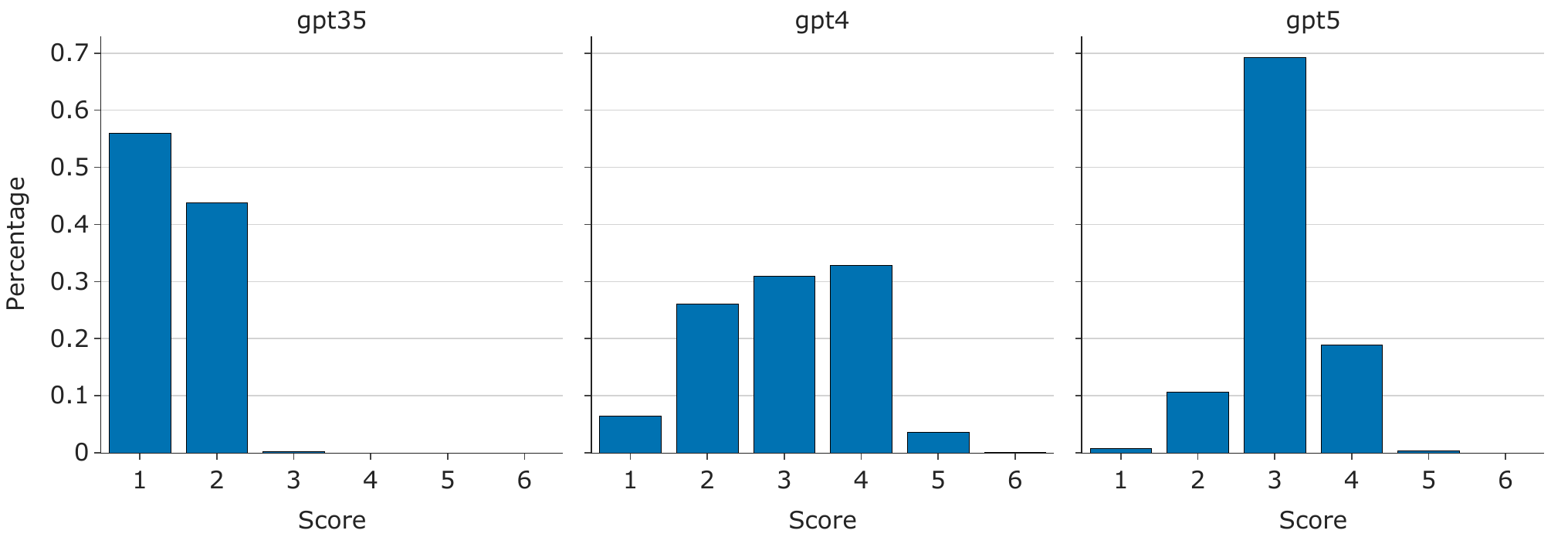}
    \caption{GPT models}
\end{subfigure}
\caption{Distribution of essay scores assigned by Llama and GPT models on the ASAP Task 1 dataset. Llama models tend to assign mid-range scores (around 2-4), while GPT models show stronger variation across versions: GPT-3.5 concentrates on very low scores (1-2), whereas GPT-5 produces mostly mid-range scores.}
\label{fig:score_distribution_asap_1_llama_gpt}
\end{figure}

\begin{figure}[!ht]
\centering
\begin{subfigure}{\textwidth}
    \centering
    \includegraphics[width=\textwidth]{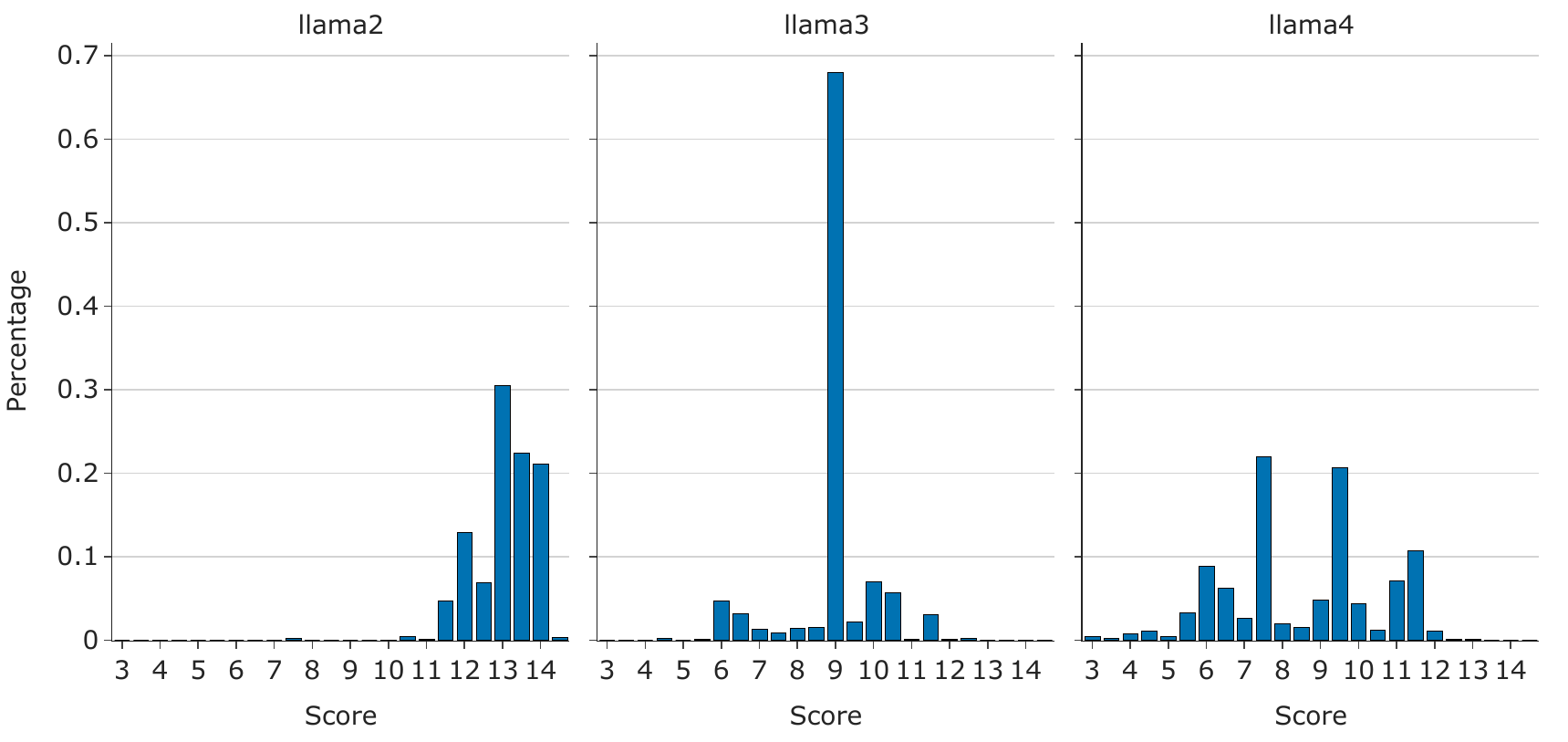}
    \caption{Llama models}
\end{subfigure}
\vspace{0.5cm}
\begin{subfigure}{\textwidth}
    \centering
    \includegraphics[width=\textwidth]{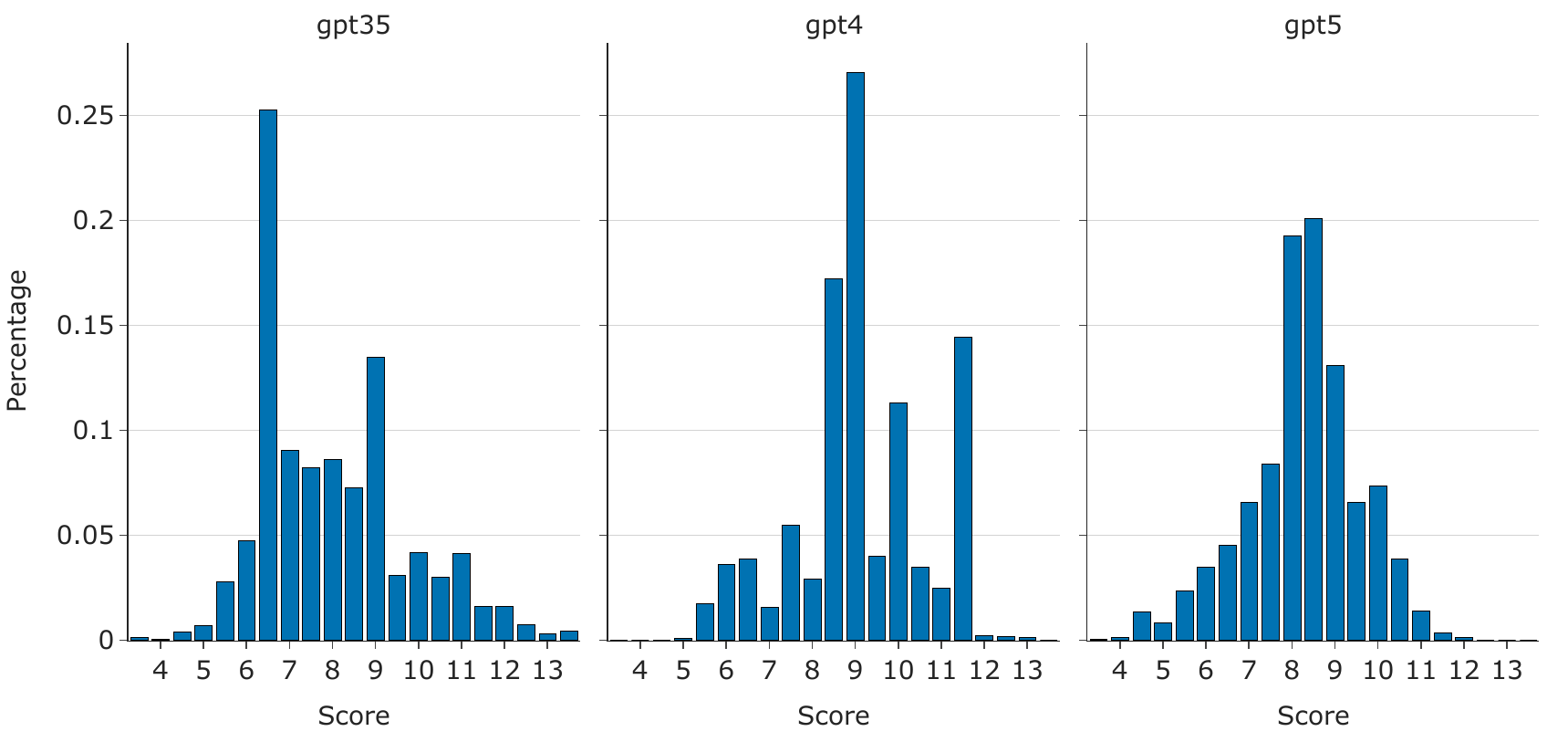}
    \caption{GPT models}
\end{subfigure}
\caption{Distribution of essay scores assigned by Llama and GPT models on the DREsS dataset. Most models concentrate scores in the mid-range (7-10), though the distributions vary across models, with Llama 2 assigning many high scores and GPT-5 showing a smoother, centered distribution..}
\label{fig:score_distribution_dress_llama_gpt}
\end{figure}

\FloatBarrier
\section{List of trait terms}
\label{sec:trait_terms}

\begin{table*}[t]
\centering
\scriptsize
\begin{tabularx}{\textwidth}{l l X}
\toprule
\textbf{Dataset} & \textbf{Trait} & \textbf{Extracted Terms} \\
\midrule

\multirow{7}{*}{ASAP Task 1}
& Content & position, claim, support, reason, detail, evidence, argument, causal reasoning, counterarguments, reasoning, thesis, elaboration, development, explanation \\
& Organization & organization, structure, paragraph, coherence, logical flow, progression, topic sentence \\
& Word Choice & language, word, phrase, vocabulary, word choice, choice of words, use of words, command of words \\
& Sentence Fluency & sentence, sentence fluency, quality of sentences, sentence structure, sentence variety, flow of sentences, sentence flow, clarity \\
& Conventions & conventions, spelling, punctuations, writing conventions, capitalization, grammar, grammatical, mechanics \\
& Transitions & transition, transitional language, transitional phrases, transitional words, connective, linking word \\
& Audience Awareness & audience awareness, awareness of audience, awareness of the audience, audience, reader, engagement, rhetorical questions, rhetorical strategies \\

\midrule

\multirow{4}{*}{ASAP Task 7}
& Ideas & ideas, details, story, development, topic, content, elaboration, explanation, example \\
& Organization & organization, connection, paragraph, transition, structure, flow, coherence, order \\
& Style & language, writing style, command of language, vocabulary, word choice, choice of words, use of words, command of words, wording, sentence, sentence structure, clarity \\
& Conventions & conventions, grammar, grammatical, spelling, capitalization, punctuation, mechanics \\

\midrule

\multirow{3}{*}{DREsS}
& Content & content, reason, example, thesis statement, argument, stance, explanation, evidence, idea development, counterargument \\
& Organization & organization, structure, coherence device, topic sentence, transition, linking phrase, flow, coherence, cohesion, progression of ideas \\
& Language & language, vocabulary, collocation, grammar, usage rules, spelling, punctuation, verb tense, subject-verb agreement, article usage, word choice, sentence structure, sentence variety, phrasing, clarity, fluency \\

\bottomrule
\end{tabularx}
\caption{Trait-specific lexical cues used to identify praise and criticism mentions in model-generated feedback across datasets.}
\label{tab:trait_terms}
\end{table*}

\FloatBarrier
\section{Human-LLM agreement across different prompts}
\label{sec:human_llm_agreement_prompt_variants}

\begin{figure}[!ht]
\centering
\includegraphics[width=\textwidth]{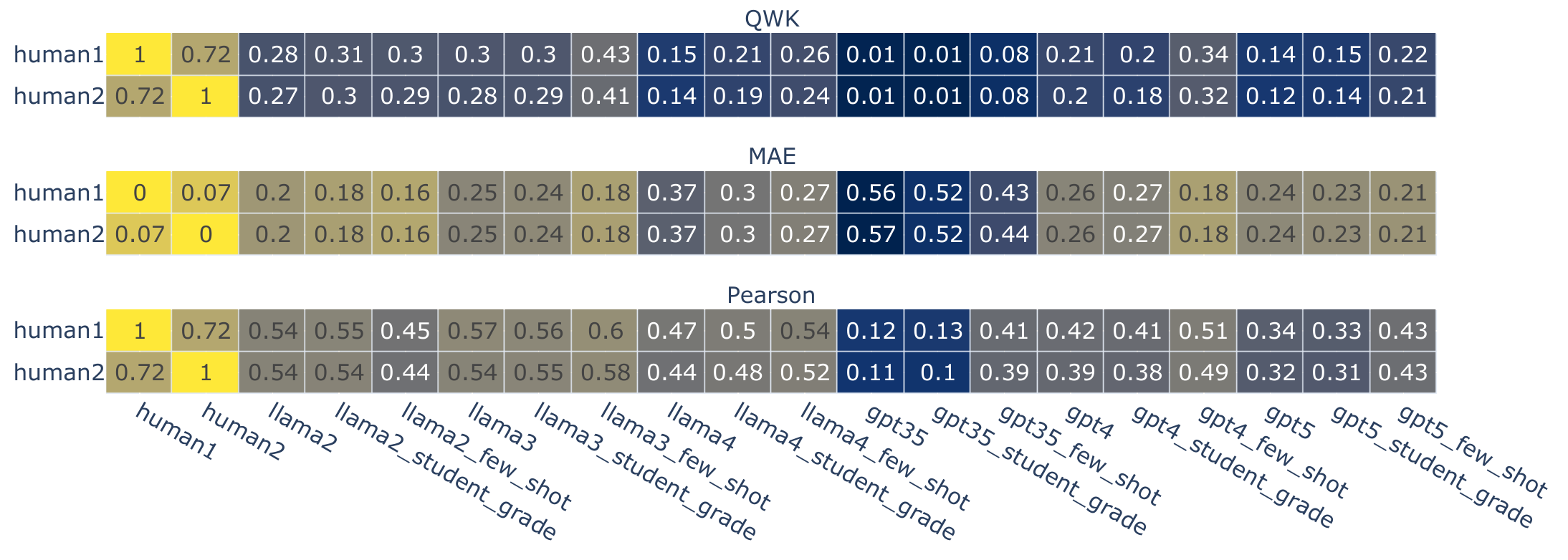}  
\caption{Agreement between LLM-generated scores and human grades under different prompt configurations on ASAP Task 1, measured using QWK, MAE, and Pearson correlation. Lighter cells indicate higher agreement (and lower error for MAE). Human raters show substantially stronger agreement with each other than with LLM predictions, while prompt variants yield only modest and inconsistent improvements across models.}\label{fig:prompt_variant_agreement_with_human_asap1}
\end{figure}

\begin{figure}[!ht]
\centering
\includegraphics[width=\textwidth]{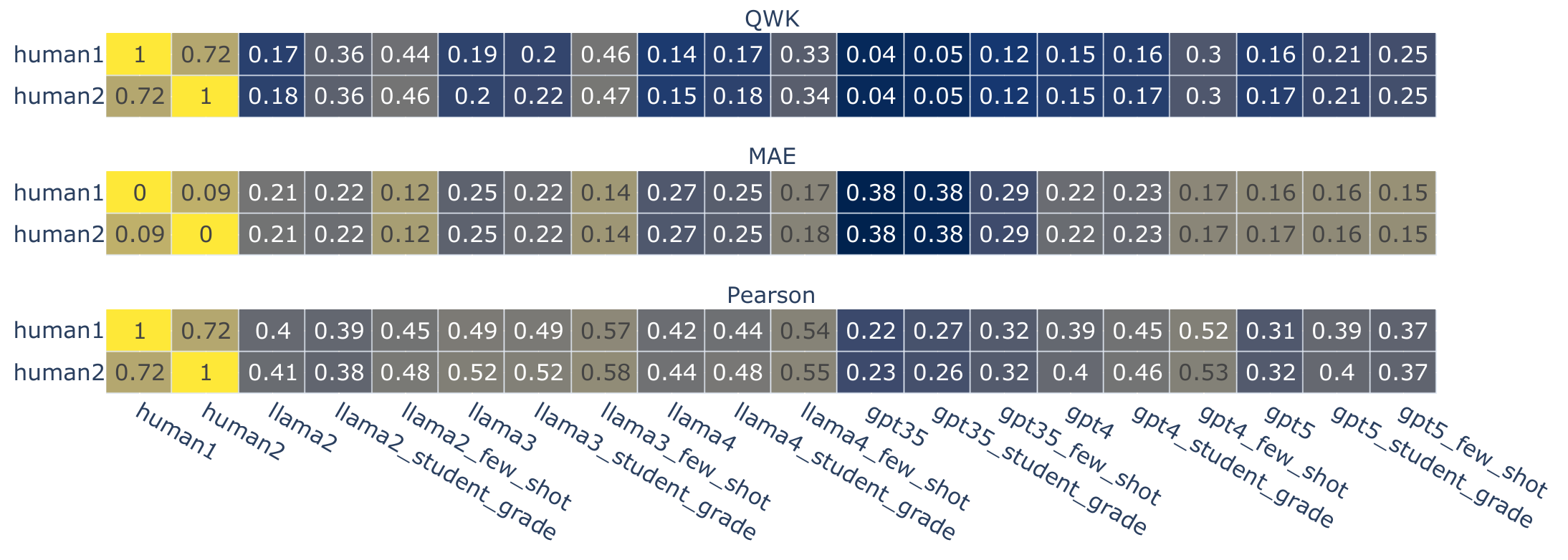}  
\caption{Agreement between LLM-generated scores and human grades under different prompt configurations on ASAP Task 7, measured using QWK, MAE, and Pearson correlation. Lighter cells indicate higher agreement (and lower error for MAE). Human-human agreement remains substantially stronger than LLM-human agreement, while prompt variants produce only limited and inconsistent improvements.}\label{fig:prompt_variant_agreement_with_human_asap7}
\end{figure}

\begin{figure}[!ht]
\centering
\includegraphics[width=\textwidth]{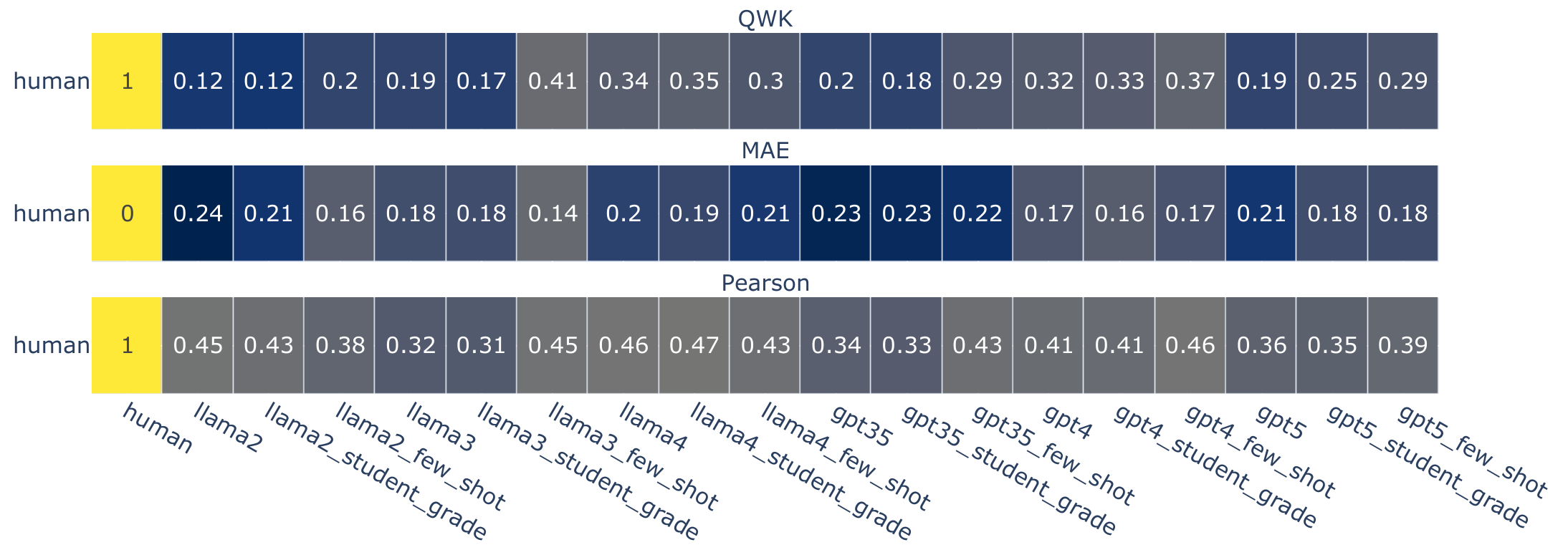}  
\caption{Agreement between LLM-generated scores and human grades under different prompt configurations on the DREsS dataset, measured using QWK, MAE, and Pearson correlation. Lighter cells indicate higher agreement (and lower error for MAE). Agreement levels vary across models, and prompt variants lead to only modest and inconsistent changes.}\label{fig:prompt_variant_agreement_with_human_dress}
\end{figure}

\FloatBarrier
\section{LLM-LLM agreement across prompts}
\label{sec:llm_llm_prompt_agreement}

\begin{figure}[!ht]
\centering
\includegraphics[width=\textwidth]{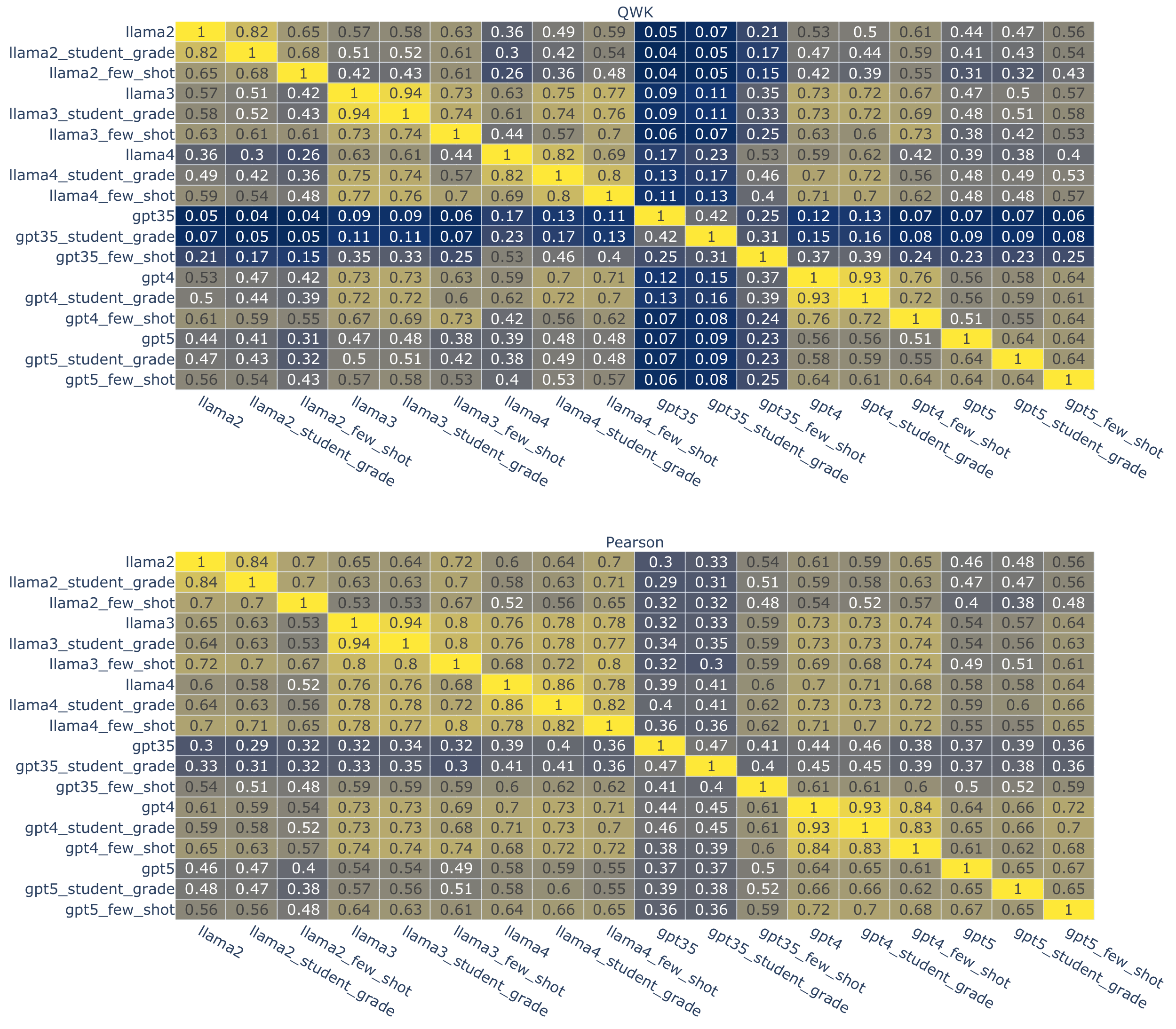}  
\caption{Pairwise agreement between LLM-generated scores on the ASAP Task 1 dataset, measured using QWK (top) and Pearson correlation (bottom). Lighter cells indicate higher agreement. LLMs from the same family and prompt configuration tend to show stronger agreement with each other, while agreement between different model families is generally lower.}\label{fig:llm_pairwise_agreement_asap_1}
\end{figure}

\begin{figure}[!ht]
\centering
\includegraphics[width=\textwidth]{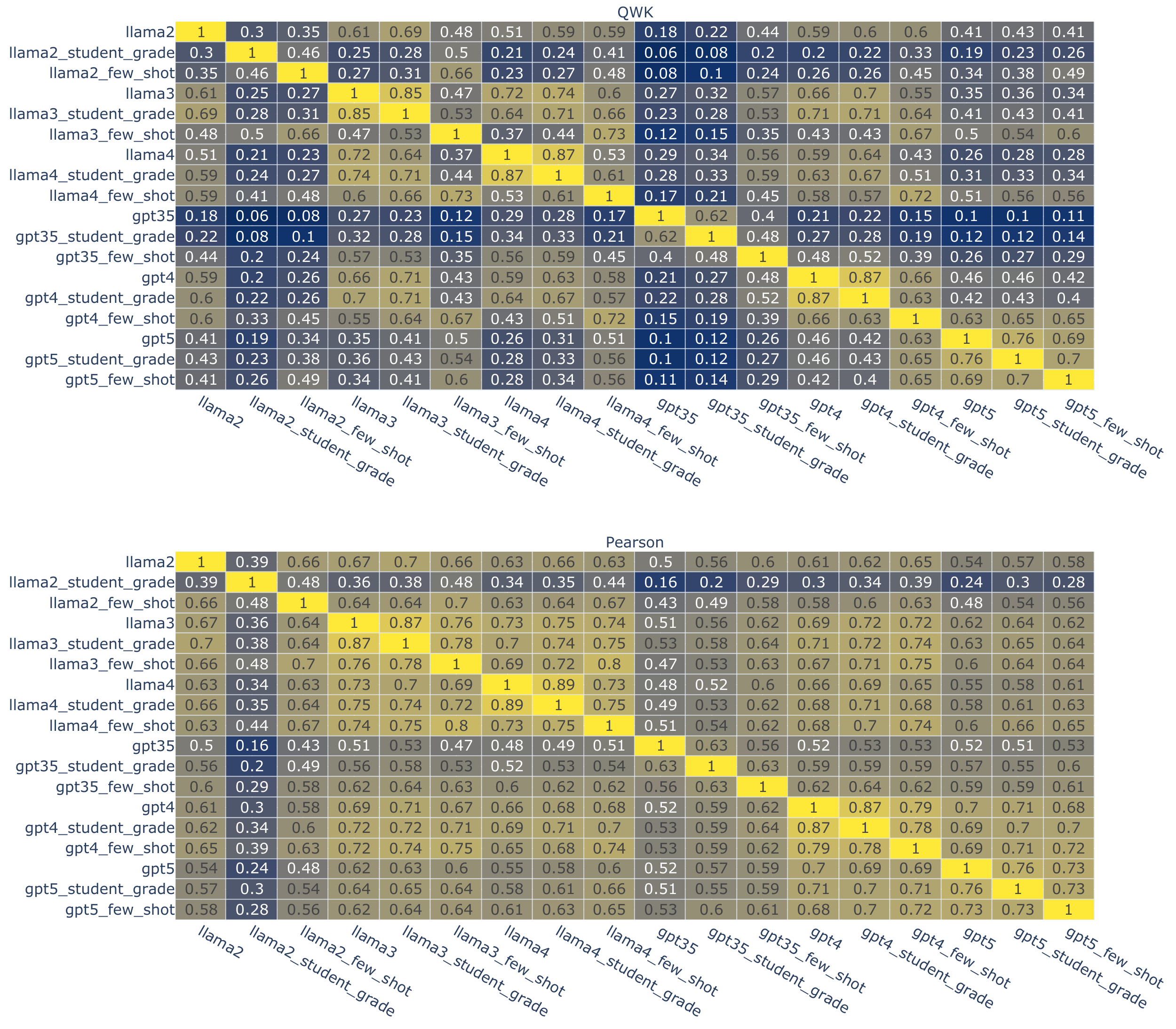}  
\caption{Pairwise agreement between LLM-generated scores on the ASAP Task 7 dataset, measured using QWK (top) and Pearson correlation (bottom). Lighter cells indicate higher agreement. Agreement is generally stronger between variants of the same model family and prompt configuration than across different models.}\label{fig:llm_pairwise_agreement_asap_7}
\end{figure}

\begin{figure}[!ht]
\centering
\includegraphics[width=\textwidth]{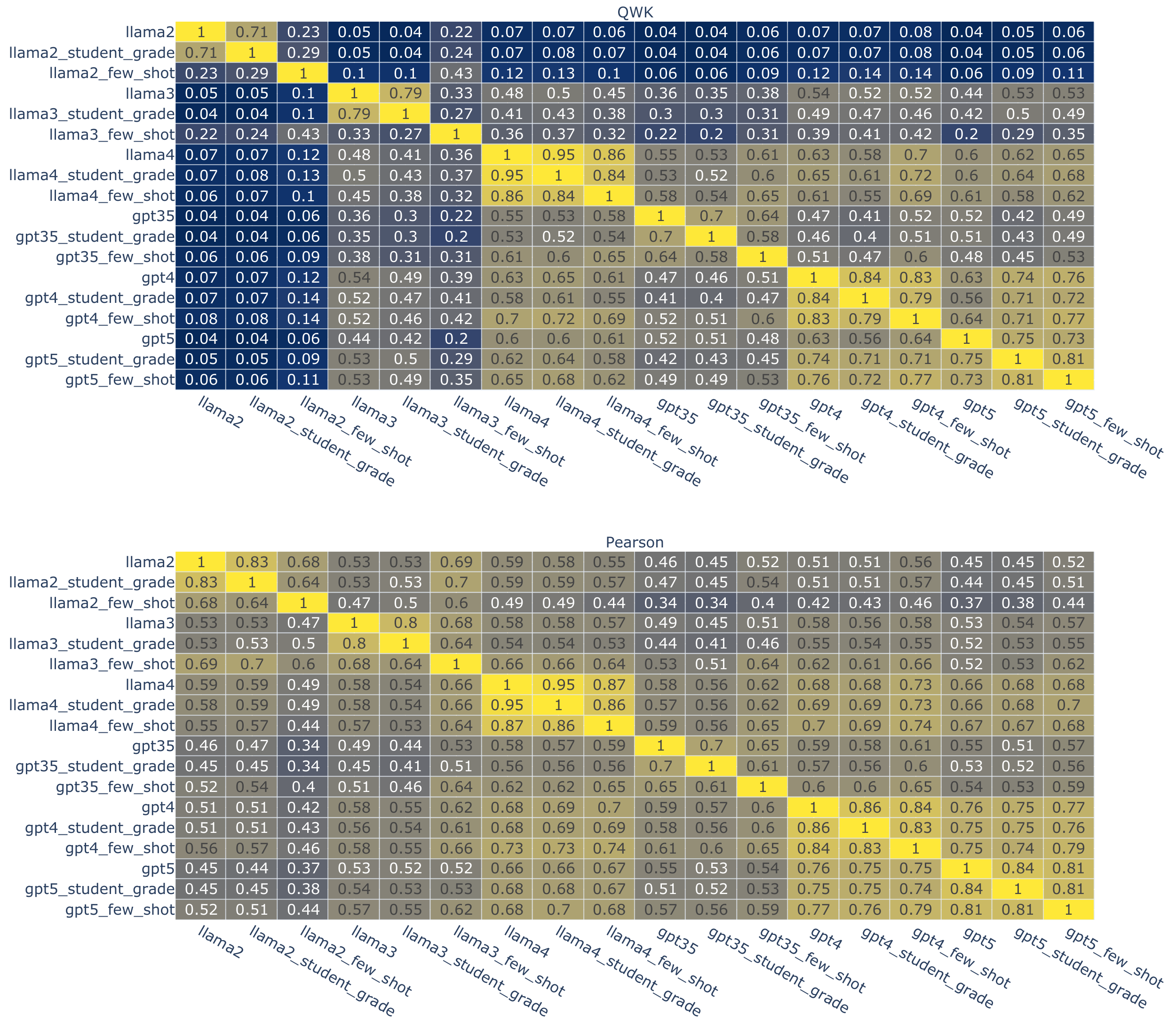}  
\caption{Pairwise agreement between LLM-generated scores on the DREsS dataset, measured using QWK (top) and Pearson correlation (bottom). Lighter cells indicate higher agreement. Agreement is generally stronger between variants of the same model family and prompt configuration than across different models.}\label{fig:llm_pairwise_agreement_dress}
\end{figure}

\FloatBarrier
\section{Sample essays with high Human-LLM disagreement}

\begin{tcolorbox}[title=Essay 2108 (DREsS), colback=gray!5, colframe=blue!30!black]

\footnotesize
\linespread{1}\selectfont

\textbf{Prompt}  
What is a good hobby for college students? Give specific reasons and details to support your opinion.

\medskip
\textbf{Essay}  
In Korea, most of high school students stay with very regular schedule and eat healthy lunch. However, in college they start drinking and often eat fast food. Their health going bad!

College students have much free time and they can do their hobby more. Then, spending time to exercise. Exercise can help college studentMany college people want to make girl or boyfriend. If they start exercise and make healthy body, it can be positive effect to people.

\medskip
\textbf{Human Score} 3/15 (Content: 1/5, Organization: 1/5, Language: 1/5)

\medskip
\textbf{GPT-4 Score} 5.5/15 (Content: 2/5, Organization: 2/5, Language: 1.5/5)

\medskip
\textbf{GPT-4 Feedback}  
The essay addresses the question and identifies exercise as a good hobby for college students, connecting this to health and social benefits. However, the content lacks strong supporting reasons and detailed examples, and the argument is underdeveloped. The organization is weak, with abrupt transitions and ideas presented in a somewhat disjointed manner, making it difficult for the reader to follow a clear line of reasoning. Coherence devices are mostly absent. Language control is limited; there are several grammatical errors, awkward phrasing, incomplete sentences, and issues with word choice that impede comprehension. The vocabulary is basic, and there are mistakes with grammar, usage, and punctuation throughout the essay.

\medskip
\textbf{Other LLM Scores} 

LlaMA-2: 7.5/15 (2.5 + 3 + 2) \\
LlaMA-3: 6.5/15 (2.5 + 2 + 2) \\
LLaMA-4: 5/15 (2 + 1.5 + 1.5) \\
GPT-3.5: 5/15 (1.5 + 1.5 + 2) \\
GPT-5: 4.5/15 (2 + 1.5 + 1) \\

\end{tcolorbox}
\captionof{figure}{DREsS essay with a low human score but higher scores assigned by LLMs, illustrating a case where models reward prompt relevance despite weak content and language quality.}
\label{fig:llm_human_disagreement_lowest_score}

\newpage
\begin{tcolorbox}[title=Essay 19052 (ASAP Task 7), colback=gray!5, colframe=blue!30!black]
\scriptsize
\linespread{0.95}\selectfont
\textbf{Prompt}  
Write about patience. Being patient means that you are understanding and tolerant. A patient person experience difficulties without complaining. 

Do only one of the following: write a story about a time when you were patient OR write a story about a time when someone you know was patient OR write a story in your own way about patience.

\medskip
\textbf{Essay}  
Have you ever been in a situation when you know something good is coming or is going to happen and you just @CAPS1’t control yourself? You ask your parents, “when” and they say, “soon just have some patience!” well this has happened to me multiple times, such as when we were going to @LOCATION1 or @LOCATION2, but on this special occasion, getting our new dog. I decided to be a mature teenager and be patient. It was @DATE1 @TIME1, the day my family was getting a dog and I was so excited. My stomach was filled with butterflies, trying to fly out and explode in happiness. Everyone else in my family was the same way, but just then, I remembered my goal about being a mature teenager and patient. So I carried on with my normal, everyday life. Eat breakfast, comb hair, get dressed, brush teeth, and go to school. Later, when I came home, my mom announced that my dad will arrive with my puppy in @NUM1 minutes. So I sat on my chair, in my room and watched the clock tic and toc for about @NUM2 minutes. Then, I walked in to the living room, where the rest of my family was sitting and waited till I heard a door slam. Suddenly. Boom! The door slammed on my dads truck. My mom, brother, and sister bolted to the door as I simply walked to see the puppy. “aww!” everyone screamed. “@CAPS1 I hold her,” and “she’s my dog,” and “let’s name her” also came in the conversation. Then, my dad glared at me, standing their quietly and handled me the dog. “you are being so patient, you @CAPS1 be the one to hold her first!” I was so proud of myself. I was patient and I was the one that got to hold her first. Being patient @CAPS1 pay off!

\medskip
\textbf{Human Score} 12/12 (Ideas: 3/3, Organization: 3/3, Style: 3/3, Conventions: 3/3)

\medskip
\textbf{GPT-3.5 Score} 9/12 (Ideas: 3/3, Organization: 2/3, Style: 2/3, Conventions: 2/3)

\medskip
\textbf{GPT-3.5 Feedback}  
The essay demonstrates a clear focus on the topic of patience through a personal story about waiting for a new dog to arrive. The writer effectively conveys the concept of patience by describing their internal struggle to control their excitement and their ultimate success in being patient. The narrative is engaging and specific, showcasing the idea of patience without explicitly stating it. However, the organization could be improved for better flow and coherence. The events are somewhat logically sequenced but transitions between paragraphs could be smoother to enhance the overall structure. In terms of style, the writer shows some command of language with descriptive details and varied sentence structure, contributing to the reader's understanding and engagement. Some areas could benefit from more refined language choices and sentence structures to elevate the writing further. Regarding conventions, there are minimal errors in standard English conventions, with consistent use of grammar, punctuation, and spelling. Overall, the essay effectively conveys the theme of patience through a personal anecdote, with room for improvement in organization and language sophistication to enhance the overall impact.

\medskip
\textbf{Other LLM Scores} 

LlaMA-2: 12/12 (3 + 3 + 3 + 3) \\
LlaMA-3: 7/12 (2 + 2 + 2 + 1) \\
LLaMA-4: 7/12 (2 + 2 + 2 + 1) \\
GPT-4: 7/12 (2 + 2 + 2 + 1) \\
GPT-5: 9/12 (3 + 3 + 2 + 1) \\

\end{tcolorbox}
\captionof{figure}{Example essay from ASAP Task 7 receiving a perfect human score (12/12) but lower scores from several LLMs, illustrating human-LLM disagreement in the evaluation of organization and language quality.}
\label{fig:llm_human_disagreement_highest_score}

\newpage
\begin{tcolorbox}[title=Essay 17891 (ASAP Task 7), colback=gray!5, colframe=blue!30!black]
\footnotesize
\linespread{1}\selectfont
\textbf{Prompt}  
Write about patience. Being patient means that you are understanding and tolerant. A patient person experience difficulties without complaining. 

Do only one of the following: write a story about a time when you were patient OR write a story about a time when someone you know was patient OR write a story in your own way about patience.

\medskip
\textbf{Essay}  
It had been about one??? after I got my ipod touch and I really wanted to get this certain??? no it was a movie. The movie was I @CAPS1 @CAPS2. I had already seen it once from my cousins ipod but it was so cool, I had to watch it again. It had so much action in it. And zombies??? attack. Multiple words @CAPS3

\medskip
\textbf{Human Score} 12/12 (Ideas: 3/3, Organization: 3/3, Style: 3/3, Conventions: 3/3)

\medskip
\textbf{GPT-5 Score} 3/12 (Ideas: 0/3, Organization: 1/3, Style: 1/3, Conventions: 1/3)

\medskip
\textbf{GPT-5 Feedback}  
- Ideas (0): The essay does not address the assigned topic of patience. Instead it briefly describes wanting to watch a movie on an iPod and mentions action and zombies. Because the response is off-task and undeveloped with no relevant details about being patient (or someone being patient), it earns a score of 0 under the rubric.  
- Organization (1): There is a very minimal sequence of events (received iPod, wanted a movie, had seen it before), but the events are presented in a fragmented, choppy way with abrupt breaks and no clear logical development or transitions. This weak organization fits a score of 1.  - Style (1): Word choice and sentence structure are limited and informal. Repeated punctuation ("???"), interjections, and placeholder tokens interrupt flow and reduce clarity. The language shows minimal variety and does not effectively support a purpose or audience, so it receives a 1.  
- Conventions (1): The essay contains multiple errors and nonstandard elements: inconsistent capitalization (e.g., "ipod" and "i"), unexplained placeholder tokens (@CAPS1, @CAPS2, @CAPS3), strange punctuation, and other grammatical/spelling issues. These limited uses of conventions earn a score of 1.  Suggestions for improvement: stay on topic by describing a situation that demonstrates patience; develop the story with specific, relevant details; organize events clearly with transitions; and revise for standard grammar, punctuation, and spelling (remove placeholder tokens and fix capitalization).

\end{tcolorbox}
\captionof{figure}{Example essay from ASAP Task 7 receiving a perfect human score (12/12) but a much lower score from the LLM (3/12), illustrating human-LLM disagreement in the interpretation of topic relevance and writing quality.}
\label{fig:sample-essay}

\FloatBarrier
\section{Essay readability and length by score}
\label{sec:essay_readability_length}

\begin{figure}[!ht]
\centering
\includegraphics[width=\textwidth]{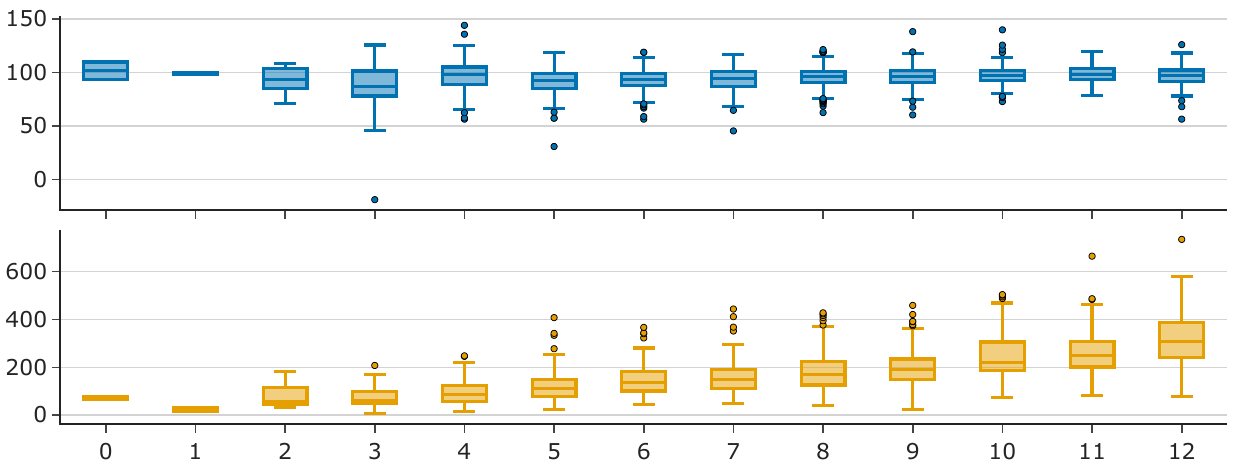}
\caption{Readability scores (Flesch Reading Ease, top) and essay lengths (number of tokens, bottom) for the ASAP Task 7 dataset. Each boxplot represents essays within a specific score range. Essay length increases consistently with score, while readability varies less systematically across score levels.}
\label{fig:readability_length_asap7}
\end{figure}

\begin{figure}[!ht]
\centering
\includegraphics[width=\textwidth]{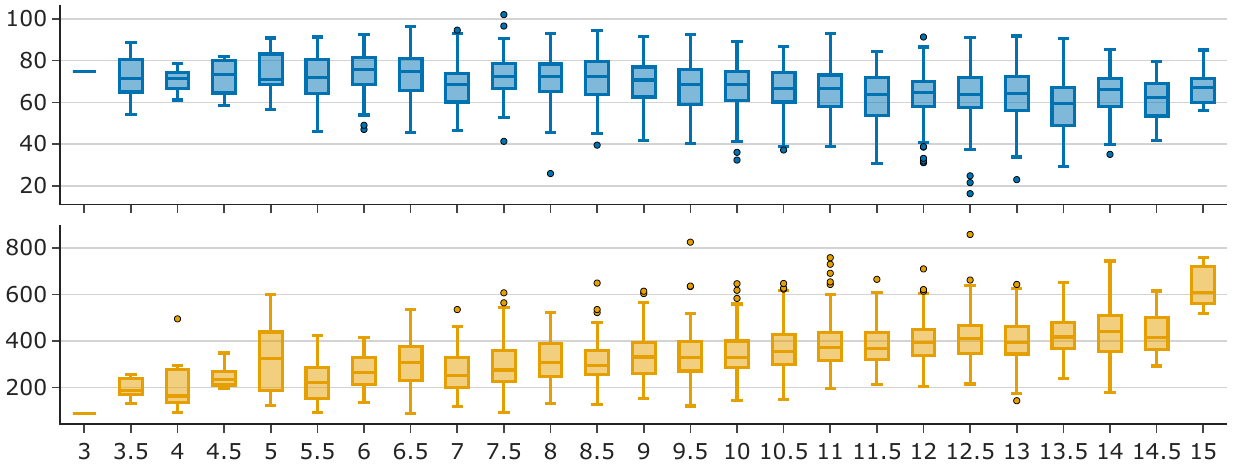}
\caption{Readability scores (Flesch Reading Ease, top) and essay lengths (number of tokens, bottom) for the DREsS dataset. Each boxplot represents essays within a specific score range. Similar to ASAP Task 7, essay length increases with score, while readability shows weaker variation across score levels.}
\label{fig:readability_length_dress}
\end{figure}

\FloatBarrier
\section{Trait importance analysis with SHAP}
\label{sec:shap_analysis}

\begin{figure}[!ht]
\centering

\begin{subfigure}{0.32\textwidth}
    \centering
    \includegraphics[width=\textwidth]{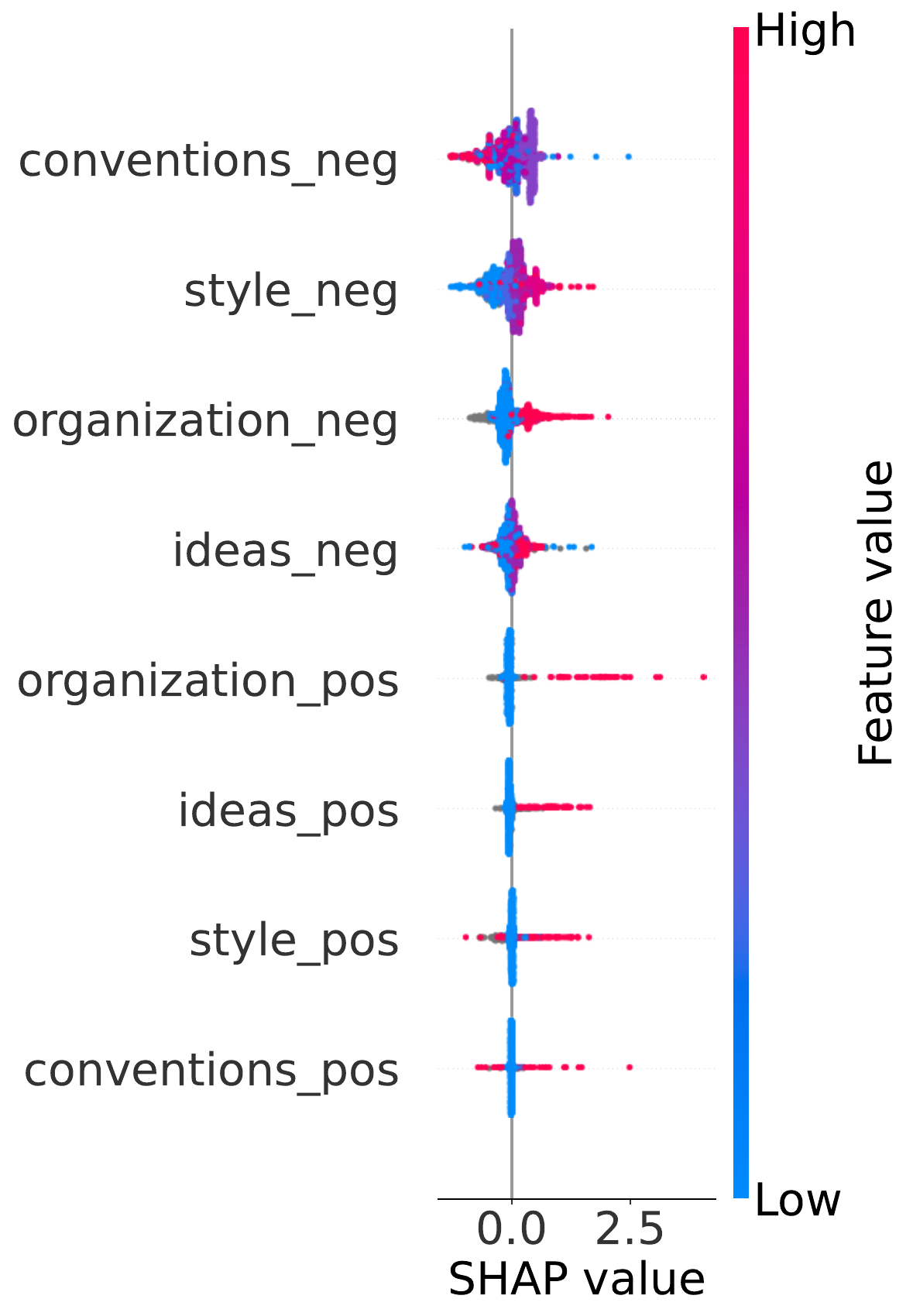}
    \caption{GPT-3.5}
\end{subfigure}
\hfill
\begin{subfigure}{0.32\textwidth}
    \centering
    \includegraphics[width=\textwidth]{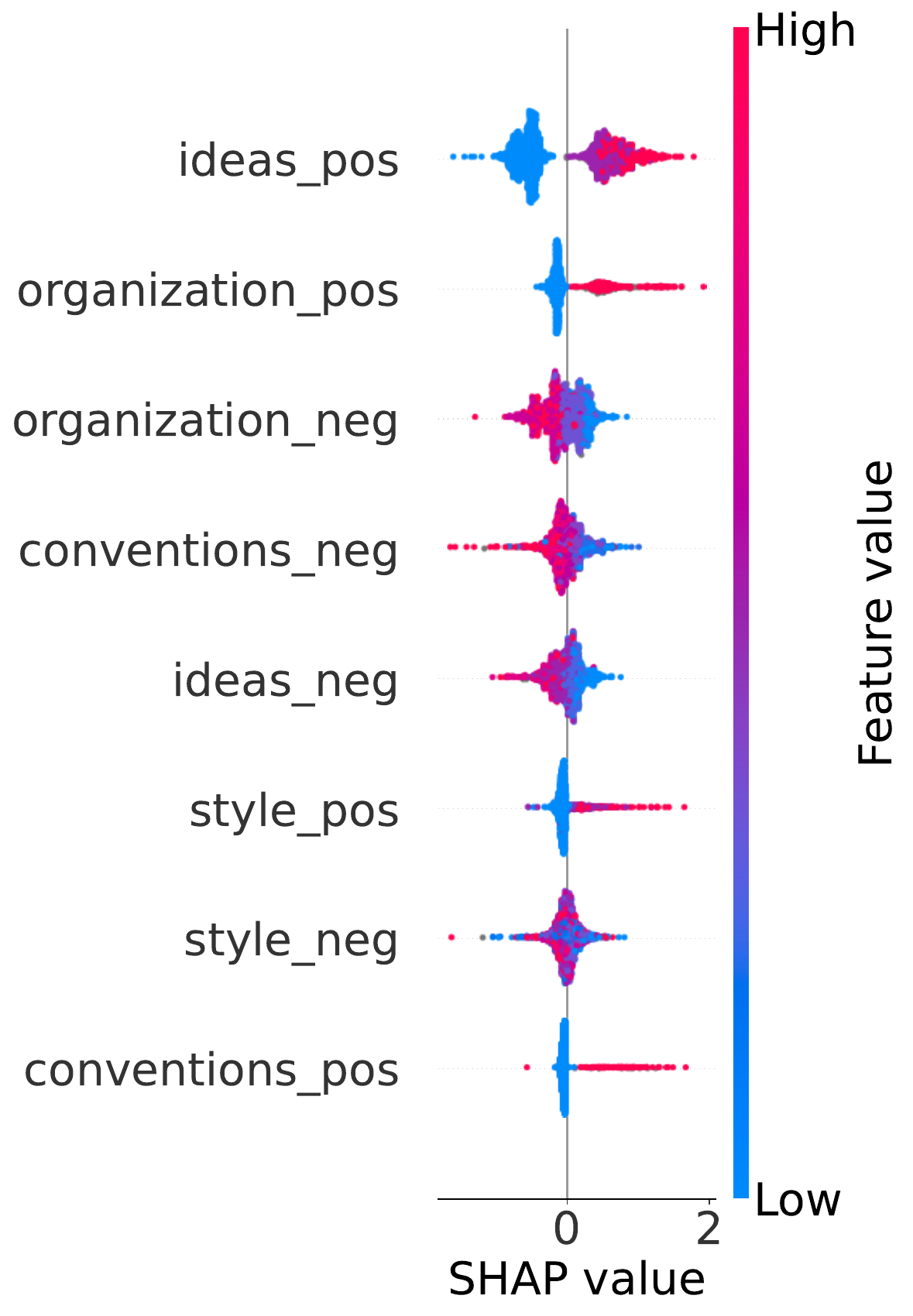}
    \caption{GPT-4}
\end{subfigure}
\hfill
\begin{subfigure}{0.32\textwidth}
    \centering
    \includegraphics[width=\textwidth]{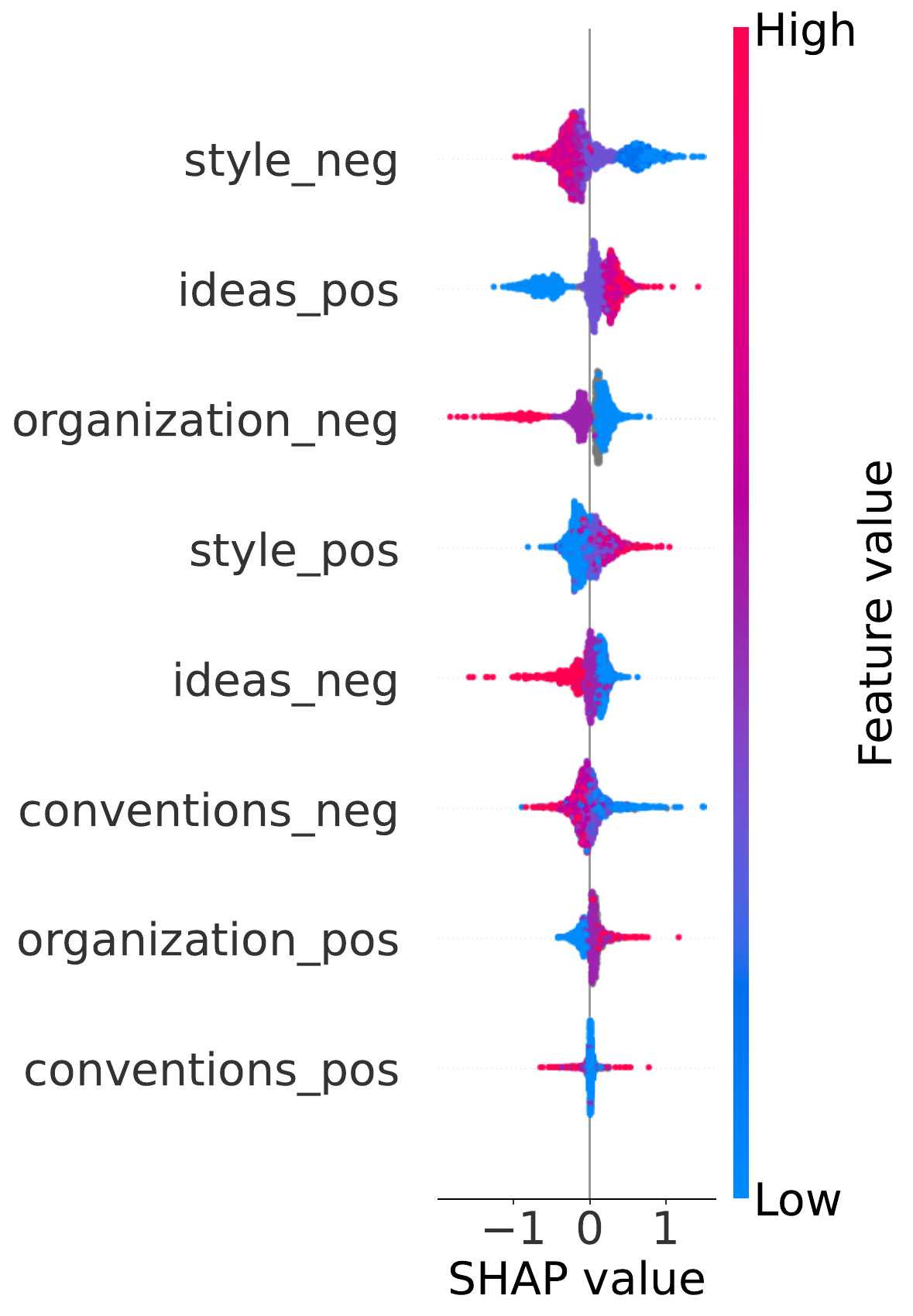}
    \caption{GPT-5}
\end{subfigure}

\caption{SHAP beeswarm plots showing the impact of trait-specific praise and criticism mentions on predicted essay scores for GPT models on ASAP Task 7. Features are sorted in decreasing order of importance (mean absolute SHAP value). Points represent individual essays, and horizontal position indicates the contribution to the predicted score. Praise-related features tend to increase predicted scores, while criticism-related features decrease them.}
\label{fig:beeswarm_gpt_asap7}
\end{figure}

\begin{figure}[!ht]
\centering

\begin{subfigure}{0.32\textwidth}
    \centering
    \includegraphics[width=\textwidth]{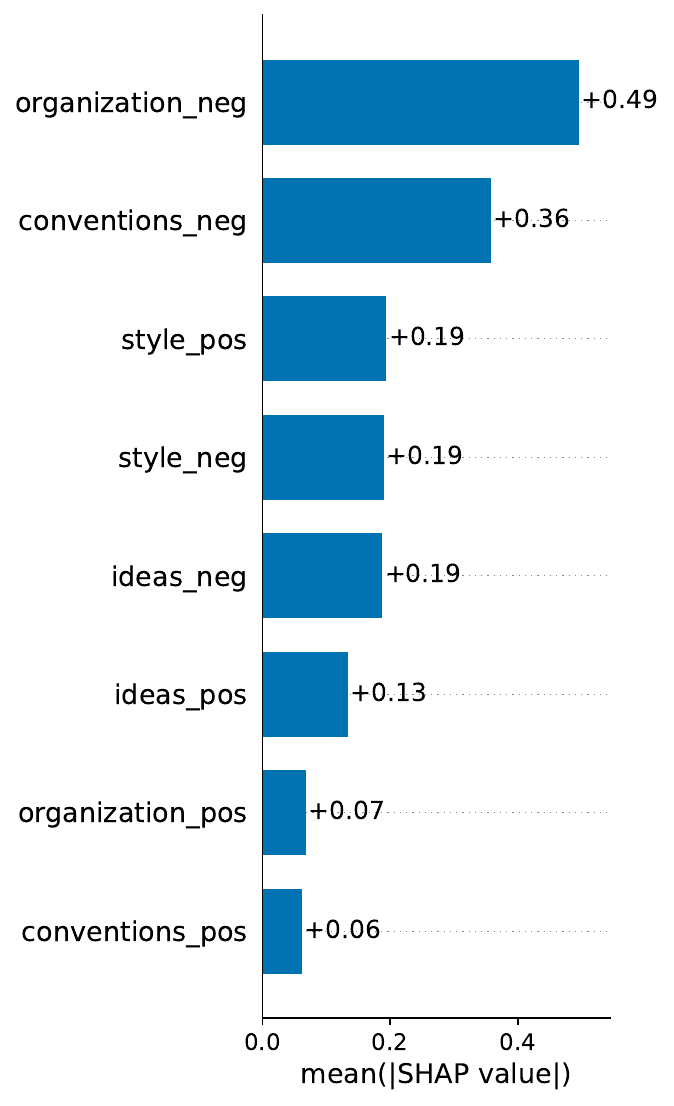}
    \caption{Llama 2}
\end{subfigure}
\hfill
\begin{subfigure}{0.32\textwidth}
    \centering
    \includegraphics[width=\textwidth]{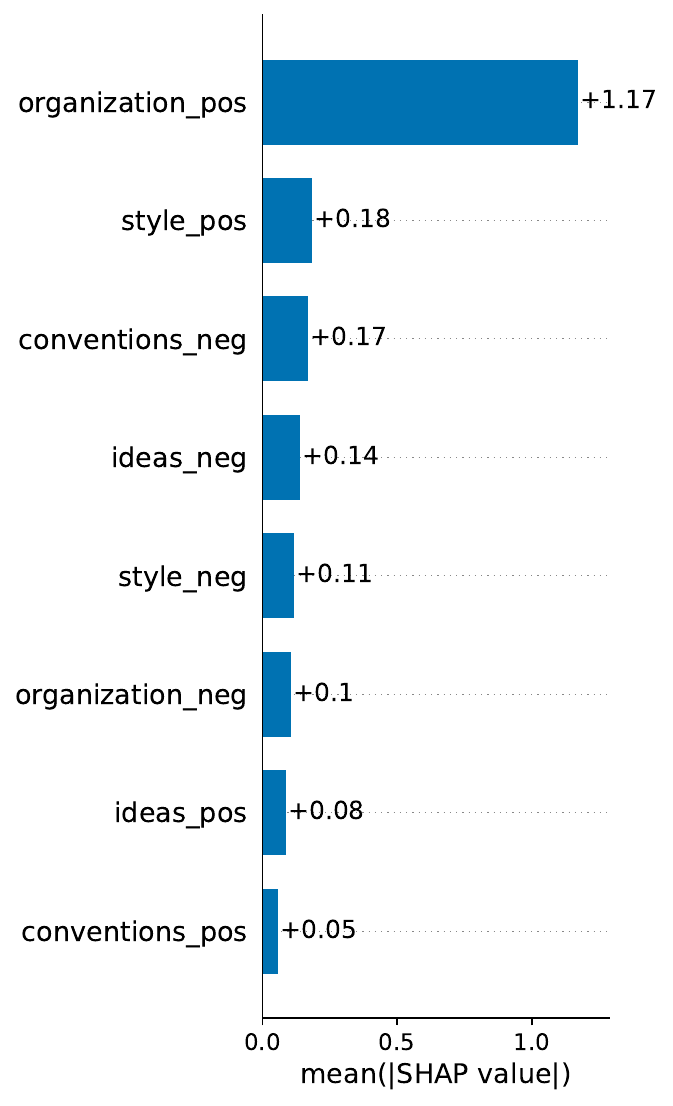}
    \caption{Llama 3}
\end{subfigure}
\hfill
\begin{subfigure}{0.32\textwidth}
    \centering
    \includegraphics[width=\textwidth]{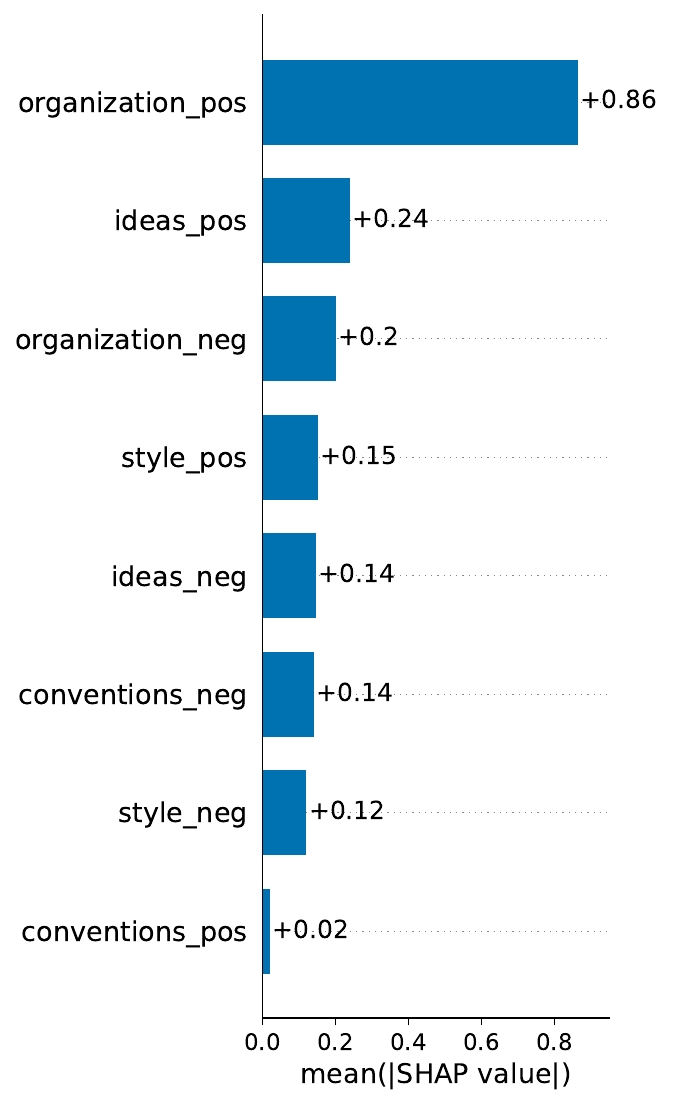}
    \caption{Llama 4}
\end{subfigure}

\caption{Average absolute SHAP values for trait-specific praise and criticism features for Llama models on ASAP Task 7. Higher values indicate features that contribute more strongly to predicted scores, revealing differences in how models prioritize rubric traits.}
\label{fig:shap_barplot_llama_asap_7}
\end{figure}

\begin{figure}[!ht]
\centering

\begin{subfigure}{0.32\textwidth}
    \centering
    \includegraphics[width=\textwidth]{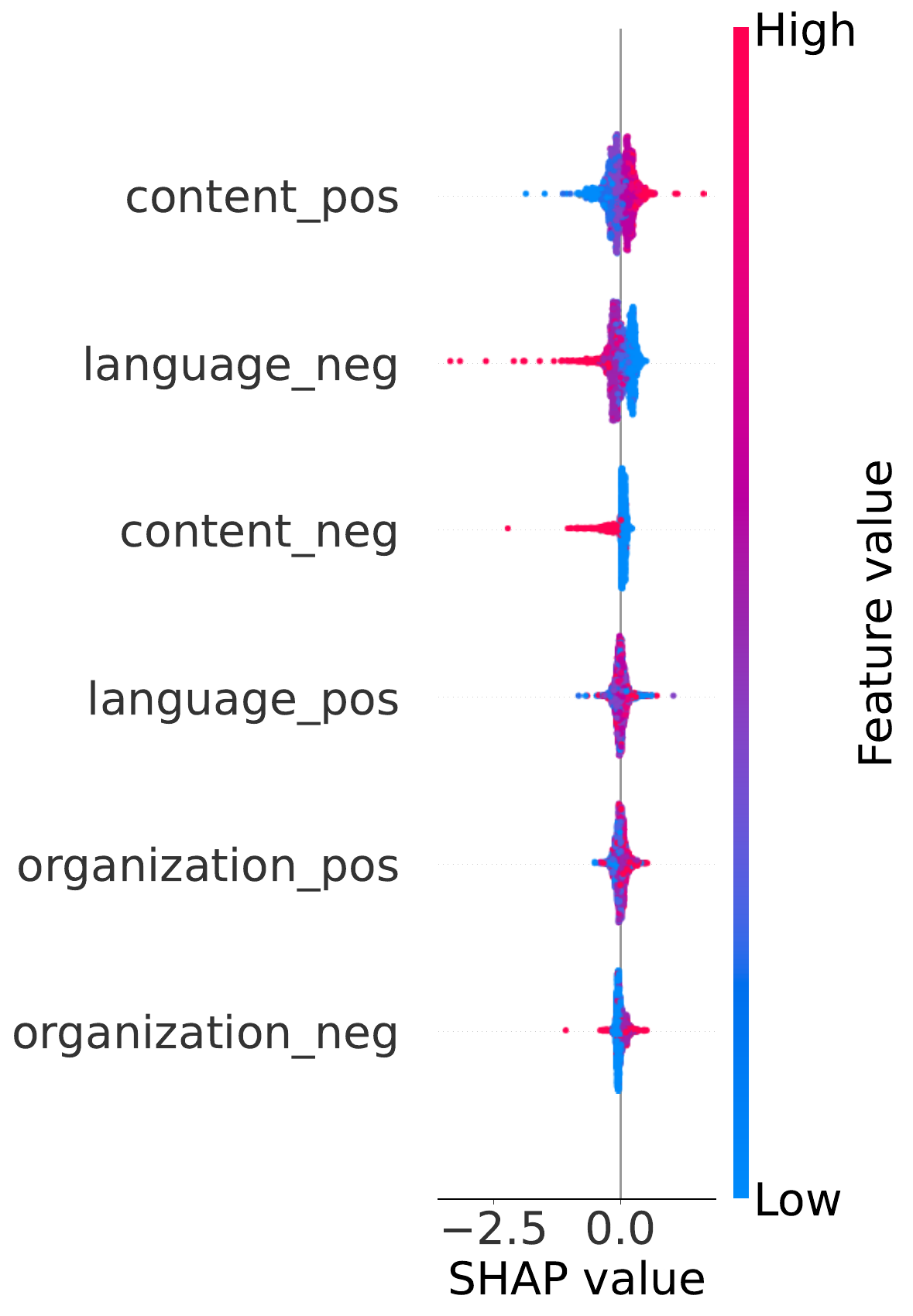}
    \caption{Llama 2}
\end{subfigure}
\hfill
\begin{subfigure}{0.32\textwidth}
    \centering
    \includegraphics[width=\textwidth]{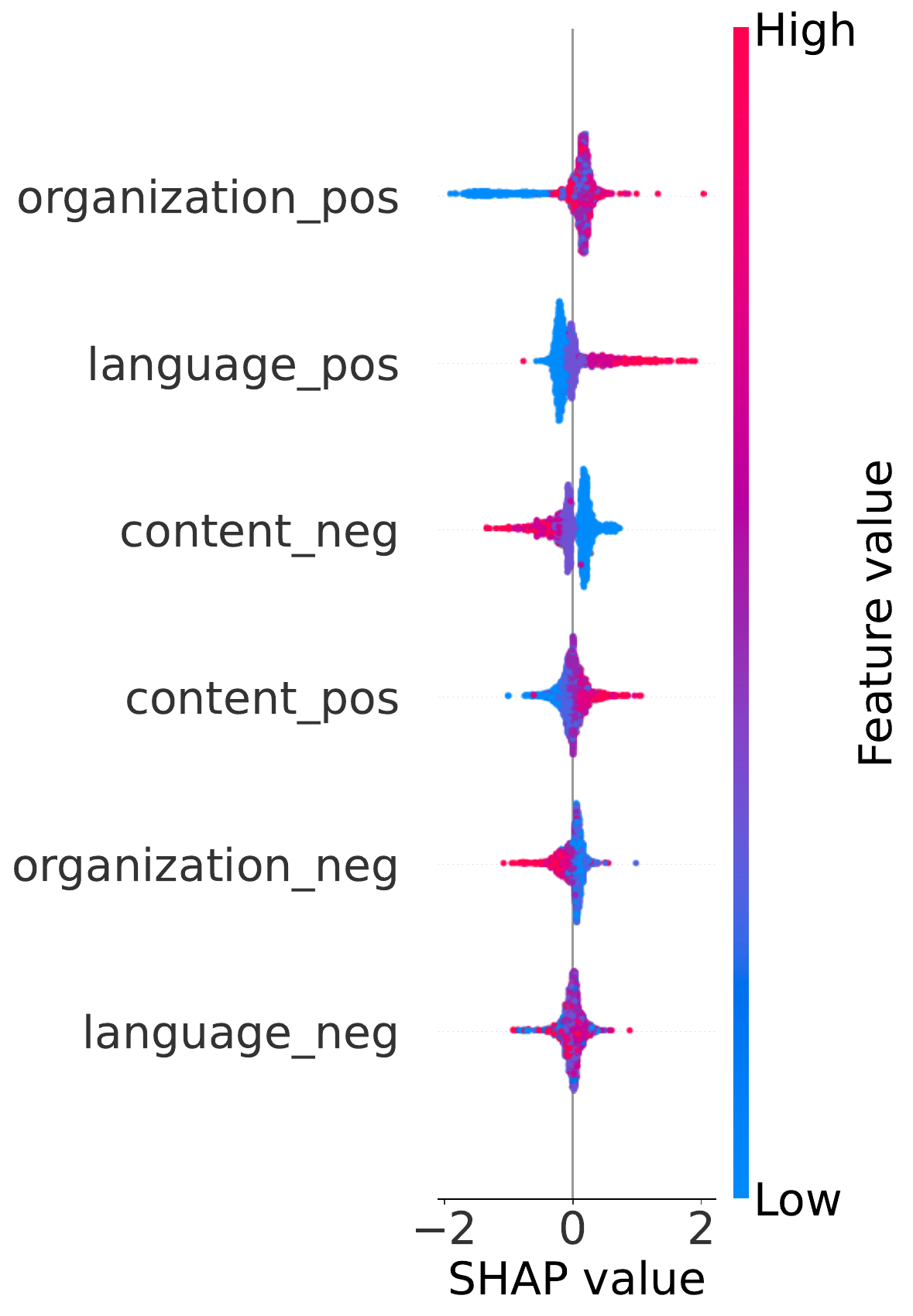}
    \caption{Llama 3}
\end{subfigure}
\hfill
\begin{subfigure}{0.32\textwidth}
    \centering
    \includegraphics[width=\textwidth]{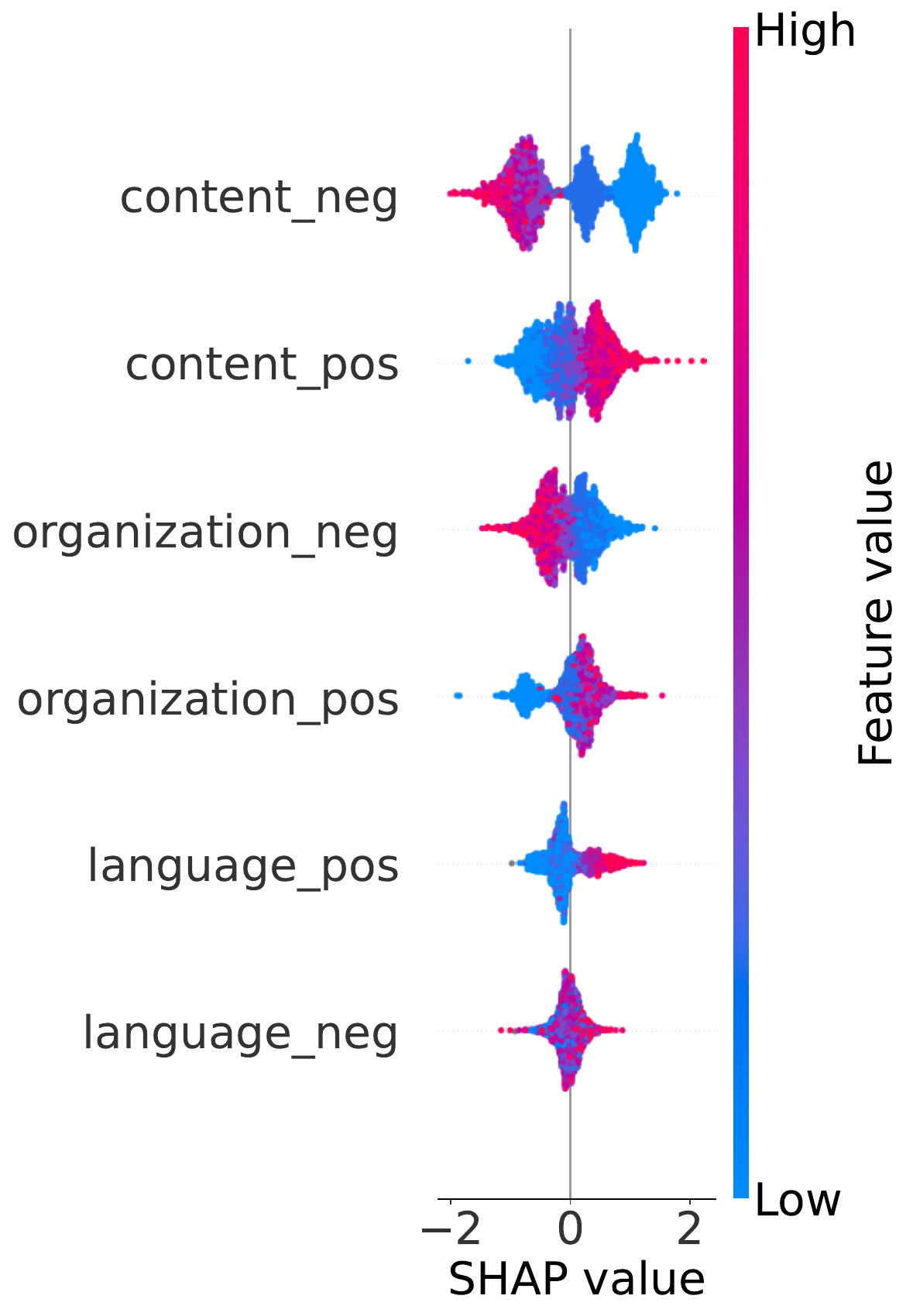}
    \caption{Llama 4}
\end{subfigure}

\caption{SHAP beeswarm plots showing the impact of trait-specific praise and criticism mentions on predicted essay scores for Llama models on DREsS. Features are sorted in decreasing order of importance (mean absolute SHAP value). Points represent individual essays, and horizontal position indicates the contribution to the predicted score. Praise-related features tend to increase predicted scores, while criticism-related features decrease them.}
\label{fig:beeswarm_llama_dress}
\end{figure}

\begin{figure}[!ht]
\centering

\begin{subfigure}{0.32\textwidth}
    \centering
    \includegraphics[width=\textwidth]{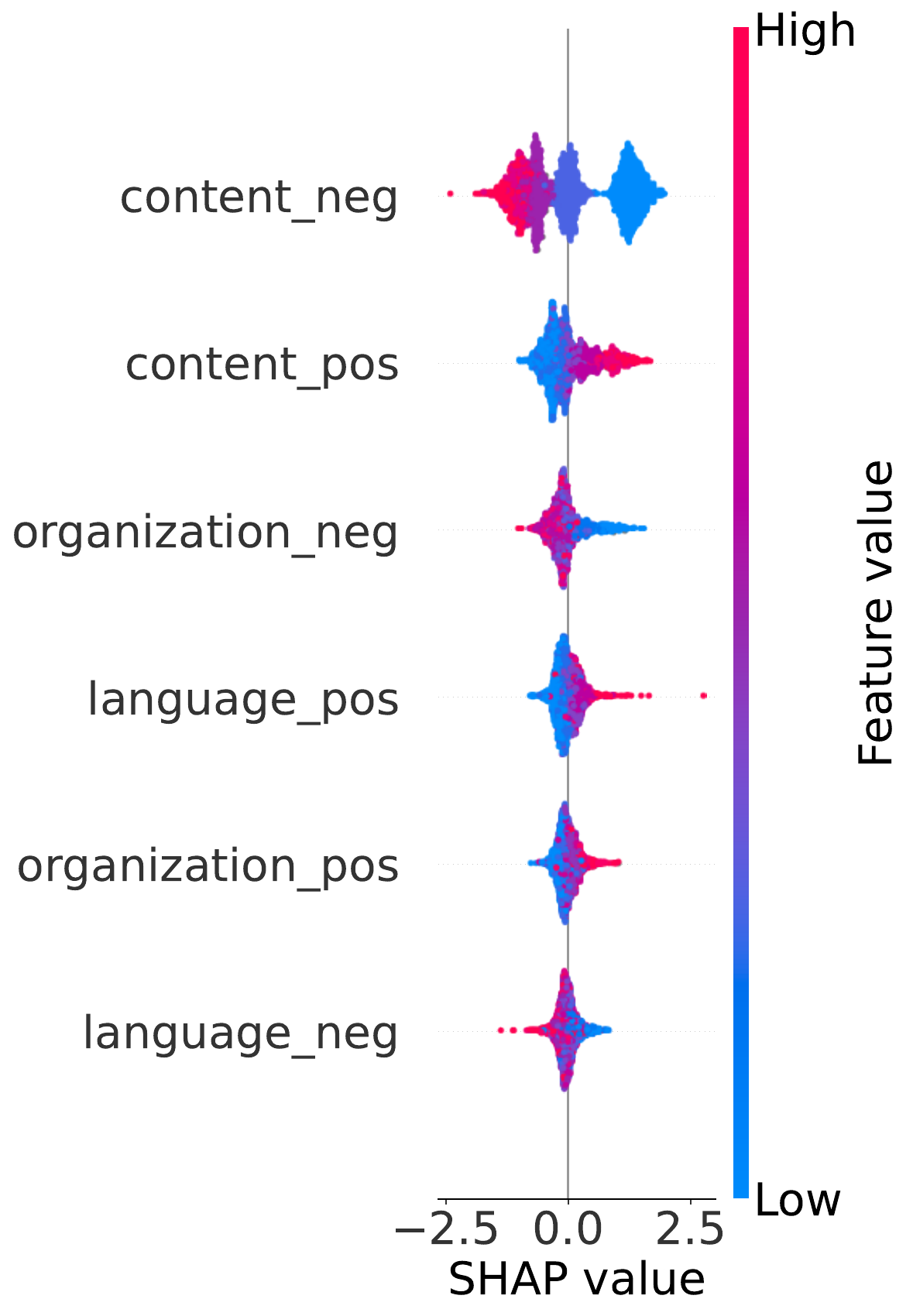}
    \caption{GPT-3.5}
\end{subfigure}
\hfill
\begin{subfigure}{0.32\textwidth}
    \centering
    \includegraphics[width=\textwidth]{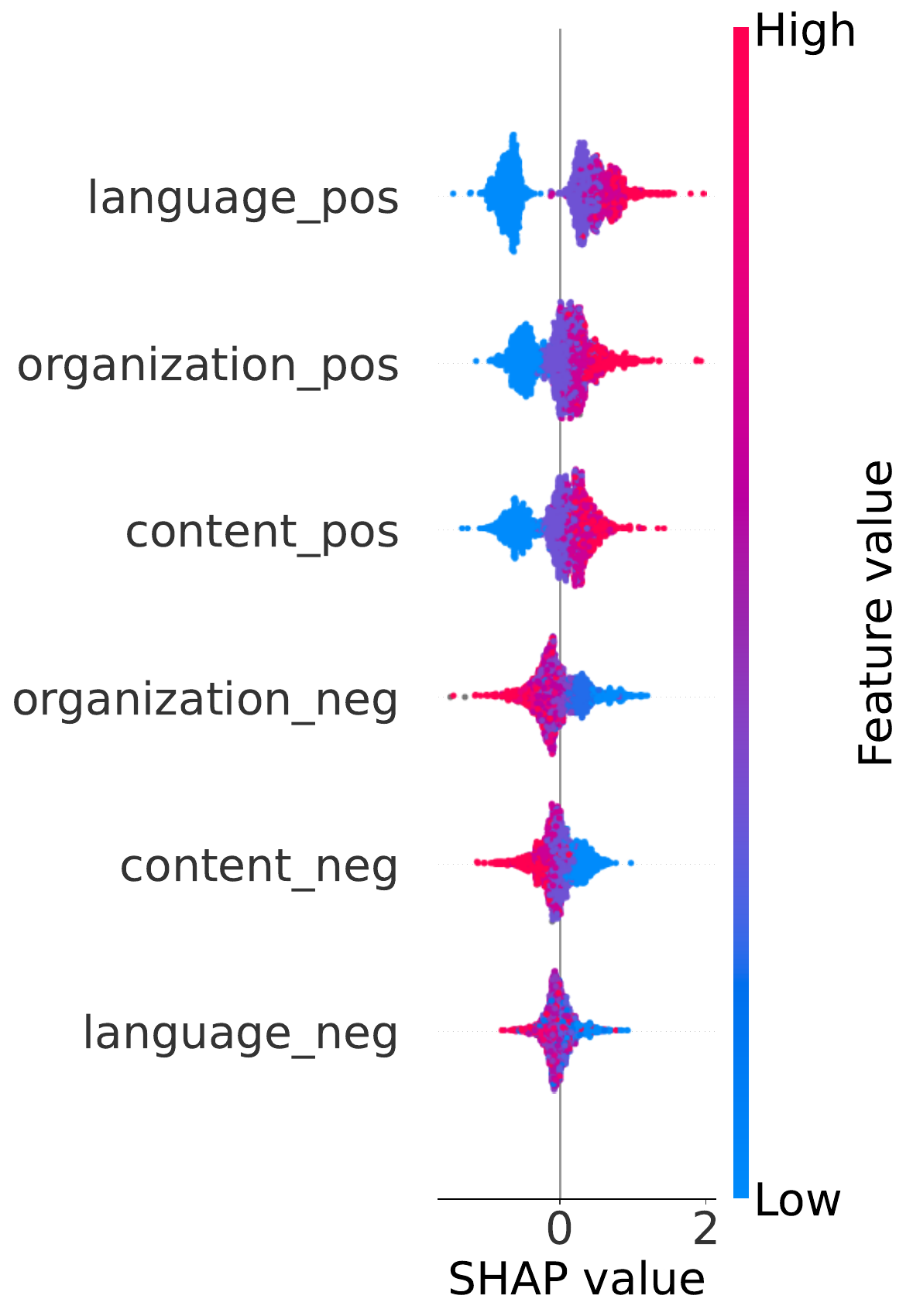}
    \caption{GPT-4}
\end{subfigure}
\hfill
\begin{subfigure}{0.32\textwidth}
    \centering
    \includegraphics[width=\textwidth]{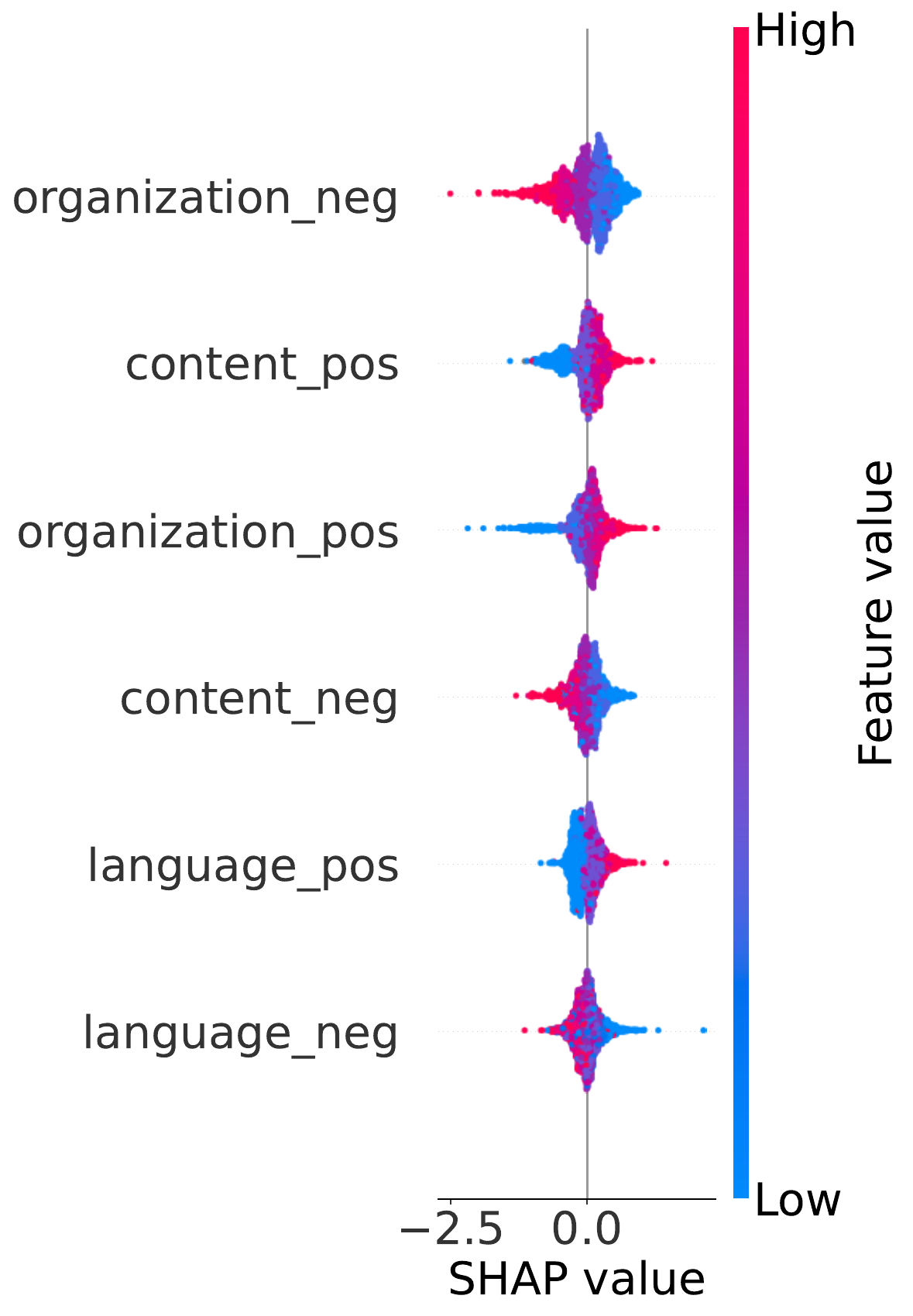}
    \caption{GPT-5}
\end{subfigure}

\caption{SHAP beeswarm plots showing the impact of trait-specific praise and criticism mentions on predicted essay scores for GPT models on DREsS. Features are sorted in decreasing order of importance (mean absolute SHAP value). Points represent individual essays, and horizontal position indicates the contribution to the predicted score. Praise-related features tend to increase predicted scores, while criticism-related features decrease them.}
\label{fig:beeswarm_gpt_dress}
\end{figure}

\begin{figure}[!ht]
\centering

\begin{subfigure}{0.32\textwidth}
    \centering
    \includegraphics[width=\textwidth]{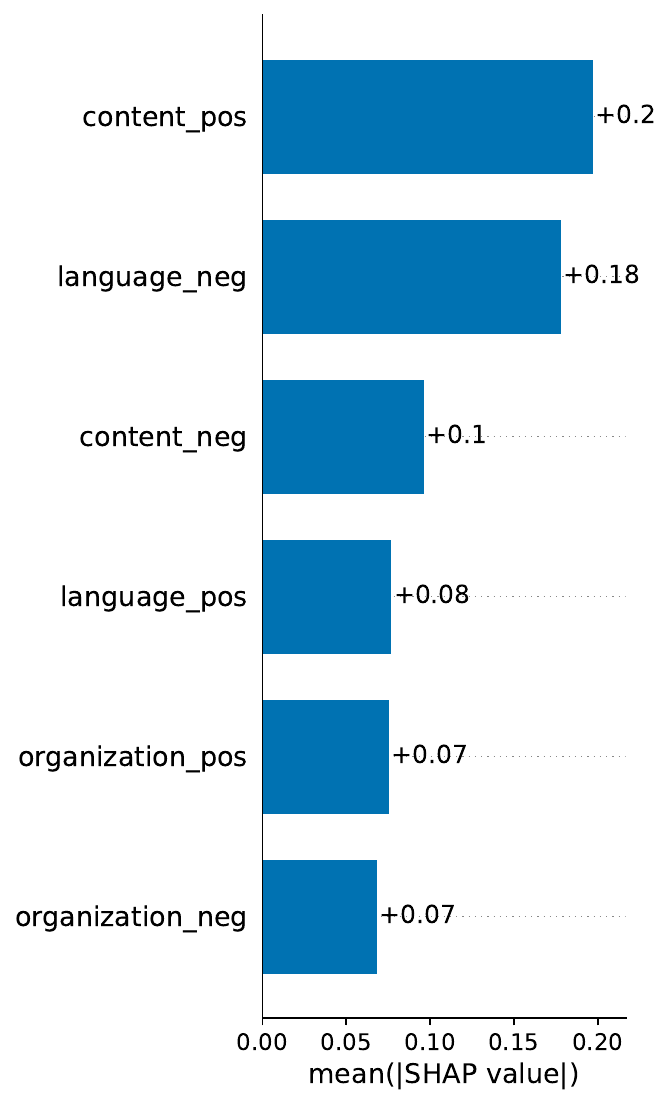}
    \caption{Llama 2}
\end{subfigure}
\hfill
\begin{subfigure}{0.32\textwidth}
    \centering
    \includegraphics[width=\textwidth]{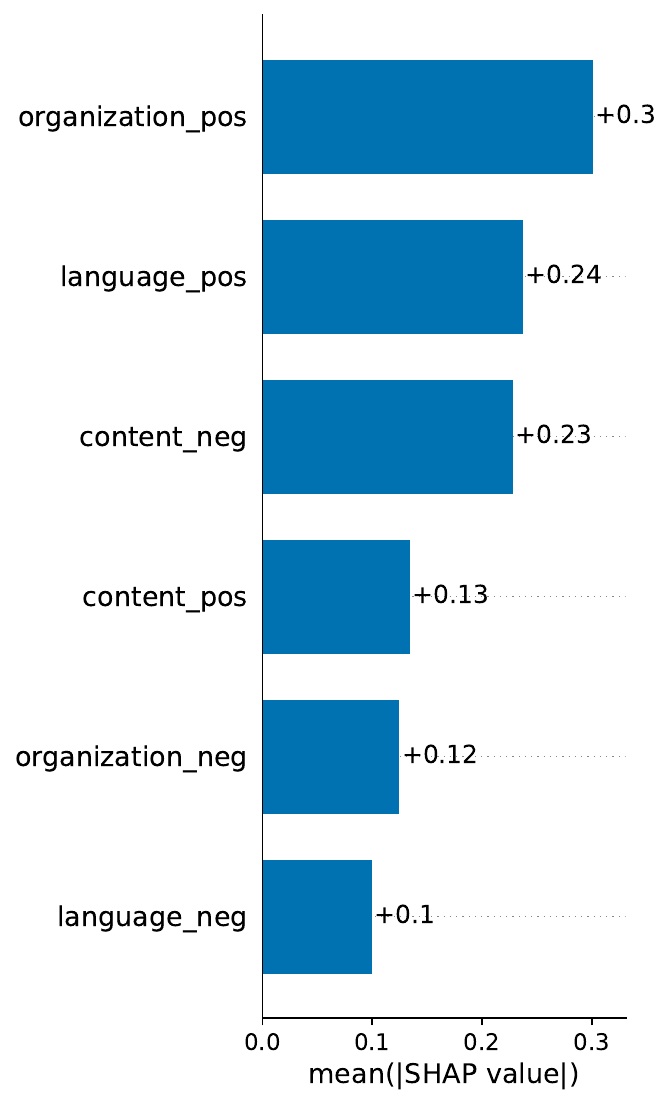}
    \caption{Llama 3}
\end{subfigure}
\hfill
\begin{subfigure}{0.32\textwidth}
    \centering
    \includegraphics[width=\textwidth]{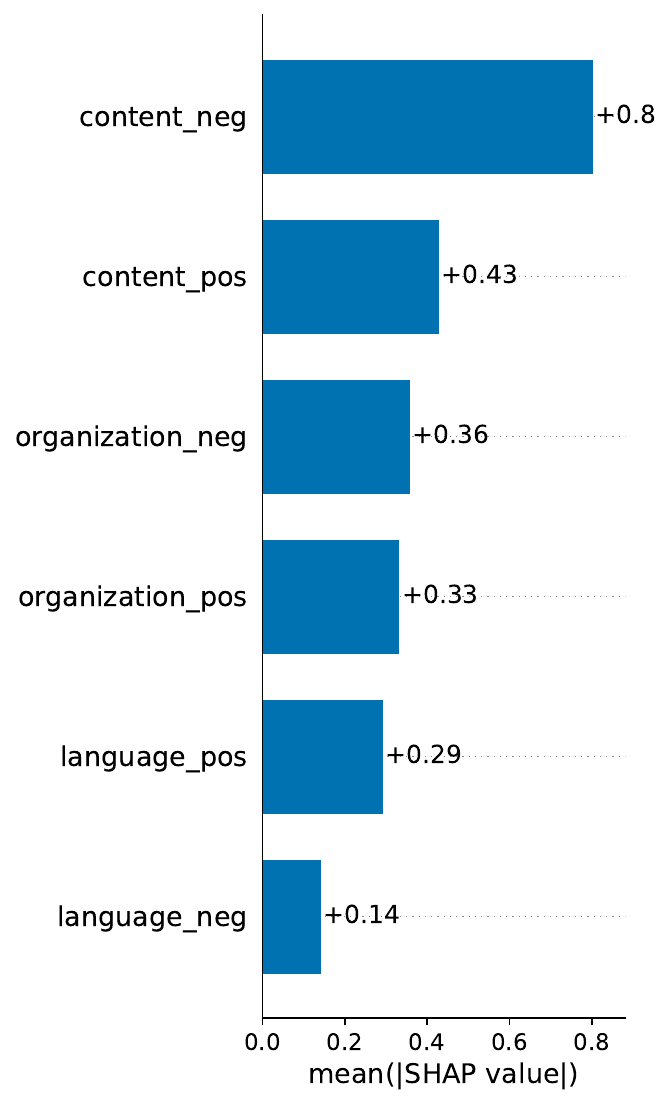}
    \caption{Llama 4}
\end{subfigure}

\caption{Average absolute SHAP values for trait-specific praise and criticism features for Llama models on DREsS. Higher values indicate features that contribute more strongly to predicted scores, revealing differences in how models prioritize rubric traits.}
\label{fig:shap_barplot_llama_dress}
\end{figure}

\begin{figure}[!ht]
\centering

\begin{subfigure}{0.32\textwidth}
    \centering
    \includegraphics[width=\textwidth]{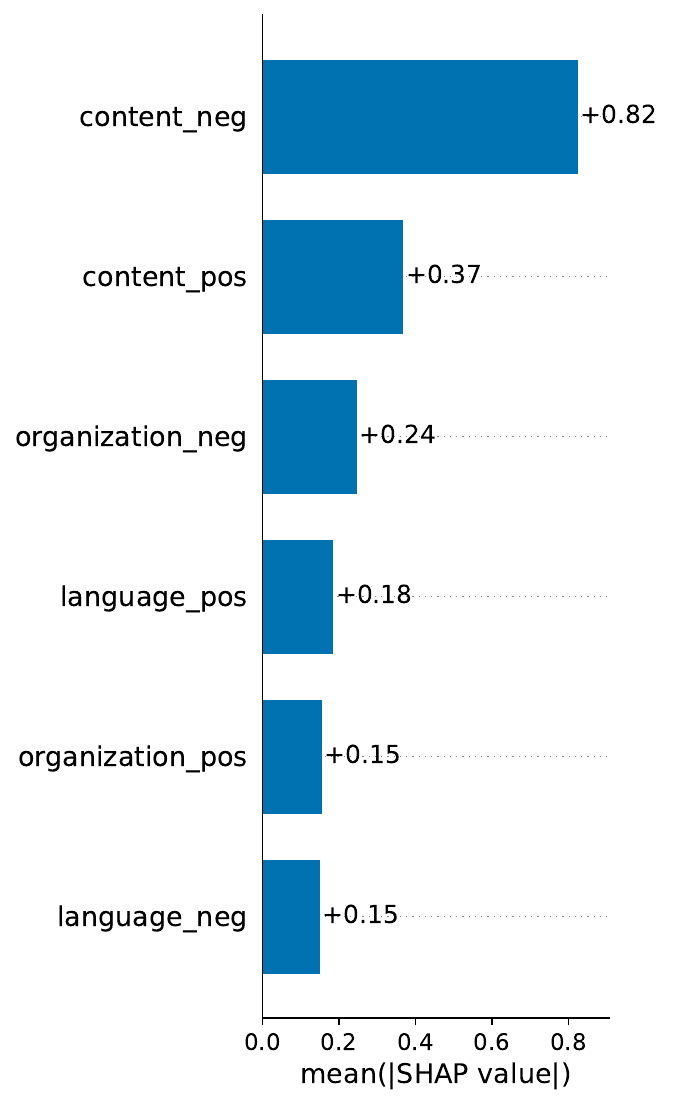}
    \caption{GPT-3.5}
\end{subfigure}
\hfill
\begin{subfigure}{0.32\textwidth}
    \centering
    \includegraphics[width=\textwidth]{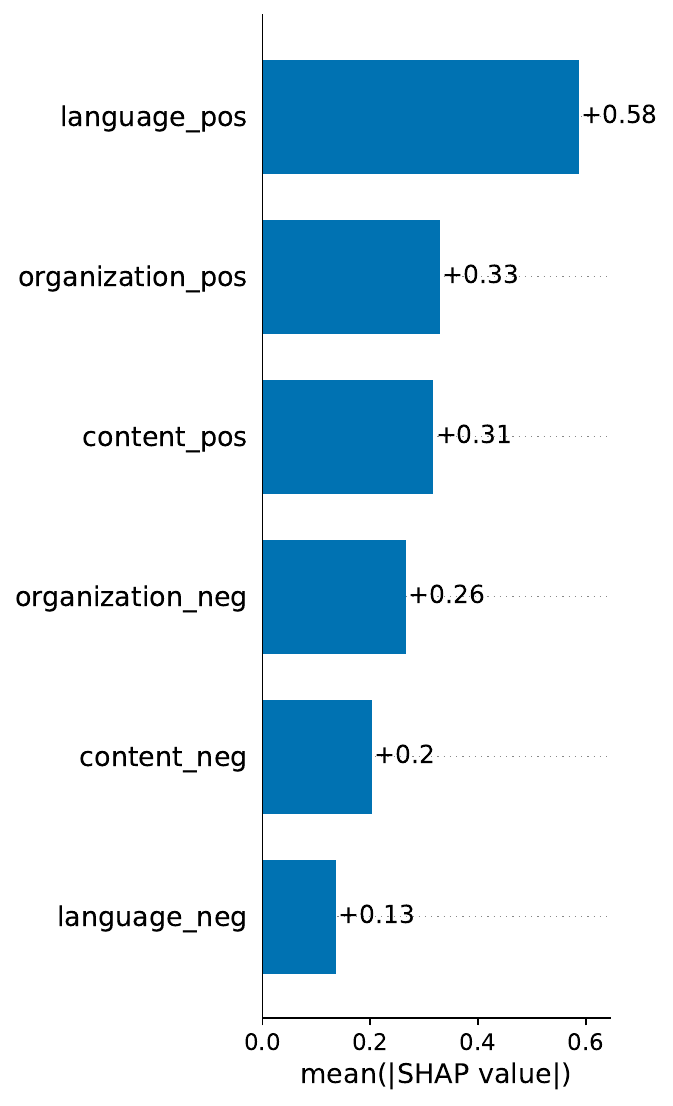}
    \caption{GPT-4}
\end{subfigure}
\hfill
\begin{subfigure}{0.32\textwidth}
    \centering
    \includegraphics[width=\textwidth]{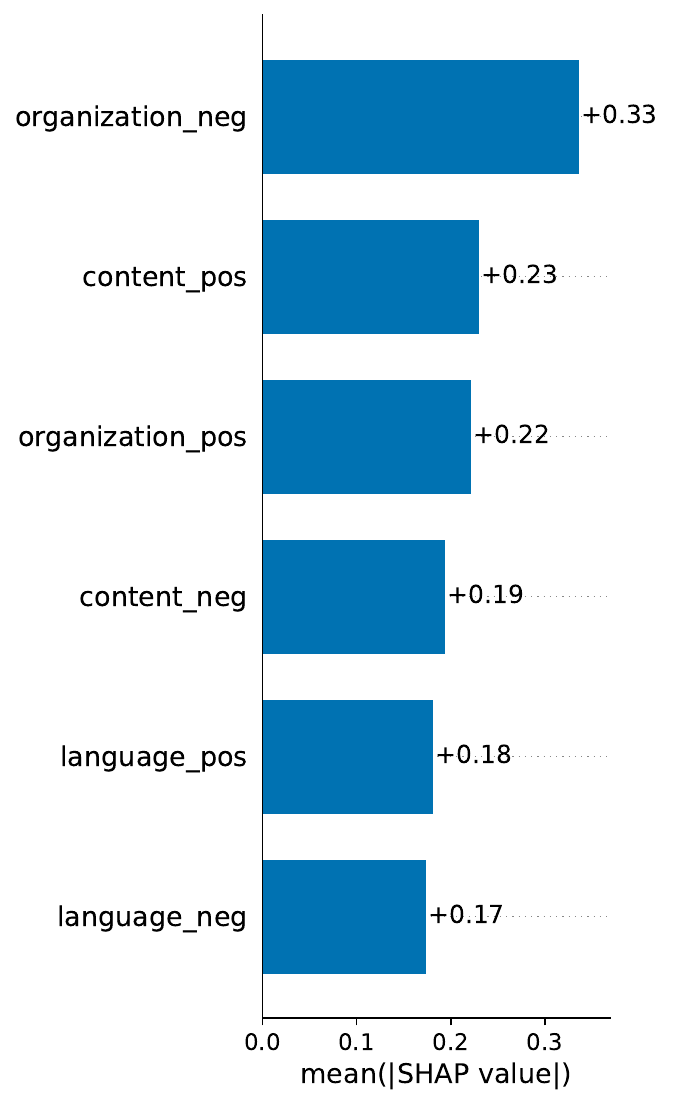}
    \caption{GPT-5}
\end{subfigure}

\caption{Average absolute SHAP values for trait-specific praise and criticism features for GPT models on DREsS. Higher values indicate features that contribute more strongly to predicted scores, revealing differences in how models prioritize rubric traits.}
\label{fig:shap_barplot_gpt_dress}
\end{figure}
\end{document}